\documentclass[runningheads]{llncs}

\usepackage{eccv}

\usepackage{eccvabbrv}

\usepackage{graphicx}
\usepackage{array}

\usepackage{booktabs}

\usepackage{booktabs}
\usepackage{multirow}
\usepackage{siunitx}
\usepackage[accsupp]{axessibility}  
\usepackage[ruled,vlined]{algorithm2e}
\usepackage{subcaption}
\usepackage[dvipsnames,table]{xcolor}
\usepackage{booktabs}
\usepackage{multirow}
\usepackage{graphicx} 
\usepackage{array}
\usepackage{float}
\usepackage{caption}
\usepackage{wrapfig}
\usepackage{graphicx}
\usepackage{makecell}

\usepackage{hyperref}

\usepackage{orcidlink}

\begin{document}

\title{Rethinking Token Reduction for Diffusion Models via Output-Similarity-Awareness} 
\titlerunning{DiTo}

\author{Hangyeol Lee\orcidlink{0009-0004-1015-588X} \and
Hyojeong Lee\orcidlink{0009-0004-0795-2044} \and
Joo-Young Kim\orcidlink{0000-0003-1099-1496}}

\authorrunning{H. Lee et al.}

\institute{KAIST, Daejeon, Republic of Korea \\
\email{\{lhg4294,hjlee8877,jooyoung1203\}@kaist.ac.kr}}

\maketitle
\vspace{-1.5em}
\renewcommand{\topfraction}{0.95}
\renewcommand{\bottomfraction}{0.95}
\renewcommand{\textfraction}{0.05}
\renewcommand{\floatpagefraction}{0.9}
\begin{figure}[h]
\centering
\setlength{\tabcolsep}{2pt}
\renewcommand{\arraystretch}{0.0}

\newcommand{\headh}{6ex} 
\newcommand{\headw}{0.155\linewidth} 

\newcommand{\mainheadh}{4ex} 

\begin{tabular}{c c@{}c@{}c@{}c@{}c}

\multicolumn{6}{c}{\parbox[c][\mainheadh][c]{0.93\linewidth}{\centering \textbf{Flux, Text-to-Image  (1024$\times$1024)}}} \\

\includegraphics[width=0.155\linewidth]{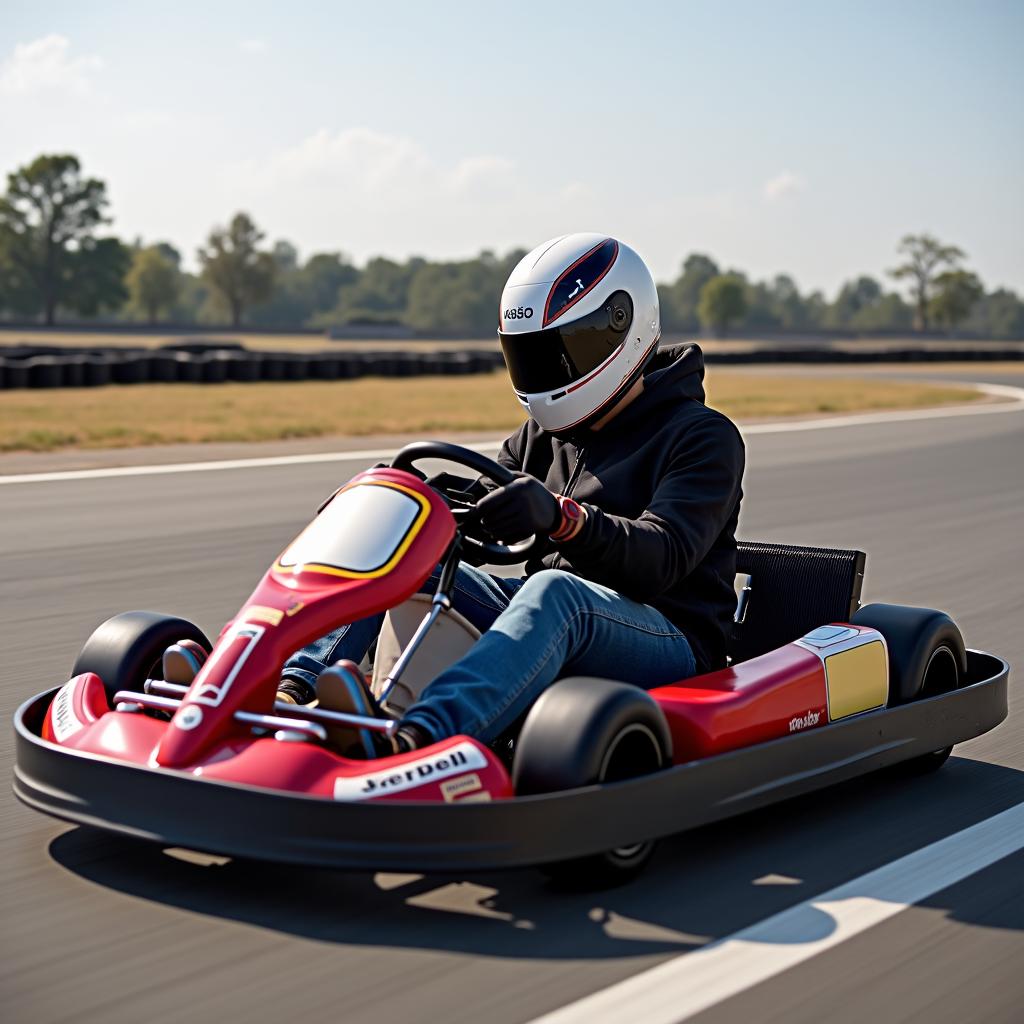} &
\includegraphics[width=0.155\linewidth]{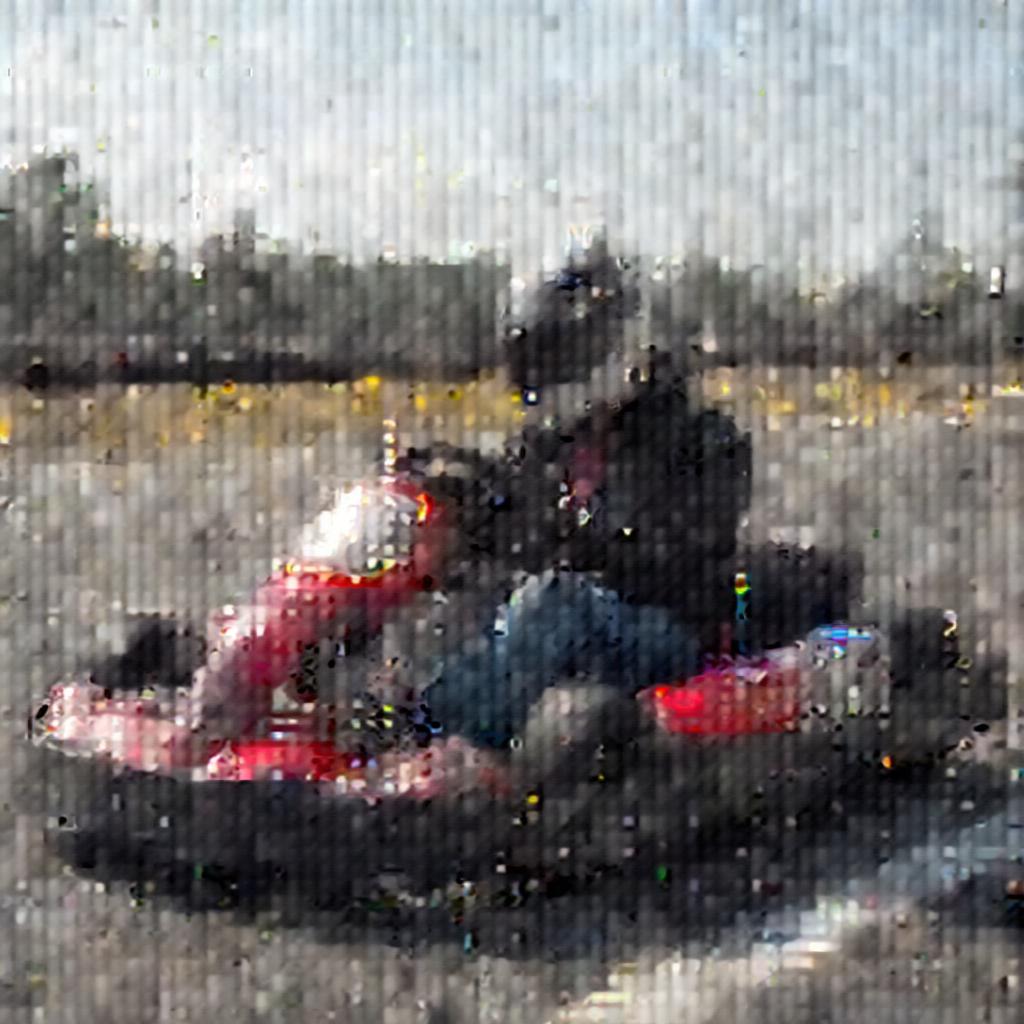} &
\includegraphics[width=0.155\linewidth]{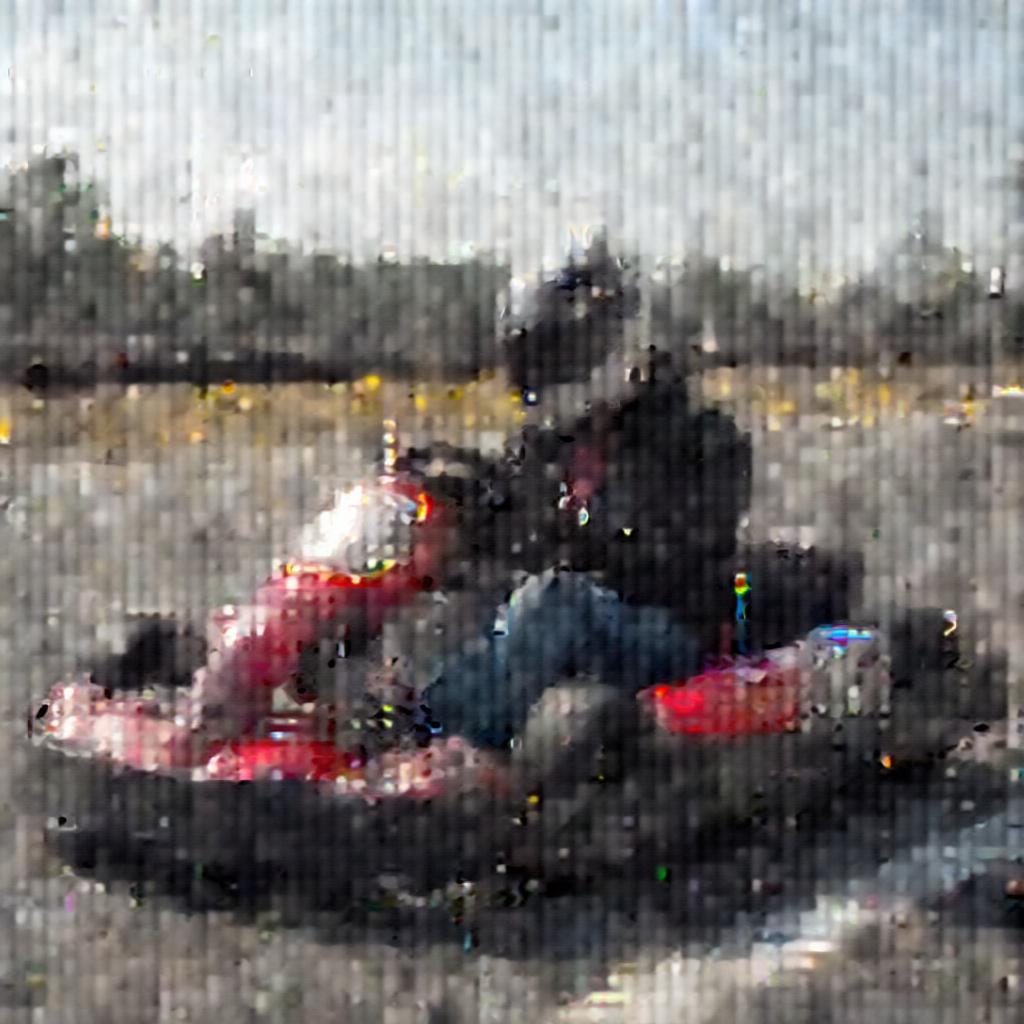} &
\includegraphics[width=0.155\linewidth]{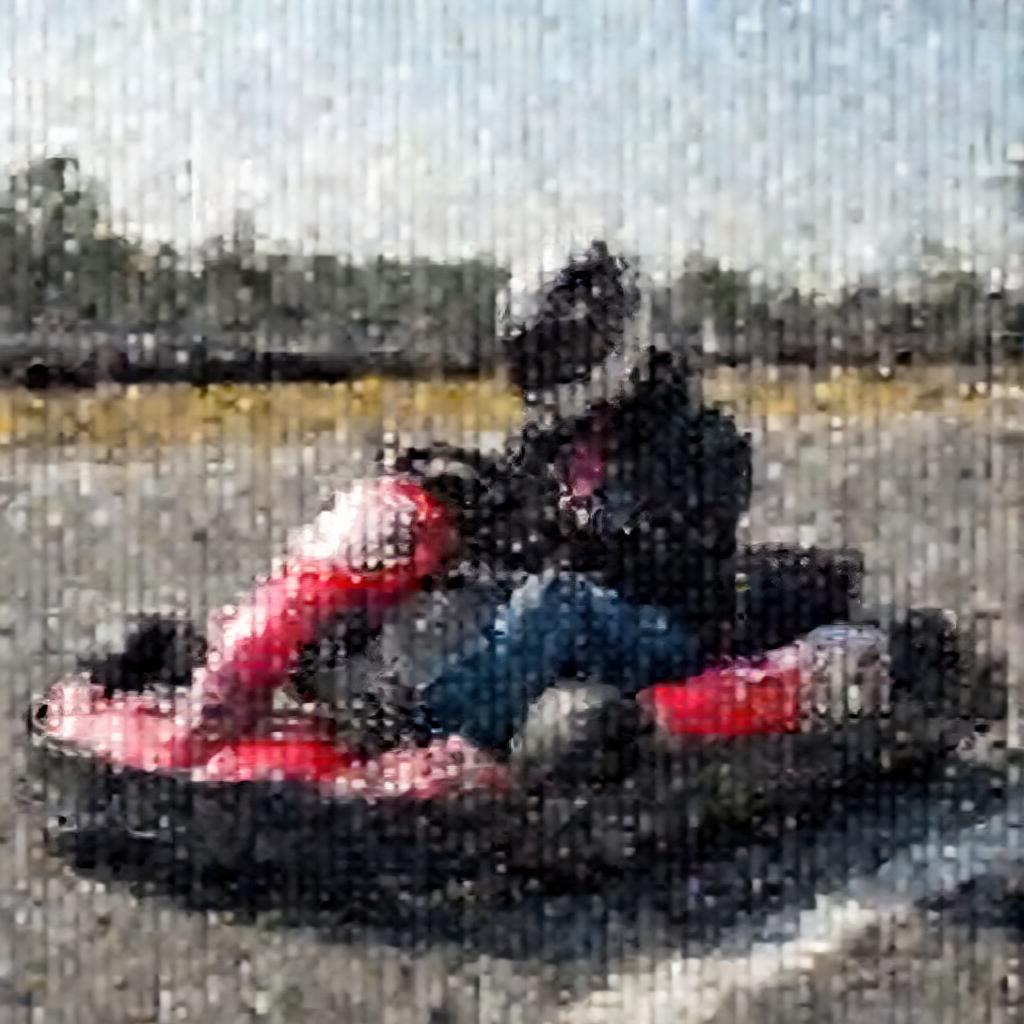} &
\includegraphics[width=0.155\linewidth]{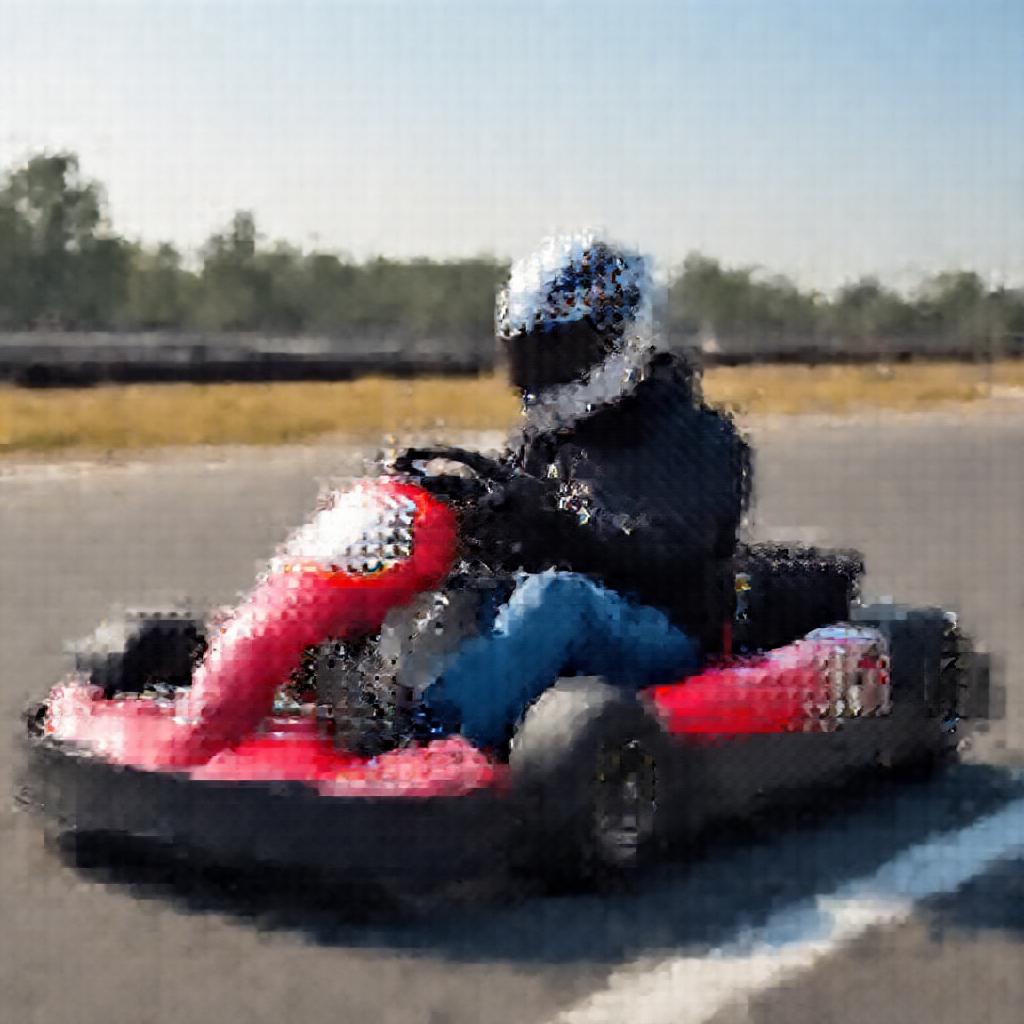} &
\includegraphics[width=0.155\linewidth]{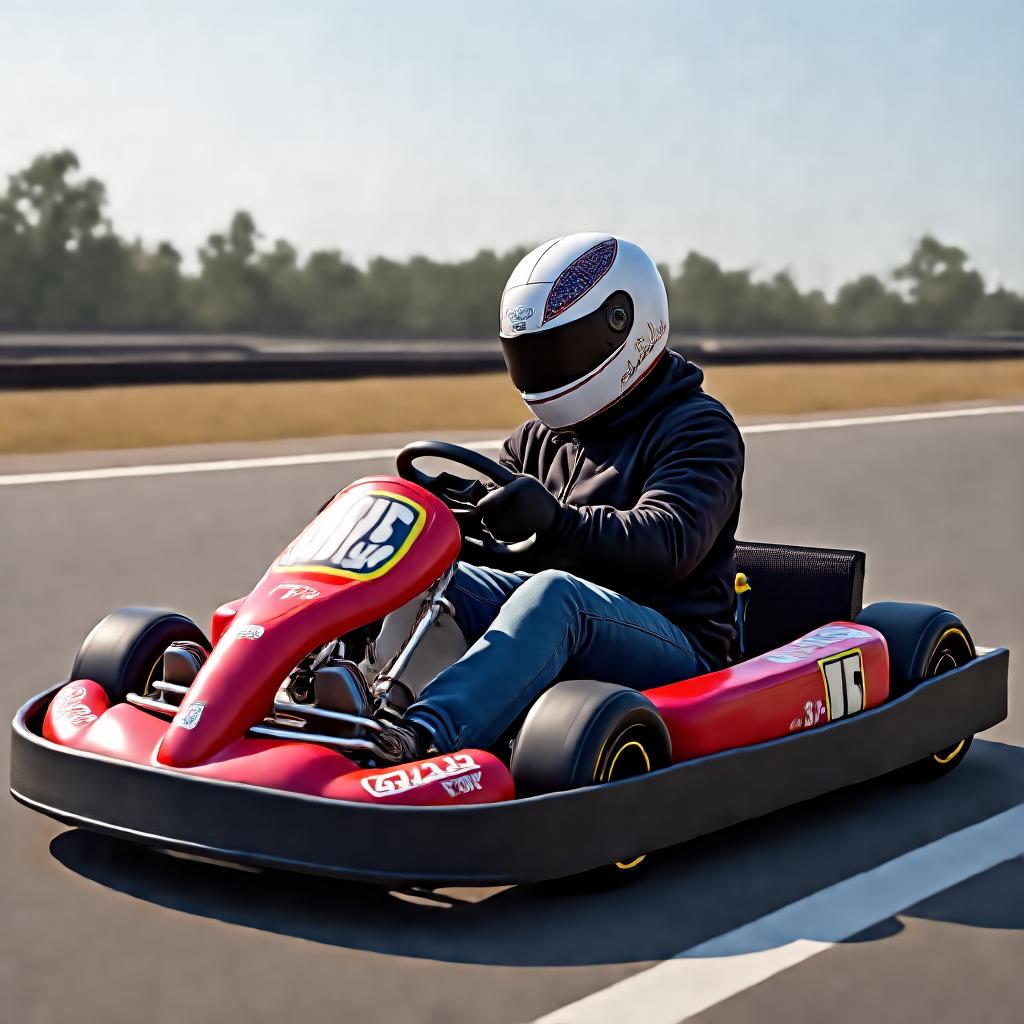} \\

\parbox[c][\headh][c]{\headw}{\centering Baseline \\ (1.00$\times$)} &
\parbox[c][\headh][c]{\headw}{\centering ToMeSD \\ (1.21$\times$)} &
\parbox[c][\headh][c]{\headw}{\centering ToFu \\ (1.21$\times$)} &
\parbox[c][\headh][c]{\headw}{\centering SiTo \\ (1.18$\times$)} &
\parbox[c][\headh][c]{\headw}{\centering ToMA \\ (1.23$\times$)} &
\parbox[c][\headh][c]{\headw}{\centering \textbf{DiTo} \\ \textbf{(1.23$\times$)}} \\
\end{tabular}

\vspace{0.1em}

\begin{tabular}{c c@{}c@{}c@{}c@{}c}

\multicolumn{6}{c}{\parbox[c][\mainheadh][c]{0.93\linewidth}{\centering \textbf{Stable Diffusion 3, Text-to-Image (1024$\times$1024)}}} \\

\includegraphics[width=0.155\linewidth]{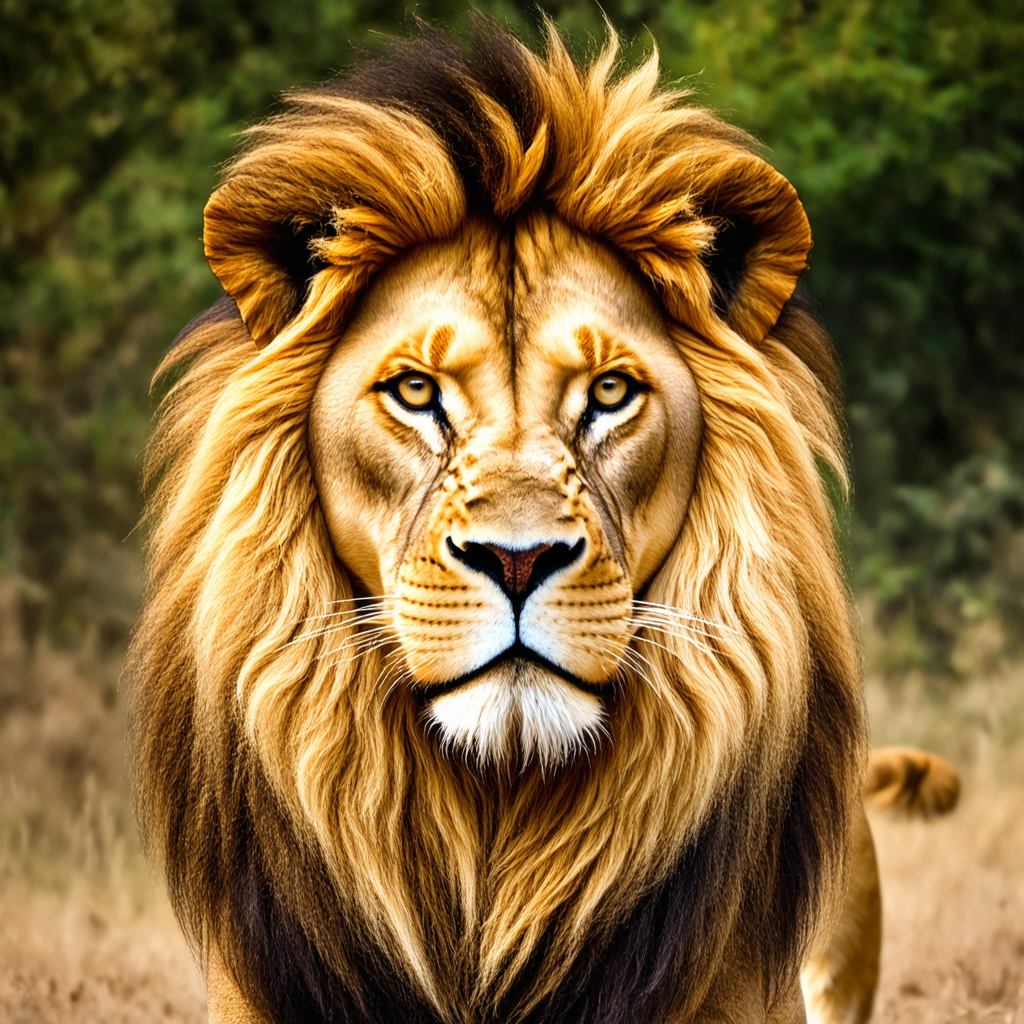} &
\includegraphics[width=0.155\linewidth]{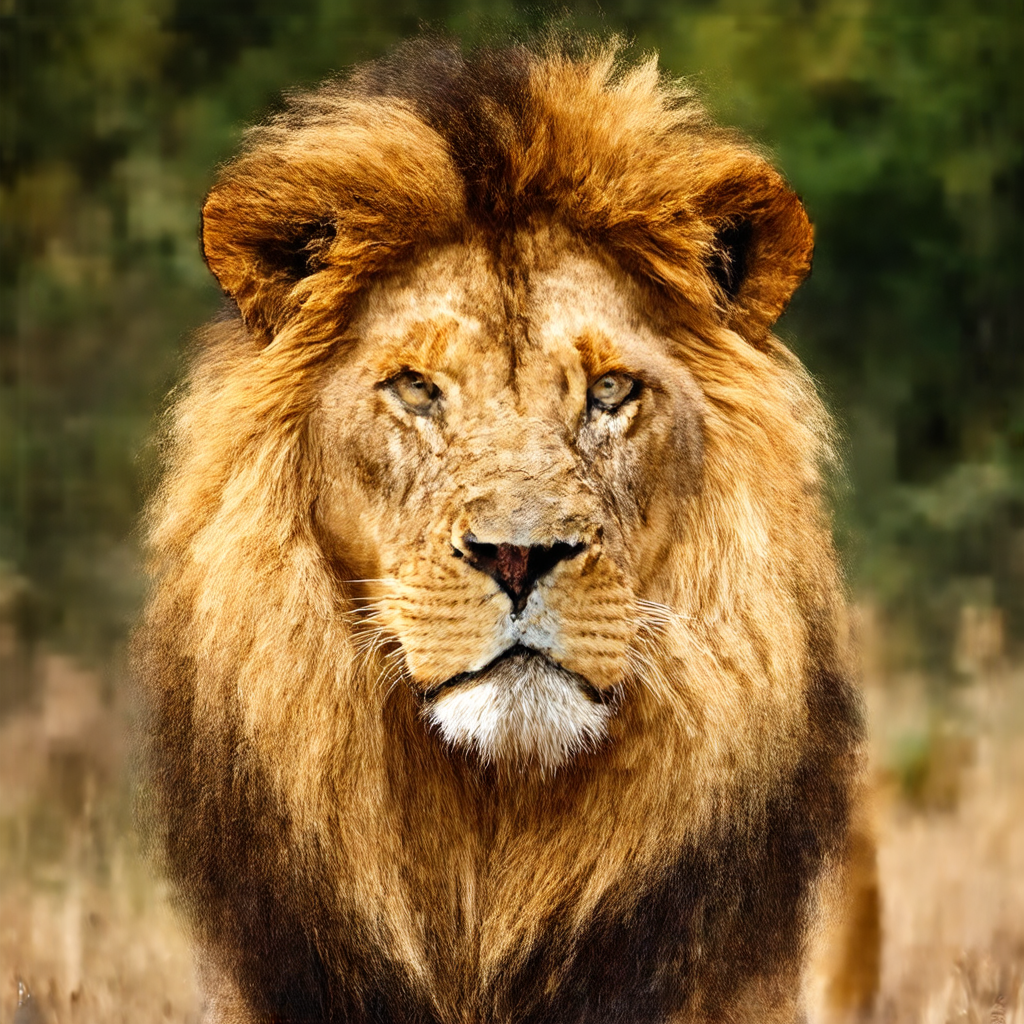} &
\includegraphics[width=0.155\linewidth]{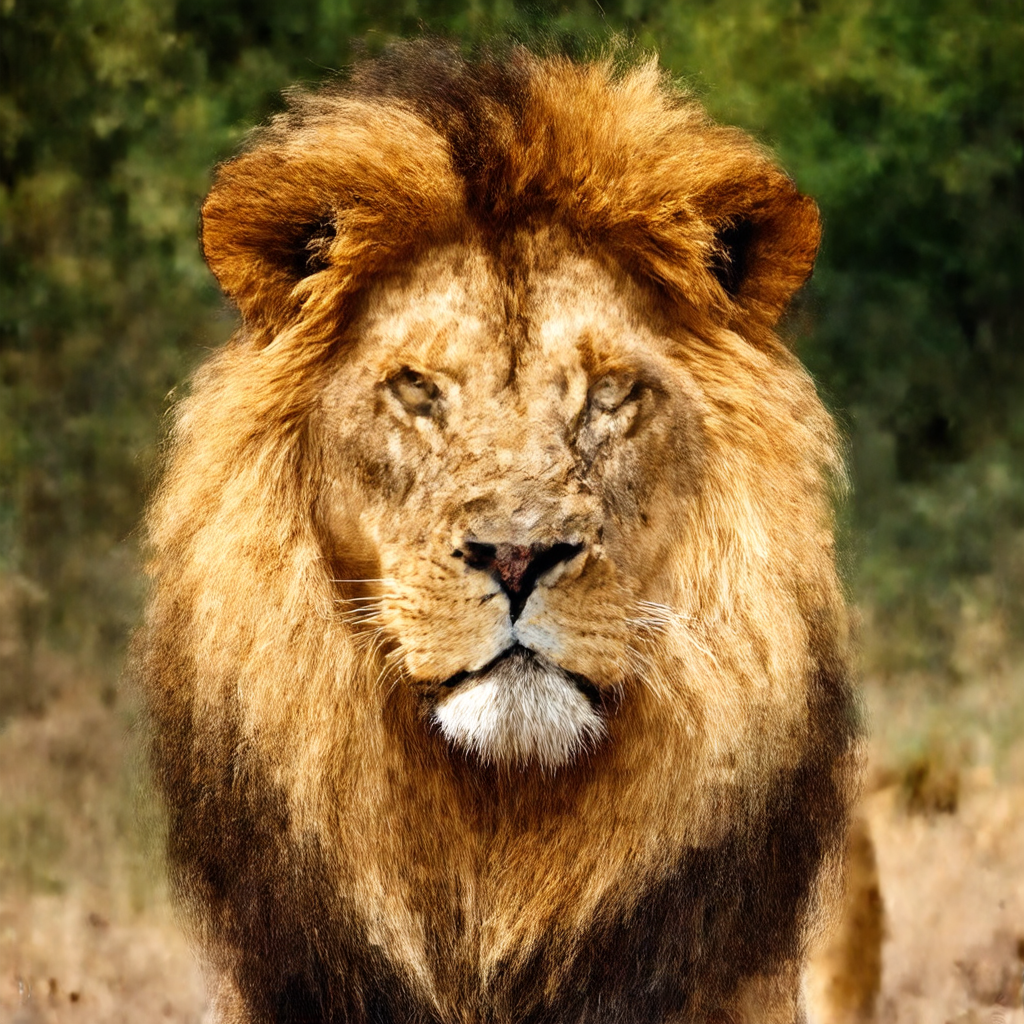} &
\includegraphics[width=0.155\linewidth]{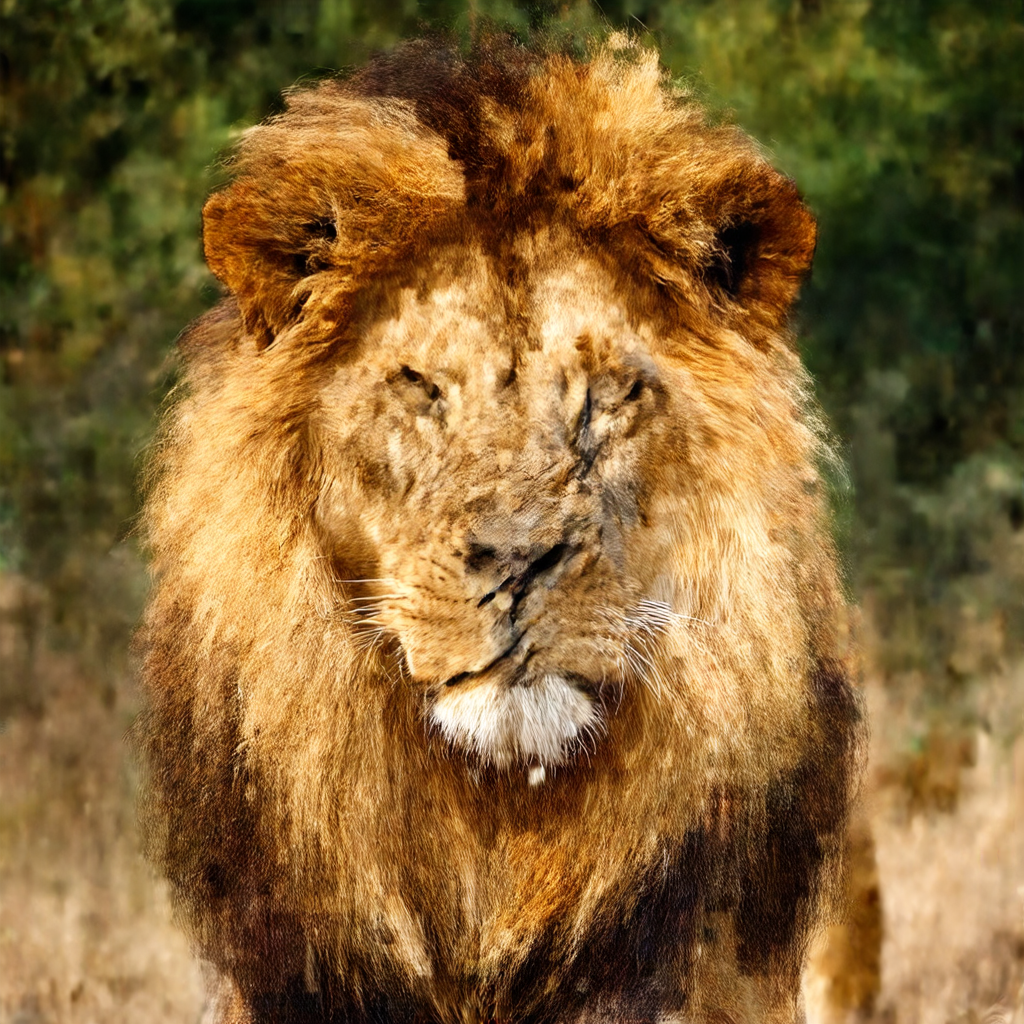} &
\includegraphics[width=0.155\linewidth]{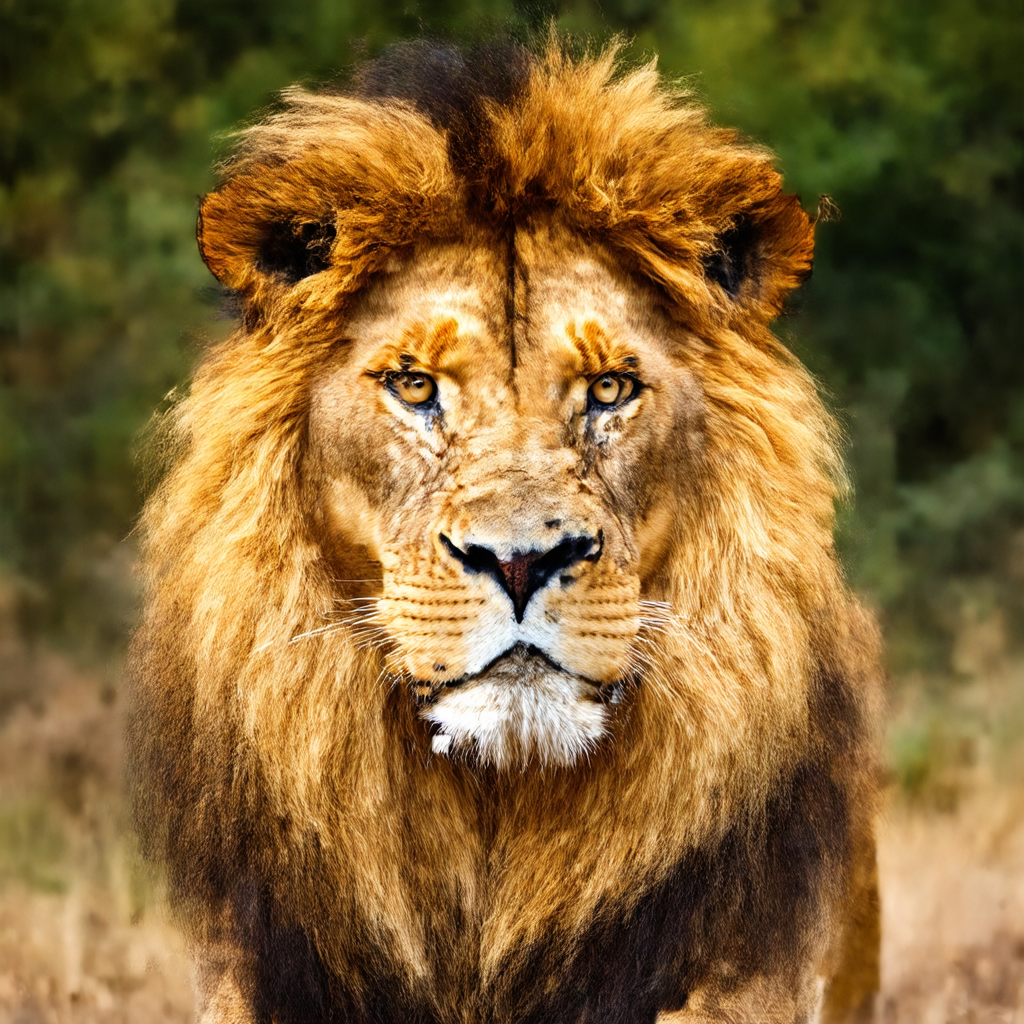} &
\includegraphics[width=0.155\linewidth]{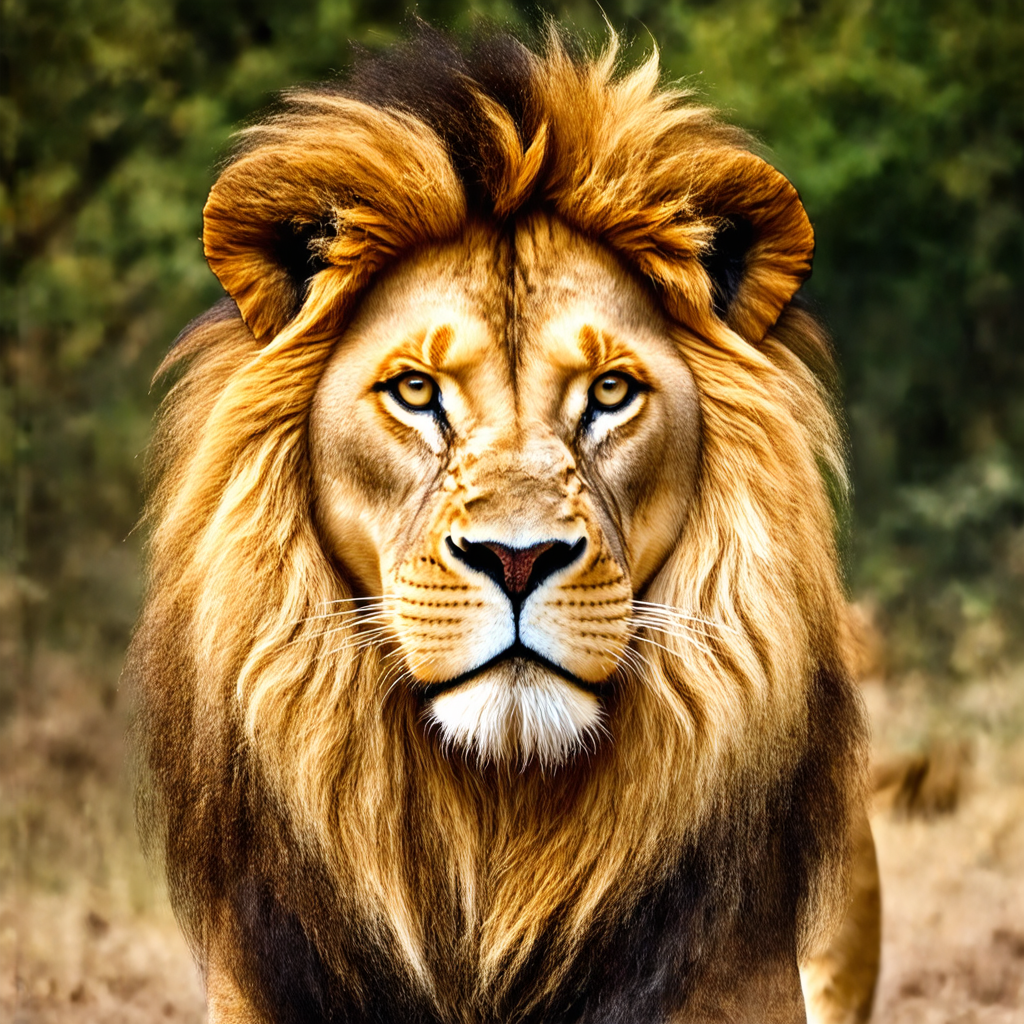} \\

\parbox[c][\headh][c]{\headw}{\centering Baseline \\ (1.00$\times$)} &
\parbox[c][\headh][c]{\headw}{\centering ToMeSD  \\ (1.17$\times$)} &
\parbox[c][\headh][c]{\headw}{\centering ToFu \\ (1.16$\times$)} &
\parbox[c][\headh][c]{\headw}{\centering SiTo \\ (1.11$\times$)} &
\parbox[c][\headh][c]{\headw}{\centering ToMA \\ (1.15$\times$)} &
\parbox[c][\headh][c]{\headw}{\centering \textbf{DiTo} \\ \textbf{(1.20$\times$)}} \\
\end{tabular}

\caption{
\textbf{Visual comparisons on Flux and SD3 with prior token reduction methods.} Under the equivalent FLOPs reduction, DiTo preserves superior visual fidelity and structural consistency with comparable latency, while others suffer from severe structural distortion.}

\label{fig:visual_comparison_w_prior_tr}
\end{figure}

\vspace{-2.5em}

\begin{abstract}
Diffusion Transformers (DiTs) achieve superior image generation quality but suffer from quadratic computational complexity relative to token count. While various token reduction (TR) methods have been proposed to mitigate this cost, they overlook the primary objective of generative models: minimizing recovery error, which requires reflecting output token similarity. They rely solely on input token similarity inherited from reduction-only ViT paradigms, leading to a fundamental misalignment with this objective.

To bridge this gap, we propose \textbf{DiTo}, a novel TR paradigm that shifts the focus toward output-centric token reduction. Based on the observation that output token similarity is consistently preserved across adjacent timesteps, DiTo utilizes prior-step similarities as an effective proxy to establish token correspondences at a Matching timestep, which are then reused across multiple subsequent Reduction timesteps. To optimize this interleaved scheduling, we propose Pair Match Ratio (PMR)-guided Interval Scheduling to determine the optimal matching frequency. Furthermore, to mitigate localized approximation errors and resulting blocking artifacts caused by repeated reuse, we propose Frequency-aware Token Matching by incorporating a selection-frequency penalty. Extensive experiments demonstrate that DiTo consistently outperforms existing TR methods with 1.6–3.9 dB higher PSNR at comparable speedups, achieving a superior Pareto frontier.
\end{abstract}
\section{Introduction}

\vspace{-2.0em}

\begin{figure}[h]
  \centering
  \begin{subfigure}{0.48\textwidth}
    \centering
    \includegraphics[width=\linewidth]{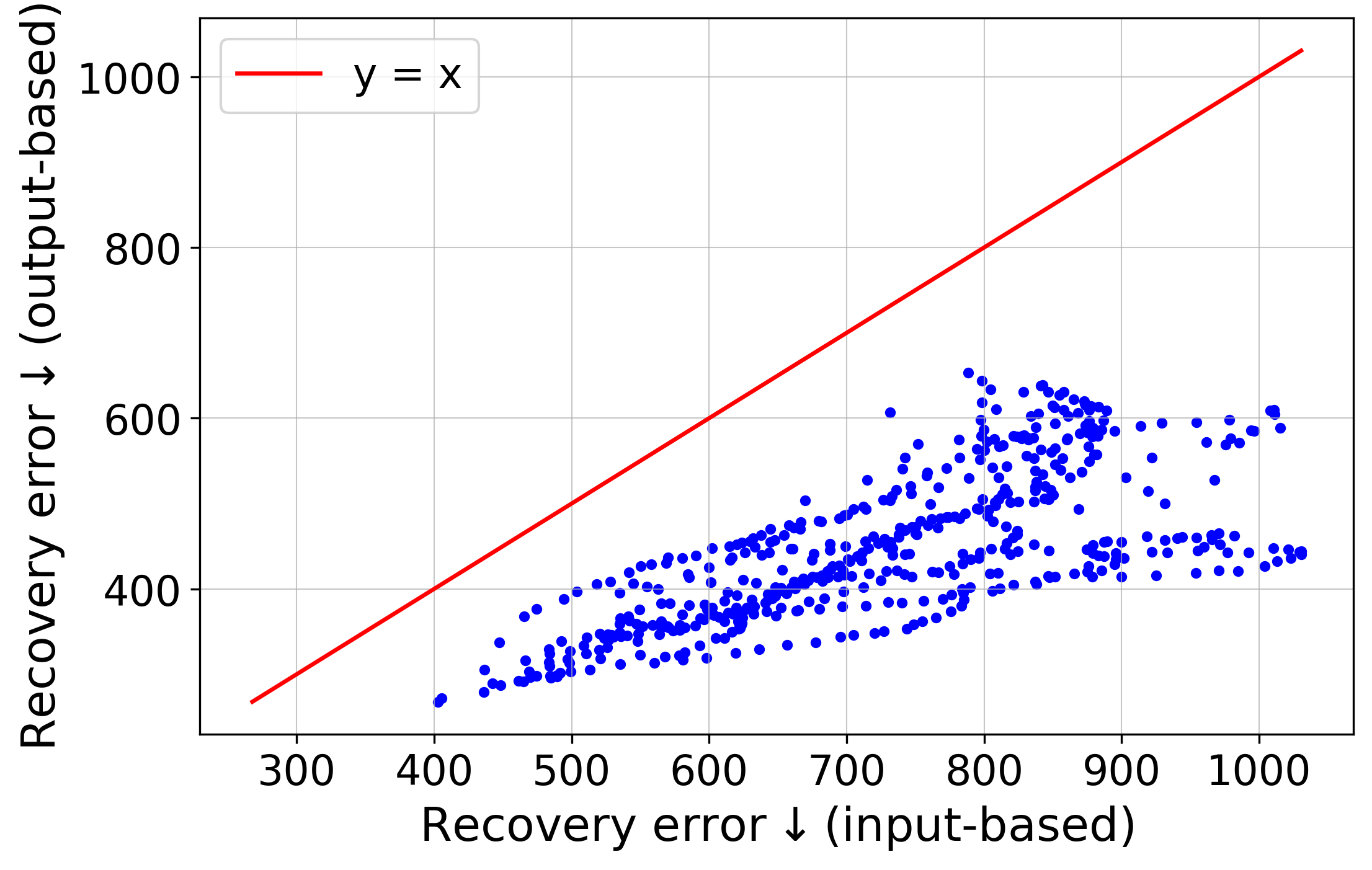}
    \caption{Comparison of recovery errors.}
    \label{fig:scatter_recovery_error}
  \end{subfigure}
  \hfill
  \begin{subfigure}{0.48\textwidth}
    \centering
    \includegraphics[width=\linewidth]{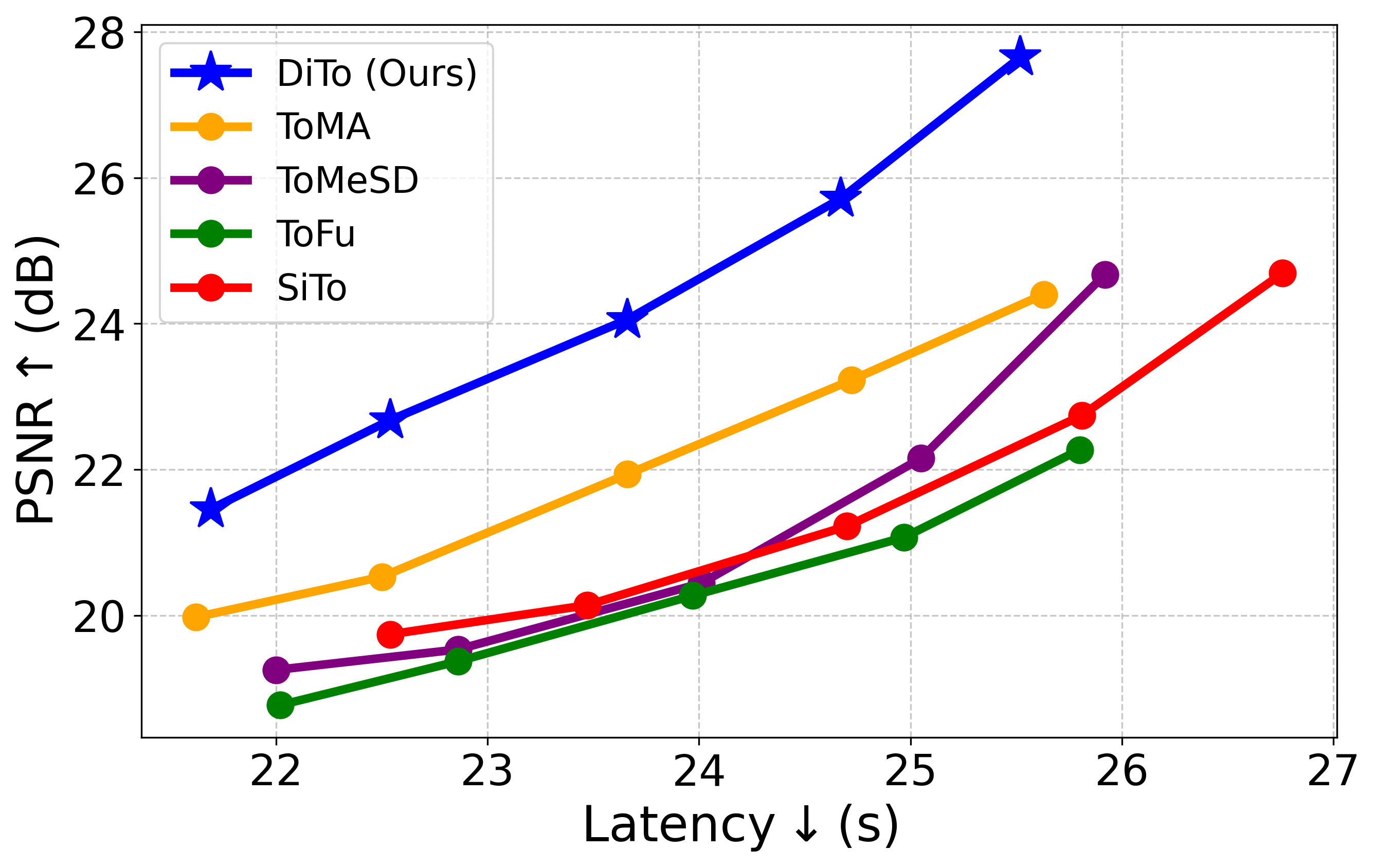}
    \caption{Quality vs. Efficiency trade-off.}
    \label{fig:perto_tradeoff}
  \end{subfigure}
  
  \caption{\textbf{Analysis of recovery error and performance trade-offs.} 
  (a) All 500 randomly selected samples fall below the $y=x$ line, indicating that output-based TR methods consistently achieve lower recovery error than existing input-based methods.
  (b) DiTo achieves a superior Pareto frontier in terms of generation quality (PSNR) and efficiency compared to existing methods.}

  \label{fig:overall_analysis}
\end{figure}
\vspace{-1.0em}
Diffusion Transformers (DiTs) \cite{Peebles2022DiT, esser2024scaling, chen2024pixartsigma, flux2024} have emerged as a powerful paradigm for high-fidelity image synthesis, primarily due to their scalable architecture composed of transformer blocks. Despite their impressive performance, the attention mechanism—the core component of transformers—exhibits quadratic complexity relative to the number of tokens. This scaling property leads to prohibitive computational overhead, establishing a significant bottleneck, particularly for high-resolution generation where the token count is substantial.

While FlashAttention \cite{dao2023flashattention2} and xFormers \cite{xFormers2022} have successfully accelerated transformers through kernel-level optimization, they do not address the underlying computational complexity. To mitigate this, token reduction (TR) approaches have emerged as a promising solution\cite{bolya2023tome, meng2022adavit, Yin_2022_CVPR, ATS}, aiming to directly reduce the token count by pruning or merging redundant tokens. For Vision Transformers (ViTs), these methods are primarily designed for discriminative tasks, focusing only on redundancy removal\cite{liang2022not, rao2021dynamicvit, ATS, pan2021iared2, kong2022spvit, Kim_2024_WACV, yang2025kff, marin2023tokenpooling, visapp25, Yin_2022_CVPR, bolya2023tome, meng2022adavit}. In contrast, generative tasks like Diffusion Models (DMs) necessitate an additional recovery stage, making such reduction-only methods difficult to apply directly.

In DMs, tokens correspond to specific spatial positions within the output; thus, the original token count must be preserved through an explicit recovery stage.
Accordingly, the TR process begins with the token matching stage, which establishes token correspondences that govern both the reduction and recovery stages by specifying how tokens are to be reduced and subsequently restored.
Existing TR methods \cite{bolya2023tomesd, Kim_2024_WACV, lu2025toma,zhang2025training,smith2024todo} inherit ViT-style paradigms that rely on input token similarity for the matching stage.
However, within this recovery-inclusive TR framework, the primary objective shifts toward minimizing the recovery error—the discrepancy between the restored output and the original dense output. 
Since the recovery stage reconstructs reduced tokens by copying the outputs of their matched tokens, the error is fundamentally determined by how closely the matched tokens' outputs represent the original outputs of the reduced ones. Therefore, to ensure a high-fidelity reconstruction, TR must be designed to reflect output token similarity rather than existing TR methods.
Our empirical analysis in \cref{fig:scatter_recovery_error} supports this necessity by evaluating the recovery errors. In this scatter plot, the $x$ and $y$ axes represent the recovery errors when applying TR via input-based and output-based matching, respectively. Notably, every evaluated sample falls below the $y=x$ line, demonstrating that output-based TR consistently achieves lower recovery errors than conventional input-based TR, which fail to align with the goal of error minimization.
\captionsetup{justification=centering}
\begin{figure}[tb]
  \centering
  \begin{subfigure}{0.48\linewidth}
    \centering
    \includegraphics[width=\linewidth,keepaspectratio]{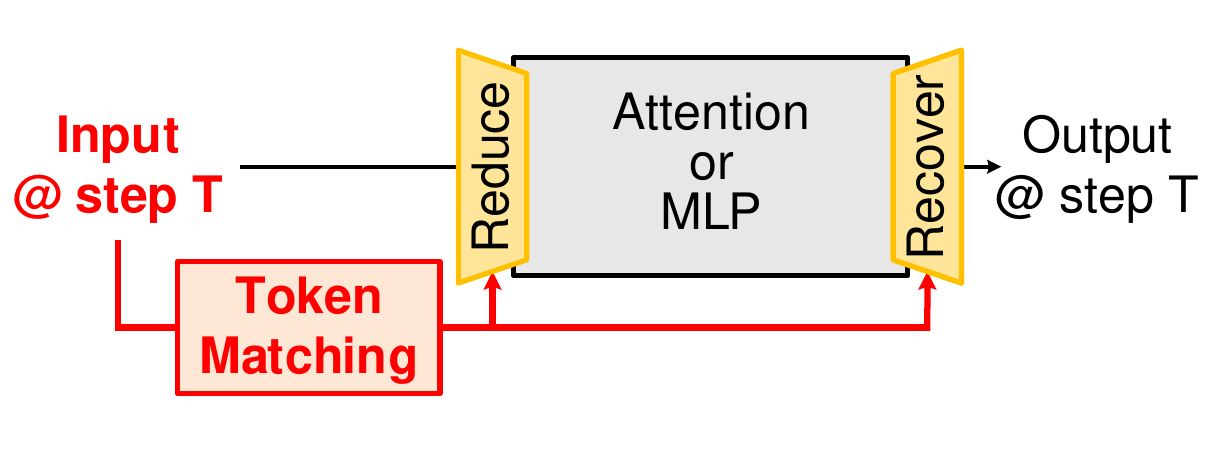}
    \caption{Existing Token Reduction Method \\ (Input-based)}
    \label{fig1:existing_method}
  \end{subfigure}
  \hfill
  \begin{subfigure}{0.48\linewidth}
    \centering
    \includegraphics[width=\linewidth,keepaspectratio]{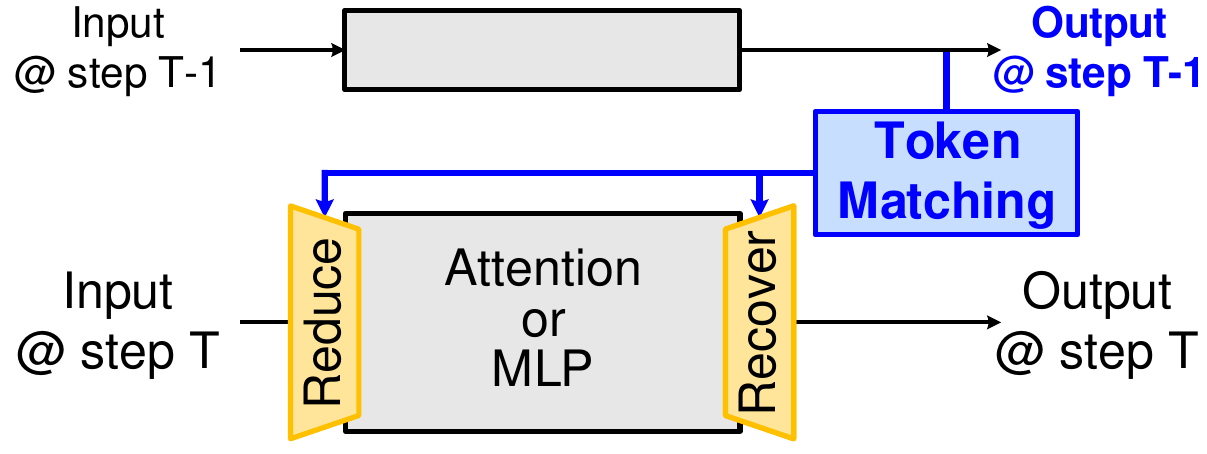}
    \caption{Proposed Token Reduction Method \\ (Output-based)}
    \label{fig1:our_method}
  \end{subfigure}
  \captionsetup{justification=justified}
    \caption{\textbf{Comparison of token reduction methods.} (a) Existing input-based methods rely on current input token similarity. (b) The output-based method utilizes prior output token similarity, minimizing recovery error and enhancing generation quality.}
  \label{fig1}
\end{figure}
\captionsetup{justification=justified}


To bridge this gap, we propose \textbf{Di}ffusion-specific \textbf{To}ken Reduction (\textbf{DiTo}), which marks a paradigm shift towards output-centric token reduction, explicitly reflecting output token similarity to minimize recovery error.
Since the current output is not yet available during the token matching stage, DiTo exploits the temporal consistency inherent in DMs as an effective proxy. 
While it is well-established that activations remain highly correlated across adjacent timesteps \cite{BiDM, parallel_denoising, streamlined}, we further observe that token similarity is likewise consistently preserved.
Based on this insight, we leverage the output token similarity from the prior timestep to guide the matching stage for the current reduction-recovery stages (\cref{fig1:our_method}).
This strategy partitions the diffusion timesteps into an interleaved schedule of matching timesteps, where token correspondences are established, and subsequent reduction timesteps that reuse these correspondences across multiple steps to perform efficient reduction and recovery.
To effectively optimize the matching intervals, we introduce the Pair Match Ratio (PMR), a new metric that quantifies the temporal alignment with the current-step output’s matching results. Since DiTo relies on prior-step similarities, PMR serves as a reliable indicator for determining how long a matching result remains valid before alignment degrades. Leveraging this metric, we propose PMR-guided Interval Scheduling to find an optimal balance between computation efficiency and matching accuracy in an offline manner. While DiTo improves efficiency through the reuse of matching results, repeated application can exacerbate token-selection imbalance for reduction, concentrating approximation errors in localized regions and causing blocking artifacts. To mitigate this, we propose Frequency-aware Token Matching, which incorporates a similarity penalty based on token selection history to effectively suppress these artifacts.

With these designs, DiTo consistently outperforms existing input-based token reduction methods on representative DiT models, including \texttt{Flux} and \texttt{SD3}. DiTo markedly outperforms existing methods in visual preservation (\cref{fig:visual_comparison_w_prior_tr}), achieving 1.6--3.9 dB higher PSNR at similar or slightly higher speedups. 
These results demonstrate that DiTo effectively breaks the quality-efficiency trade-off, achieving a superior Pareto frontier (\cref{fig:perto_tradeoff}).
\section{Related Work}

\subsection{Acceleration for DMs}

While diffusion models deliver exceptional generative performance, they incur high computational costs due to the extensive number of sampling timesteps and the heavy computational cost required per step.
Accordingly, acceleration methods primarily focus on either reducing the sampling steps $T$ or lowering the per-step cost. 
To reduce $T$, sampler-based approaches~\cite{song2021ddim, lu2022dpmsolver, lu2025dpmsolverpp, zhao2023unipc, zhou2024fastode} employ higher-order numerical solvers, while distillation-based approaches~\cite{salimans2022progressive,meng2023guided_distill,luo2023lcm,yin2024dmd2} compress multi-step teacher models into few-step students. But these approaches typically require additional training.

To reduce the per-step cost, various methods have been proposed: layer-distillation methods~\cite{lee2024koala,kim2024bksdm} reduce the number of layers by training a shallower student model, pruning methods~\cite{guo2025mosaicdiff, zhu2025obsdiff} lower model complexity through parameter removal, and quantization methods~\cite{li2023qdiffusion, liu2024edadm, he2023ptqd} lower numerical precision to reduce memory and computational cost. 
Orthogonal to the aforementioned approaches, token reduction methods~\cite{bolya2023tomesd, Kim_2024_WACV, zhang2025training, lu2025toma} directly reduce the token length, effectively mitigating the quadratic attention cost. However, existing methods do not explicitly minimize recovery error, which can lead to noticeable quality degradation.

\subsection{Token Reduction in ViTs}
Token reduction (TR) has been extensively explored to enhance the efficiency of Vision Transformers (ViTs). The core objective of these methods is to identify redundant tokens based on token similarity or importance scores, thereby reducing the token count via pruning or merging. 
Importance-based methods\cite{liang2022not, rao2021dynamicvit, ATS, meng2022adavit, pan2021iared2, kong2022spvit,Yin_2022_CVPR} reduce uninformative tokens using importance scores derived from a trainable prediction module or the attention maps in self-attention layers. In contrast, Similarity-based methods \cite{bolya2023tome, Kim_2024_WACV, yang2025kff, marin2023tokenpooling, visapp25} aggregate spatially redundant tokens by leveraging their token similarity. However, these methods are primarily tailored for discriminative tasks, which focus solely on redundancy removal. In contrast, since generative models such as Diffusion Models (DMs) require an additional recovery stage, these reduction-only methods cannot be directly applied.


\subsection{Token Reduction in DMs}
In DMs, tokens correspond to specific spatial positions within the final output; thus, the original token count must be preserved through an explicit recovery stage. Recently, several TR methods incorporating such recovery stages have been proposed for DMs~\cite{bolya2023tomesd, Kim_2024_WACV, zhang2025training, lu2025toma}. These approaches generally leverage input token similarity to identify redundant tokens and their corresponding reference tokens, forming matched token pairs to apply merging or pruning. For instance, ToMeSD~\cite{bolya2023tomesd} utilizes a bipartite soft matching algorithm to efficiently find and merge matched pairs, while ToFu \cite{Kim_2024_WACV} achieves higher performance by hybridizing merging and pruning strategies. Furthermore, SiTo~\cite{zhang2025training} calculates per-token similarity scores based on global token relationships to select effective reference tokens, and ToMA~\cite{lu2025toma} identifies optimal token pairs through a parallel greedy search within local regions.

However, these methods primarily focus on optimizing the token pair selection process by relying on input token similarity—a practice inherited from ViTs—which is sub-optimal from the perspective of minimizing recovery error. In contrast, DiTo explicitly minimizes recovery error by leveraging output token similarity to guide the TR process, thereby achieving superior generation quality compared to existing methods.

\section{Preliminaries}

\subsection{Diffusion Transformers}
Self-attention is the core computation in Diffusion Transformers (DiTs). For clarity, we present the essential operations of self-attention as follows:
\begin{equation} \label{eq:self_attn}
\mathrm{Attn}(X) \;=\; \mathrm{softmax}\!\left(QK^{\top}{}\right)V
\;=\; \mathrm{softmax}\!\left((XW_Q)(XW_K)^{\top}\right)(XW_V),
\end{equation}
Where $W_Q, W_K, W_V\in\mathbb{R}^{d\times d}$ represent the projection matrices that produce the query, key, and value, respectively, and $X\in\mathbb{R}^{N\times d}$ denotes the input tensor of attention. 
$N$ represents the token length and $d$ represents the hidden dimension. 
As shown in \cref{eq:self_attn}, the computational complexity scales quadratically with the token length, i.e., $O(N^2)$, leading to a rapid increase in computation and making it a major bottleneck in Transformers.
Accordingly, token reduction (TR) methods effectively accelerate DiTs by directly reducing the token length, thereby alleviating this bottleneck and lowering the overall computational cost.

\subsection{Token Reduction Pipeline: Formulation and Objectives}
\label{token_reduction_pipeline}
\textbf{Formulation.} We summarize the TR process in DMs through a unified pipeline that generalizes existing TR methods. The pipeline primarily consists of three distinct stages: token matching, reduction, and recovery. 
First, the token matching stage establishes token correspondences based on their similarity, providing the essential mapping information required for the subsequent reduction and recovery stages. 
This stage begins with token bipartitioning, where the total $N$ tokens are split into two disjoint sets: a destination (dst) set $\mathcal{D}$ and a source (src) set $\mathcal{S}$, such that $N = |\mathcal{D}| + |\mathcal{S}|$. Subsequently, we perform dst--src pair matching by constructing a similarity map $\mathbf{A}$ between these two sets. 
For each src token $s \in \mathcal{S}$, we determine the most similar dst token and its corresponding similarity score as follows:
\begin{equation}
d^{\star}(s) = \arg\max_{d \in \mathcal{D}} \mathbf{A}_{d,s}, \quad \hat{s}(s) = \max_{d \in \mathcal{D}} \mathbf{A}_{d,s},
\label{eq:matching_selection}
\end{equation}
where $d^{\star}(s)$ forms candidate token pairs and $\hat{s}(s)$ represents the candidate similarity scores. 
To determine the final matched token pairs, we select a subset of source tokens, $\mathcal{S}_K \subset \mathcal{S}$, consisting of the top-$k$ tokens with the largest candidate similarity scores, where $k$ is determined by a predefined reduction ratio. This results in the final set of matched pairs, $\mathcal{P} = \{(d^{\star}(s), s) \mid s \in \mathcal{S}_K\}$.
Finally, the source tokens $s \in \mathcal{S}_K$ defined in $\mathcal{P}$ are reduced to decrease the token count for efficient computation during the reduction stage. Following the core operations, these reduced tokens are restored during the recovery stage by copying the output of their corresponding $d^{\star}(s)$ as specified in $\mathcal{P}$, ultimately yielding the restored output $\tilde{\mathbf{Y}}$.

\vspace{0.5em}
\noindent
\textbf{Objectives.} Unlike reduction-only TR in ViTs, DMs must include a recovery stage to restore the original token count. While solely identifying redundant tokens is sufficient in ViTs, DMs require this restoration to achieve high-fidelity reconstruction. Consequently, the primary objective in DMs is to ensure that the restored output $\tilde{\mathbf{Y}}$ closely approximates the dense output $\mathbf{Y}$ computed without TR. This goal is formally expressed as minimizing the recovery error ($\mathcal{L}_{\text{out}}$):
\begin{equation}\label{eq:lout}
\min \mathcal{L}_{\text{out}} = \left\lVert Y - \tilde{Y} \right\rVert_F^2,
\end{equation}
where $\lVert \,\cdot\, \rVert_F^2$ denotes the squared Frobenius norm between $Y$ and $\tilde{Y}$.

To minimize the recovery error $\mathcal{L}_{\text{out}}$, it is essential to focus on the underlying mechanism of the recovery stage. Since this stage reconstructs reduced tokens by copying the outputs of their matched tokens, the resulting error is fundamentally determined by how closely the matched tokens' outputs represent the original outputs of the reduced ones. Therefore, to effectively minimize this error, it is imperative that token matching reflects the true similarities of $\mathbf{Y}$, ensuring that the reduction and recovery stages are aligned with the final reconstruction. However, existing methods directly inherit matching strategies from ViT-style TR, which rely on input token similarity. This leads to a fundamental misalignment, making these approaches sub-optimal for minimizing recovery error.
\section{Method}

\captionsetup{justification=centering}
\begin{figure}[tb]
  \centering
  \begin{subfigure}{0.50\linewidth}
    \centering
    \includegraphics[width=\linewidth,keepaspectratio]{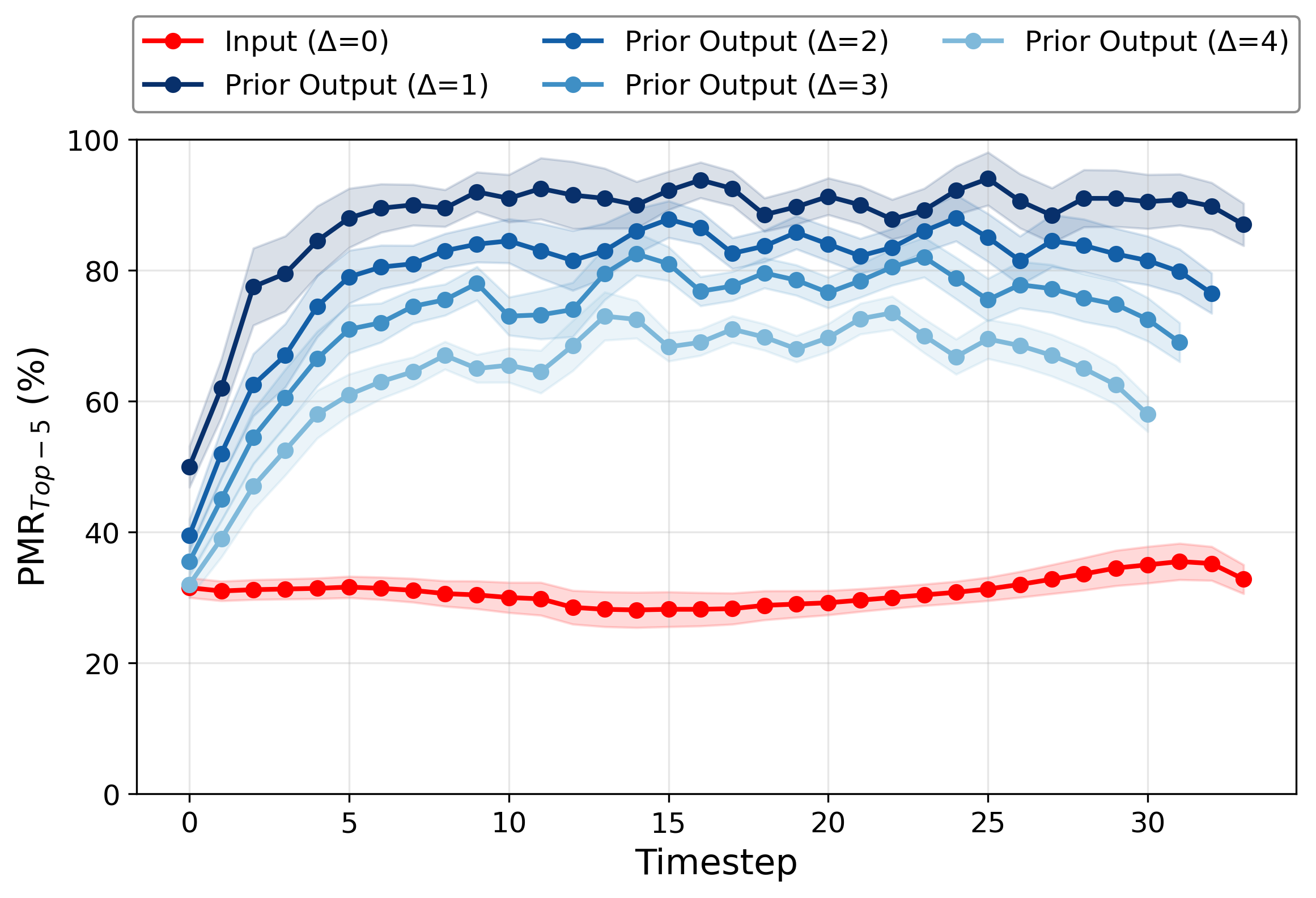}
    \caption{PMR Analysis across Intervals $\Delta$}
    \label{fig:pmr_input_output}
  \end{subfigure}
  \hfill
  \begin{subfigure}{0.48\linewidth}
    \centering
    \includegraphics[width=\linewidth,keepaspectratio]{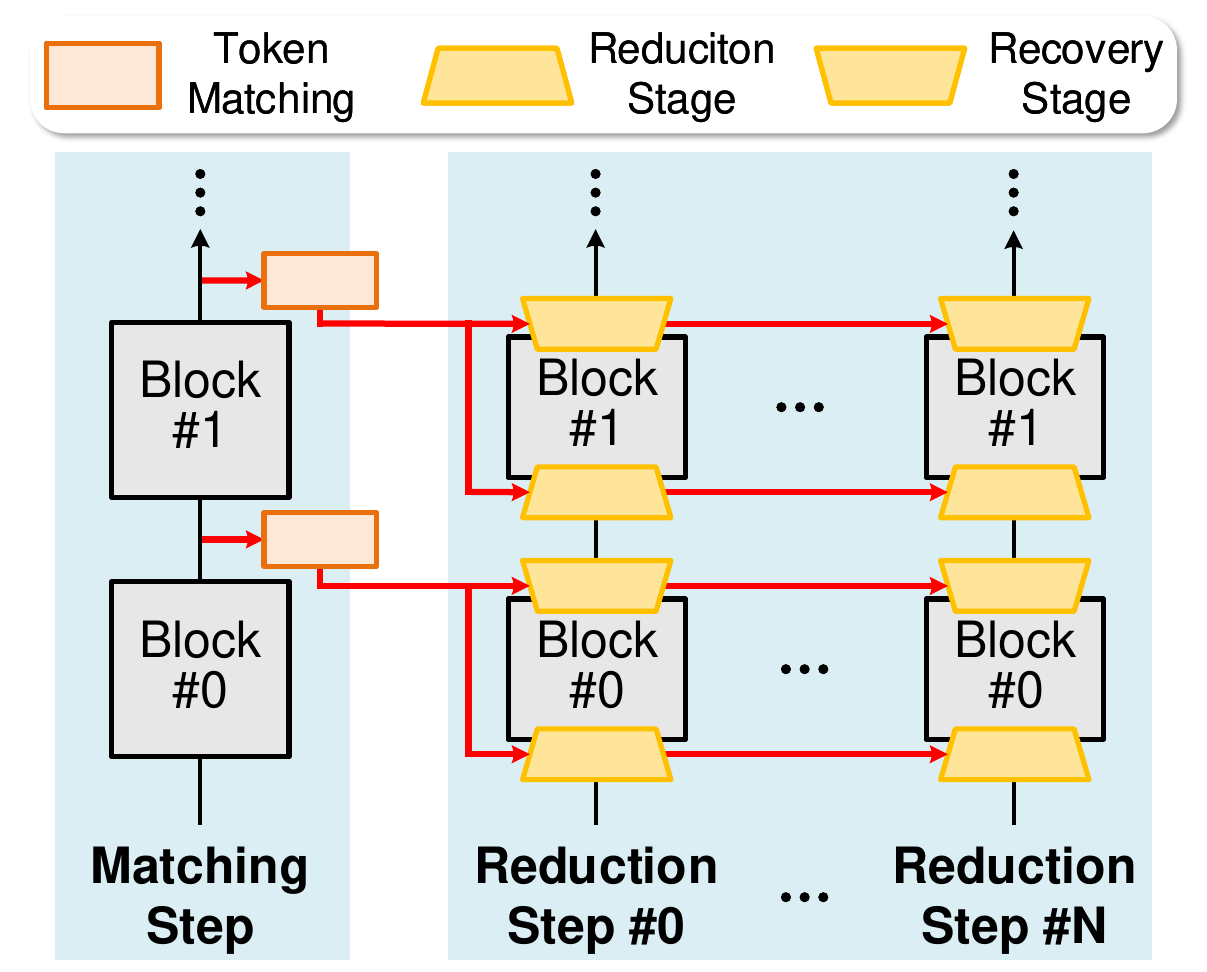}
    \caption{Overview of DiTo Pipeline}
    \label{fig:DiTo_pipeline}
  \end{subfigure}
  \captionsetup{justification=justified}
    \caption{\textbf{Quantitative analysis of token alignment and DiTo pipeline.} (a) PMR of prior outputs ($\Delta>=1$) remains consistently higher than that of the input ($\Delta=0$), despite a gradual decline as $\Delta$ increases. (b) Leveraging this observation, DiTo performs token matching using the prior output in a single Matching step and reuses the matching results across subsequent Reduction steps to maximize efficiency.}
  \label{fig1}
\vspace{-1.0em}
\end{figure}
\captionsetup{justification=justified}    
    
\subsection{Output-Similarity-Aware Token Reduction}
\textbf{Motivation.} To minimize the recovery error $\mathcal{L}_{\text{out}}$, the token matching must be performed using the output token similarity. However, since the dense output ${Y}$  is only available after completing the full computation, we cannot directly perform token matching using ${Y}$. Therefore, a practical approach requires finding a proxy that can predict the output token similarity as closely as possible. To address this, we exploit the temporal consistency inherent in DMs: since activations remain similar across adjacent timesteps, we assume that the output token similarity is likewise preserved between neighboring timesteps. 

To verify this hypothesis, we first define \textbf{Top-$k$ Pair Match Rate (PMR)} as a metric that quantifies the alignment with the golden token matching results derived from the current-step output ${Y}$. Specifically, PMR measures the proportion of source tokens whose predicted Top-$k$ destination set contains the single golden destination token. Given the competitive similarity scores among top candidates, this Top-$k$ criterion provides a reasonable margin for error compared to a strict 1:1 match. The PMR definition is given as follows:
\begin{equation}\label{eq:pmr_def}
\mathrm{PMR}_{\text{Top-}k}(t,b,\Delta)
= \frac{1}{|S|}\sum_{s\in S}\mathbb{I}
\!\left[\left|\,T^{(1)}_{t,b}(s)\ \cap\ T^{(k)}_{t-\Delta,b}(s)\,\right|>0\right].
\end{equation}
Here, $S$ is the set of source tokens $s$, while $t \in \{0, \dots, T-1\}$ and $b \in \{0, \dots, B-1\}$ denote the timestep and block index, respectively. $T^{(k)}_{t,b}(s)$ represents the set of Top-$k$ destination candidates for token $s$ at timestep $t$ and block $b$, and $\mathbb{I}[\cdot]$ is the indicator function.
The parameter $\Delta$ determines the comparison target for evaluating its alignment with the golden matching results:
\begin{itemize}
    \item[$\circ$] $\Delta = 0$: Matching results derived from the current-step input.
    \item[$\circ$] $\Delta \geq 1$: Matching results derived from the prior output at step $t-\Delta$.
\end{itemize}

Leveraging the PMR metric, \cref{fig:pmr_input_output} presents the timestep-wise average of $\mathrm{PMR}_{\text{Top-}5}$ calculated over 50 prompts. The figure compares the matching consistency of the current input ($\Delta=0$) and prior outputs ($\Delta \geq 1$) against the golden matching results, with the shaded regions visualizing the distribution across prompts.
As shown in the figure, the PMR values of prior outputs are significantly higher than those of the current input across most timesteps, validating our assumption that golden matching results are highly correlated with the results from prior steps rather than the current input.
Based on these observations, we propose \textbf{\textit{Output-Similarity-Aware Token Reduction}}, which performs token matching using the prior output to exploit the high temporal consistency of matching results across steps.

\vspace{0.5em}
\noindent
\textbf{Proposed DiTo Pipeline.} 
DiTo introduces a novel pipeline that decouples token matching from the reduction/recovery stage, which are originally handled within a single pipeline in conventional TR methods. Under this architecture, the timesteps of the diffusion process are categorized into two distinct types based on their roles. First, in the \textbf{Matching step}, the model performs full computation and utilizes the outputs for token matching to establish the token correspondences. Second, in the \textbf{Reduction step}, the model executes reduced computation by performing the reduction and recovery stages using the correspondences derived from the preceding Matching step. 
While DiTo maintains matching results throughout the Reduction steps, the associated memory overhead is negligible, as the stored information consists of compact index metadata rather than high-dimensional token features (Appendix).

Built upon this design, the overall pipeline of DiTo is illustrated in \cref{fig:DiTo_pipeline}. The process consists of a single Matching step followed by a series of multiple Reduction steps. The rationale for this interleaved structure is grounded in the analysis presented in \cref{fig:pmr_input_output}; our results demonstrate that the PMR remains robust even as the reuse interval increases, indicating that matching results maintain a valid correlation over several subsequent timesteps.
Consequently, rather than requiring a dedicated Matching step for every Reduction step, we leverage these high PMR values to reuse a single matching result across multiple subsequent Reduction steps. This strategy maximizes computational efficiency while preserving high generation quality.

\subsection{PMR-guided Interval Scheduling} 
\label{sec:intervla_scheduling}
While DiTo enhances efficiency through matching reuse, increasing the interval $\Delta$ leads to a decline in PMR, which can degrade generation quality. Furthermore, since PMR values fluctuate across different timesteps, determining the optimal reuse interval for each timestep is critical.
\begin{algorithm}[t]
\caption{\footnotesize Assign Matching/Reduction Steps from Max Intervals ($\Delta_t^{\max}$)}
\label{alg:select_steps_lookahead}
\KwIn{Max intervals $\{\Delta_t^{\max}\}_{t=0}^{T-1}$}
\KwOut{Lists $\mathcal{T}_{\mathrm{match}}$ and $\mathcal{T}_{\mathrm{reduce}}$}

\BlankLine
$\mathcal{T}_{\mathrm{match}} \leftarrow [\,]$\ \tcp*{list of Matching steps}
$\mathcal{T}_{\mathrm{reduce}} \leftarrow [\,]$\ \tcp*{list of Reduction steps}
$m \leftarrow 0$                                \tcp*{Index of latest Matching step}

\For{$t \leftarrow 0$ \KwTo $T-1$}{
    \eIf{$t = 0$}{
        \tcp{\footnotesize Initialize for the timestep 0}
        Append $0$ to $\mathcal{T}_{\mathrm{match}}$\;
        $m \leftarrow 0$\  
    }{
        $\Delta_{t+1} = (t+1) - m$\ \tcp*{interval at $t+1$}
        \eIf{$\Delta_{t+1} > \Delta_{t+1}^{\max}$}{
            Append $t$ to $\mathcal{T}_{\mathrm{match}}$\;
            $m \leftarrow t$\;
        }{
            Append $t$ to $\mathcal{T}_{\mathrm{reduce}}$\;
        }
    }
}

\tcp{\footnotesize Keep only the last timestep in consecutive Matching steps}
$\mathcal{T}_{\mathrm{match}} \leftarrow CollapseConsecutive(\mathcal{T}_{\mathrm{match}})$\;

\Return{$\mathcal{T}_{\mathrm{match}},\ \mathcal{T}_{\mathrm{reduce}}$}\;
\end{algorithm}

To address this, we propose \textbf{\textit{PMR-guided Interval Scheduling}}, which utilizes the PMR metric as a guideline for determining the optimal matching schedule in an offline manner. The primary objective of this strategy is to maintain the PMR values above a predefined threshold $\tau$ throughout the generation process while maximizing efficiency. To achieve this, the transition between Matching and Reduction steps is dynamically determined by identifying the maximum interval $\Delta_t^{\max}$ at each timestep that satisfies the PMR criterion. Specifically, we first define the timestep-wise average PMR ($\overline{\mathrm{PMR}}$) as
\begin{equation}\label{eq:pmr_block_avg}
\overline{\mathrm{PMR}}_{\text{Top-}k}(t,\Delta)
\;=\;
\frac{1}{B}\sum_{b=0}^{B-1}\mathrm{PMR}_{\text{Top-}k}(t,b,\Delta),
\end{equation}
and then construct the valid interval set at each timestep $t$
\begin{equation}\label{eq:Dt_def}
\mathcal{D}_t
\;=\;
\Big\{
\Delta \in \{1,\dots,T-1\}
\ \big|\ 
\overline{\mathrm{PMR}}_{\text{Top-}k}(t,\Delta) \ge \tau
\Big\}.
\end{equation}
The maximum allowable interval ($\Delta_t^{\max}$) at each timestep $t$ is chosen as
\begin{equation}\label{eq:delta_t_max}
\Delta_t^{\max}
\;=\;
\begin{cases}
\max \mathcal{D}_t, & \text{if } \mathcal{D}_t \neq \emptyset,\\
0, & \text{otherwise.}
\end{cases}
\end{equation}

Given the max intervals $\Delta_t^{\max}$, \cref{alg:select_steps_lookahead} assigns each timestep to either a Matching step or a Reduction step. The core scheduling rule follows a one-step look-ahead strategy: to determine whether timestep $t$ should be a Matching or Reduction step, we evaluate the condition for the subsequent timestep $t+1$. Specifically, we calculate the expected interval at the next step as $\Delta_{t+1} = (t+1) - m$, where $m$ denotes the index of the latest Matching Step. This value is then compared against the allowable limit $\Delta_{t+1}^{\max}$ to determine whether a new Matching step is required. If $\Delta_{t+1}$ exceeds the $\Delta_{t+1}^{\max}$, timestep $t$ is designated as a Matching step to refresh the $m$; otherwise, it is assigned as a Reduction step. Finally, to minimize redundant token matching operations, if Matching steps occur consecutively, we retain only the final timestep within the sequence; this is because matching results from preceding steps are overwritten by the subsequent update, making earlier computations unnecessary.

\begin{figure}[t]
    \centering
    \includegraphics[width=\linewidth,keepaspectratio]{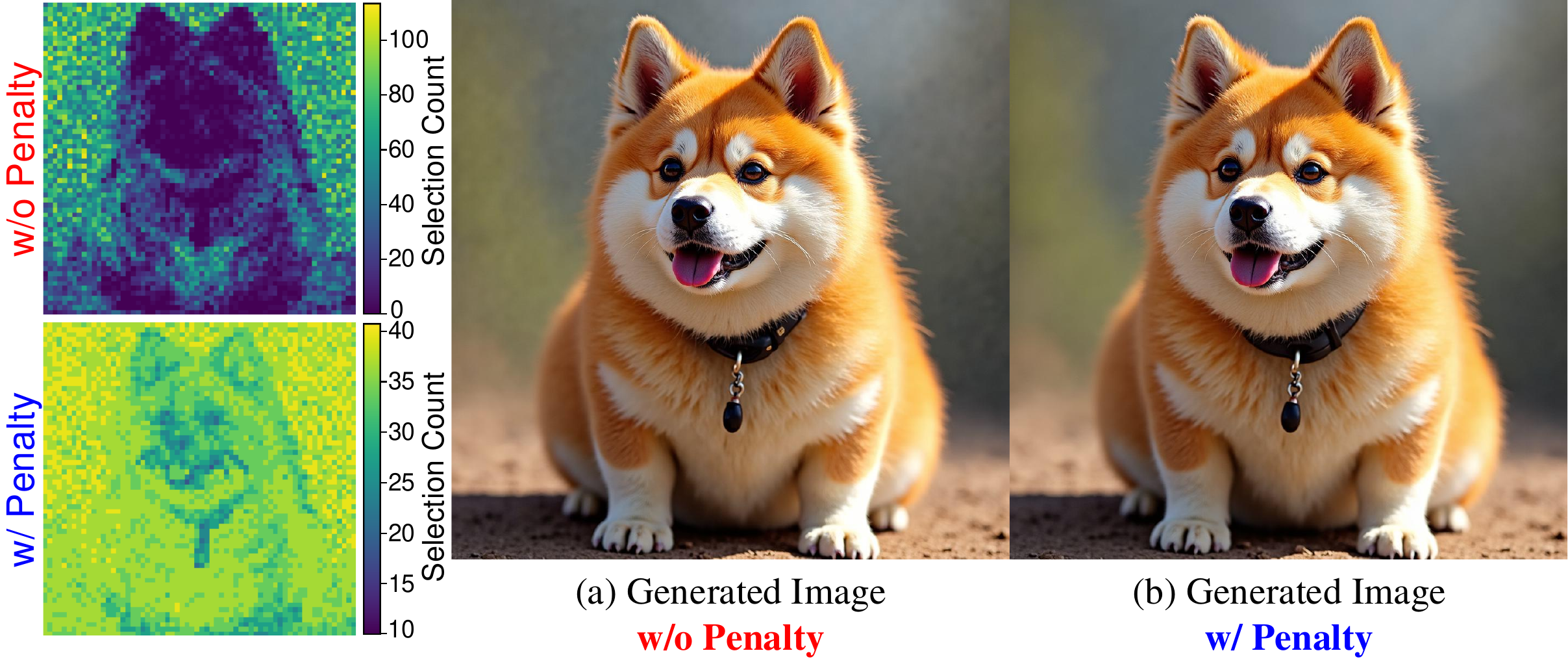}
    \caption{\textbf{Impact of token selection frequency in TR on image quality.} (a) In naïve DiTo, repeated reuse of matching results causes reduction targets to concentrate at specific spatial locations (top-left heatmap), resulting in visible blocking artifacts. (b) With our frequency-based penalty, the imbalance in selection frequency is alleviated (bottom-left heatmap). This reduces the accumulation of localized errors and successfully mitigates blocking artifacts while maintaining the token reuse efficiency.}
    
    \label{fig:penalty}
\vspace{-1.0em}
\end{figure}

\subsection{Frequency-aware Token Matching}
\textbf{Motivation.} To improve efficiency, DiTo reuses the matching results in a single Matching step across multiple Reduction steps, reducing the frequency of token matching operations. However, as shown in the top-left heatmap of \cref{fig:penalty}, this repeated reuse amplifies the imbalance in token selection frequency for reduction. Specifically, when a matching result is reused over $d$ timesteps, the disparity in selection counts between frequently and rarely selected tokens expands in proportion to $d$. This causes reduction targets to persistently concentrate on a small subset of tokens, leading to the accumulation of localized approximation errors. These errors manifest as visible blocking artifacts, which, as illustrated in \cref{fig:penalty}a, emerge at the exact spatial locations corresponding to the high-intensity regions in the heatmap.

To mitigate this issue, we propose \textit{\textbf{Frequency-aware Token Matching}}, which augments the original token matching stage with a frequency-based penalty to alleviate the imbalance in the selection frequency. By penalizing tokens that have been frequently selected as reduction targets, this method prevents excessive error accumulation at specific spatial locations. Consequently, it effectively mitigates blocking artifacts while maintaining the computational efficiency of token reuse.

\vspace{0.5em}
\noindent
\textbf{Formulation.}
To ensure a scale-invariant penalty for every matching, we normalize the penalty magnitude using the similarity interval $\Delta s = \max(\hat{\mathbf{s}}) - \min(\hat{\mathbf{s}})$ from \cref{eq:matching_selection}. This accounts for the fact that the absolute scale of similarity scores varies across different matching instances, providing a consistent reference range based on the distribution of candidate scores $\hat{\mathbf{s}}$.
Furthermore, considering that only a fraction $r$ of tokens are selected during the matching stage, we define the baseline penalty scale as $r \cdot \Delta s$. 
The final penalty strength is controlled by a hyperparameter $\lambda$.
We maintain a token selection history vector $\mathbf{C}\in\mathbb{R}^{N}$ over all $N$ tokens, where $C_i$ stores the cumulative number of times token $i$ has been selected as a reduction target.
Since the source token indices vary across token matching operations, we gather the history values for the current source set $\mathcal{S}$ as $\mathbf{C}[\mathcal{S}]\in\mathbb{R}^{S}$.
We then augment the original token matching process by subtracting a frequency-based penalty $\mathbf{p}$ from the candidate similarity scores $\hat{\mathbf{s}}$:
\begin{equation}\label{eq:freq_aware_matching_vec}
\mathbf{p} \;=\; \lambda\, \cdot r\, \cdot \Delta s \cdot \mathbf{C}[\mathcal{S}],
\qquad
\hat{\mathbf{s}}^{\mathrm{pen}} \;=\; \hat{\mathbf{s}} - \mathbf{p}.
\end{equation}

As a result, as shown in the bottom-left heatmap of \cref{fig:penalty}, the proposed penalty alleviates the imbalance in selection frequency. By reducing the accumulation of localized errors, it effectively mitigates blocking artifacts, as illustrated in \cref{fig:penalty}b. Additional visualizations demonstrating this effect are provided in the Appendix.

\section{Experiments}

\subsection{Experimental Settings}
\textbf{Models and Benchmarks.}
We evaluate DiTo on the text-to-image task using two representative DiT models: \texttt{FLUX.1-dev} (\texttt{Flux})\cite{flux2024} with 35 sampling steps and \texttt{stable-diffusion-3-medium} (\texttt{SD3})\cite{esser2024scaling} with 50 sampling steps. All experiments are conducted with the \texttt{Diffusers} framework \cite{von-platen-etal-2022-diffusers}, generating images at 1024$\times$1024 resolution for evaluation, and FlashAttention2\cite{dao2023flashattention2} is applied to all models for efficient attention computation. We use the 1,000 class names from the ImageNet-1k dataset \cite{imagenet15russakovsky} as prompts, and generate a total of 3,000 images using three random seeds for the main experiments. For the ablation study, we randomly sample 100 images per seed, resulting in 300 images in total. Latency is measured on an RTX 6000 Ada GPU\cite{nvidia_rtx6000_ada}, averaged over 100 generated images.

\textbf{Evaluation Metrics.}
We use FID\cite{FID}, CLIP\cite{CLIPScore}, PSNR\cite{psnr2008}, SSIM\cite{wang2004ssim}, and LPIPS\cite{LPIPS} as evaluation metrics. FID (Fréchet Inception Distance) measures the distance between the distributions of generated and real images in the Inception feature space, capturing overall distributional quality and diversity. We compute FID using the ImageNet-1k ground-truth images as the reference set. CLIP score measures text–image alignment based on the similarity between generated image and text embeddings. In addition, we evaluate perceptual reconstruction fidelity using PSNR, SSIM, and LPIPS. PSNR (Peak Signal-to-Noise Ratio) quantifies pixel-level reconstruction error, SSIM (Structural Similarity) evaluates structural similarity, and LPIPS measures perceptual differences in a pretrained deep feature space. For these three metrics, we use the generated images from the vanilla model as the reference. 
\begin{figure}[t]
\centering
\setlength{\tabcolsep}{2pt}
\renewcommand{\arraystretch}{0.0}

\newcommand{\headh}{6ex} 
\newcommand{\headw}{0.12\linewidth} 

\begin{tabular}{c @{\hspace{0.5mm}} c@{}c@{}c @{\hspace{3mm}} r @{\hspace{0.5mm}} c@{}c@{}c}
\parbox[c][\headh][c]{\headw}{\centering \texttt{Flux} \\ (Baseline)} &
\parbox[c][\headh][c]{\headw}{\centering DiTo \\ 10\%} &
\parbox[c][\headh][c]{\headw}{\centering DiTo \\ 30\%} &
\parbox[c][\headh][c]{\headw}{\centering DiTo \\ 50\%} &
\parbox[c][\headh][c]{\headw}{\centering \texttt{SD3} \\ (Baseline)} &
\parbox[c][\headh][c]{\headw}{\centering DiTo \\ 10\%} &
\parbox[c][\headh][c]{\headw}{\centering DiTo \\ 30\%} &
\parbox[c][\headh][c]{\headw}{\centering DiTo \\ 50\%} \\

\includegraphics[width=0.12\linewidth]{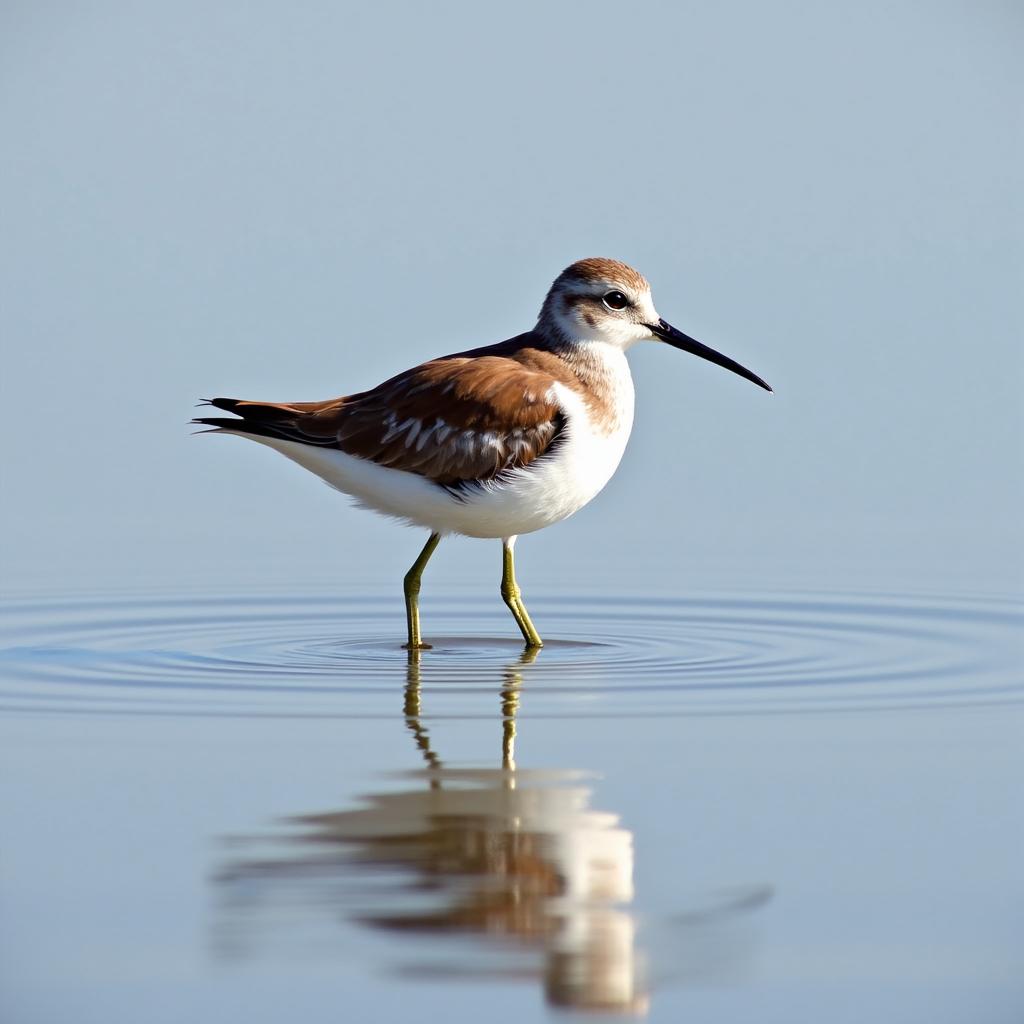} &
\includegraphics[width=0.12\linewidth]{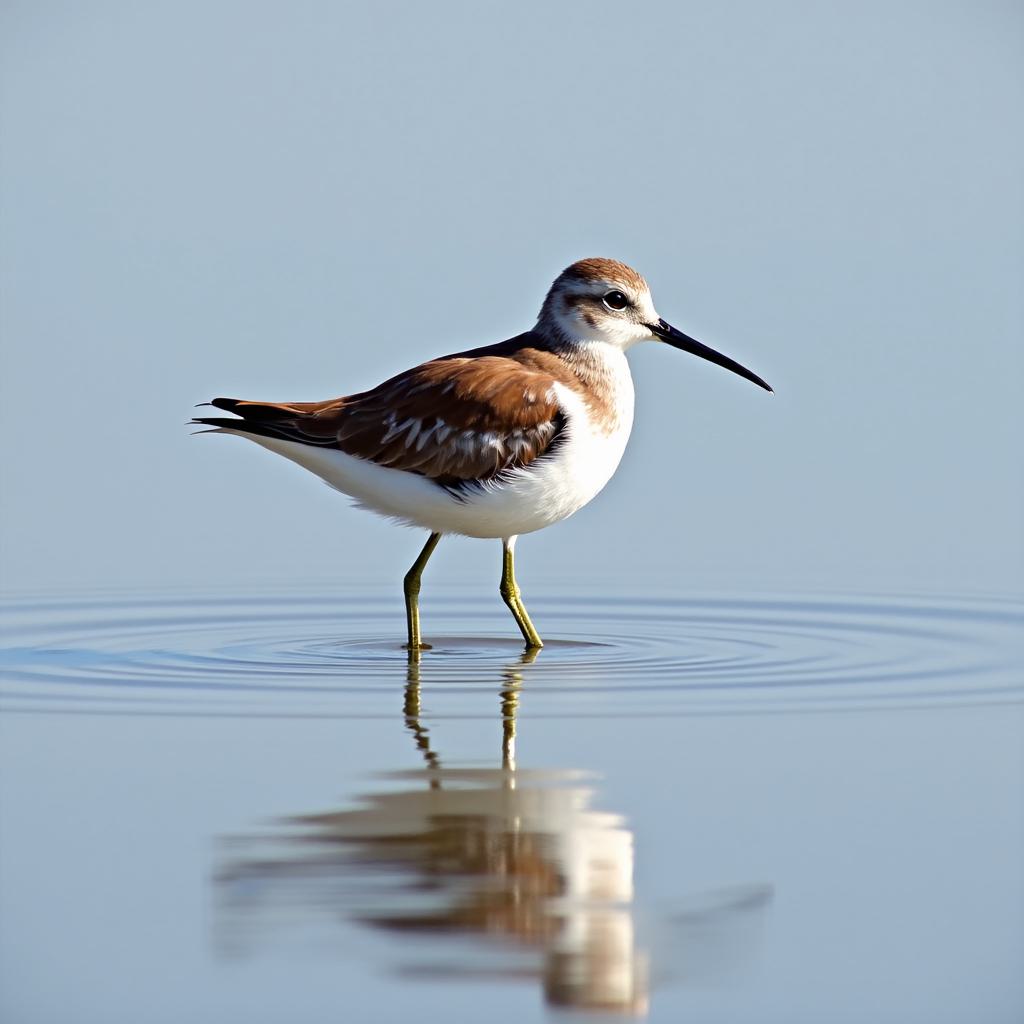} &
\includegraphics[width=0.12\linewidth]{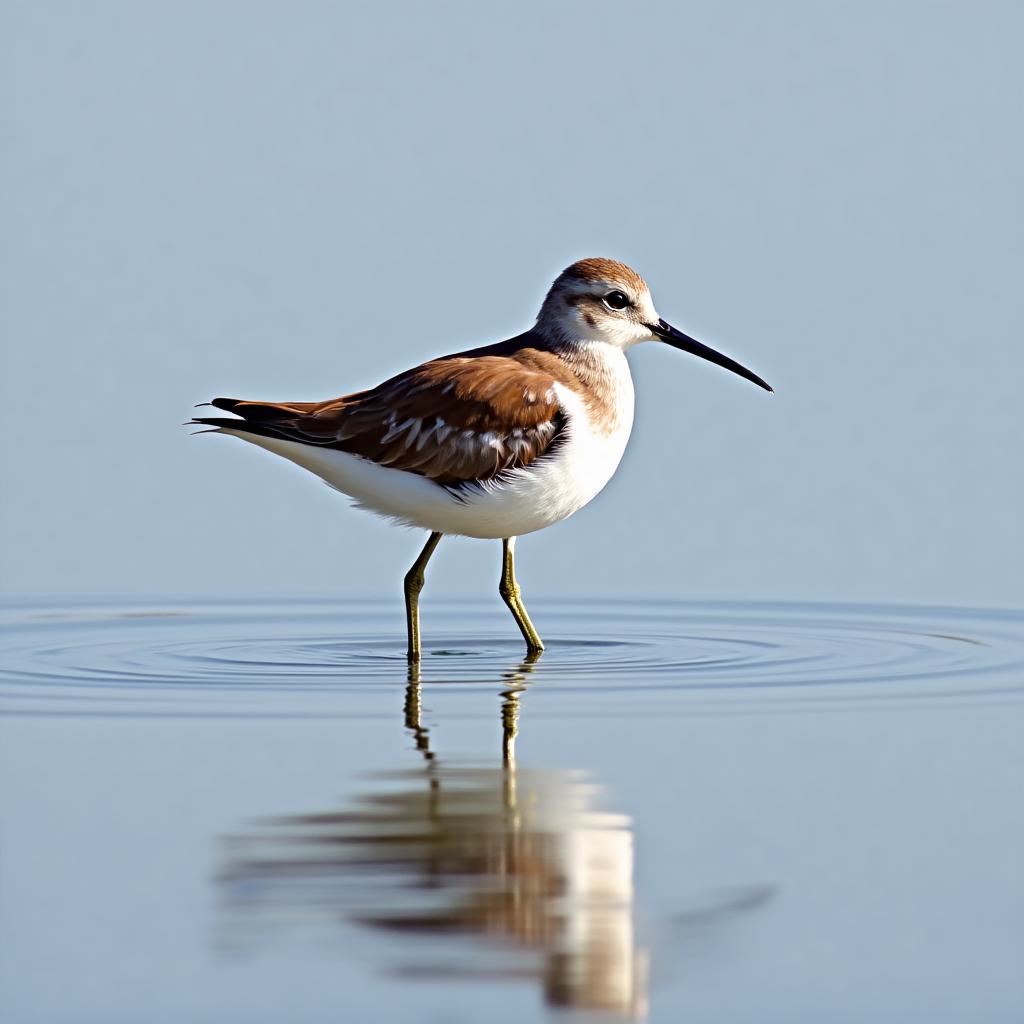} &
\includegraphics[width=0.12\linewidth]{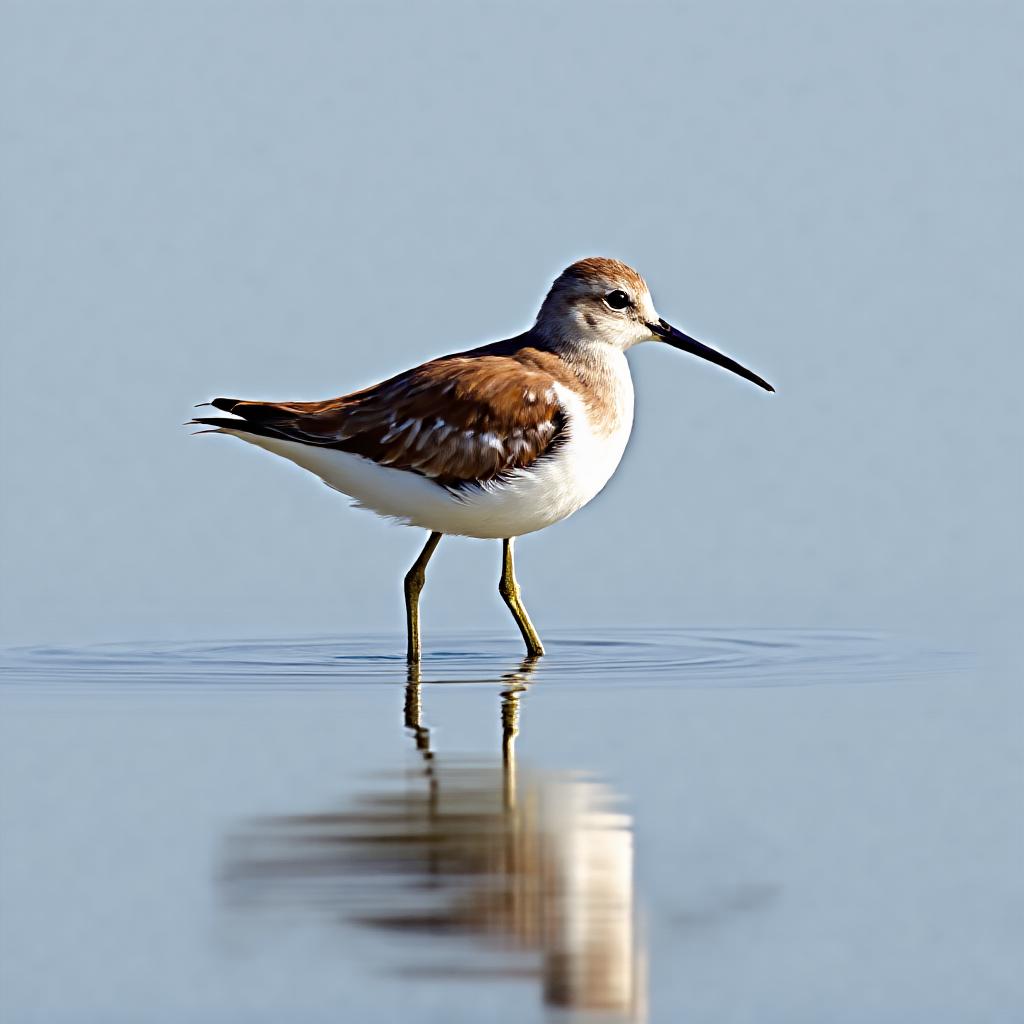} &

\includegraphics[width=0.12\linewidth]{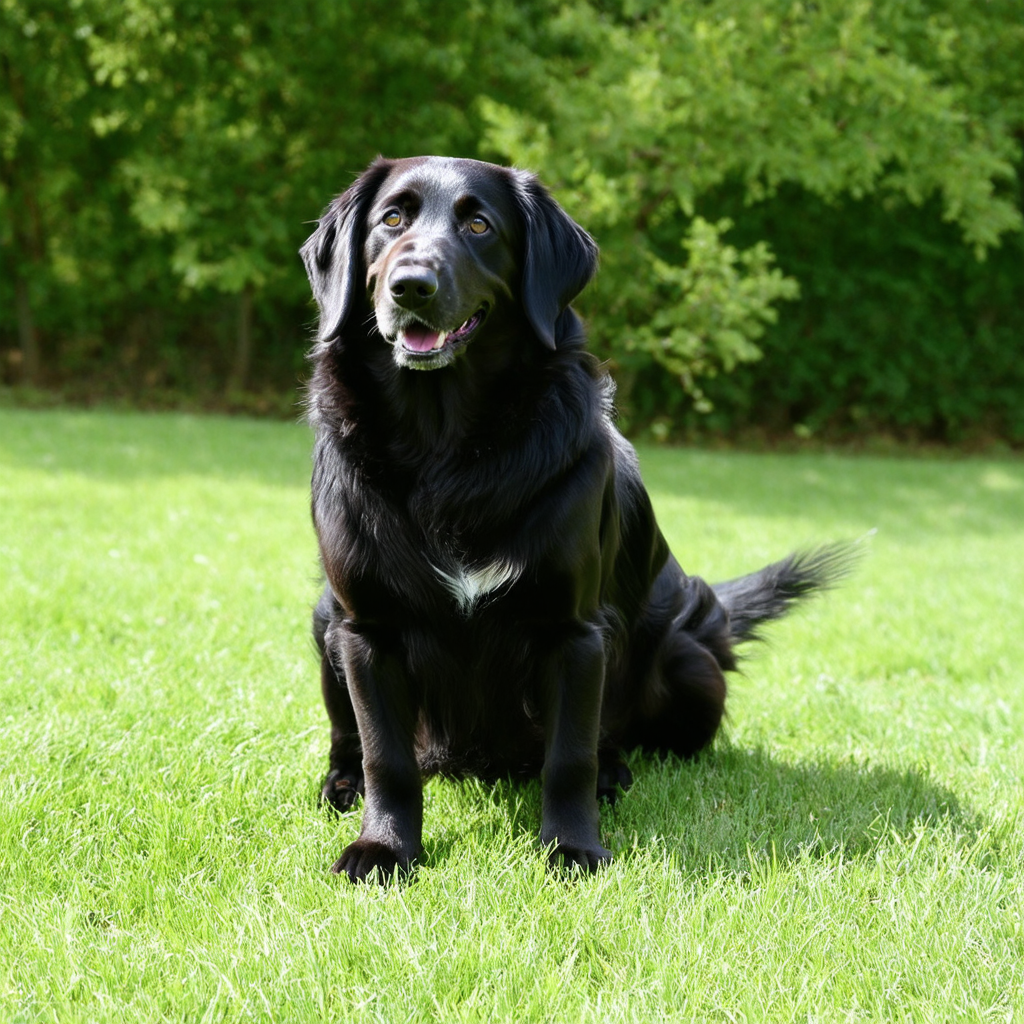} &
\includegraphics[width=0.12\linewidth]{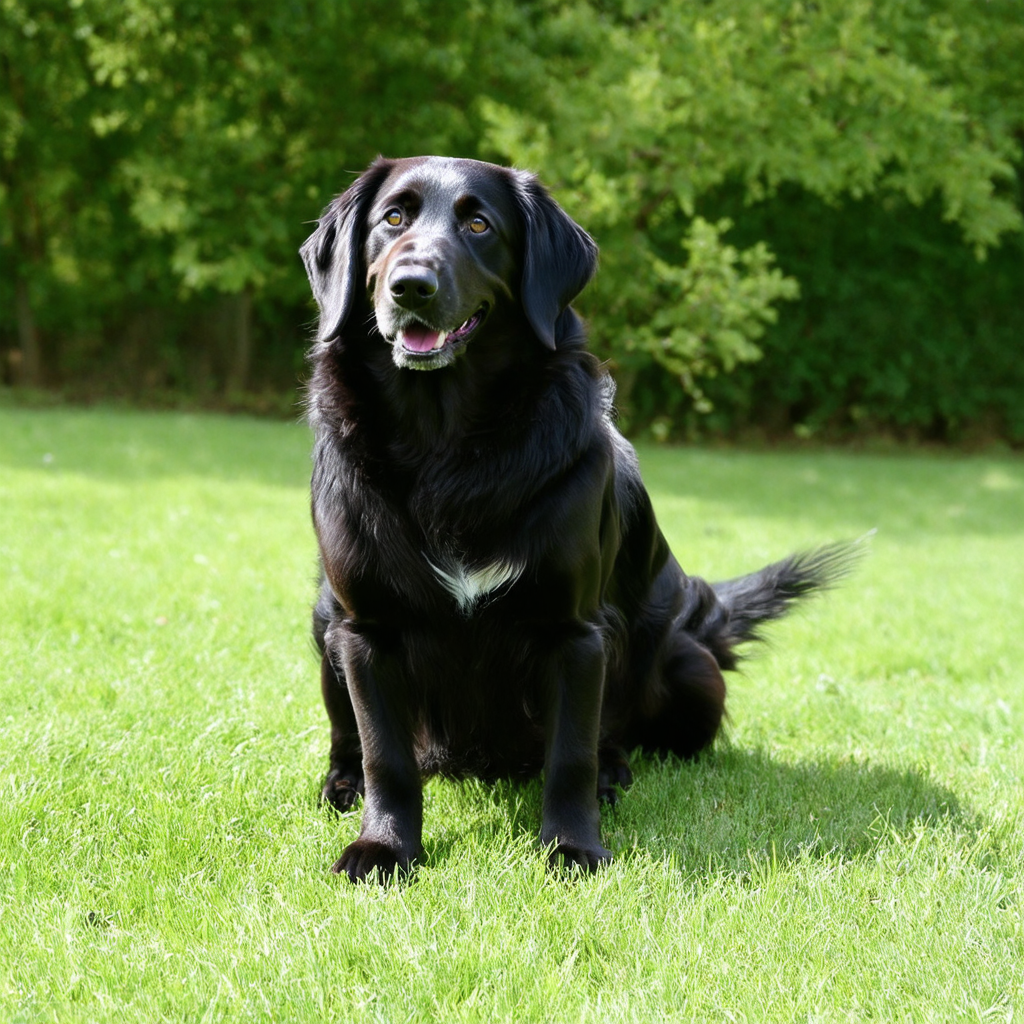} &
\includegraphics[width=0.12\linewidth]{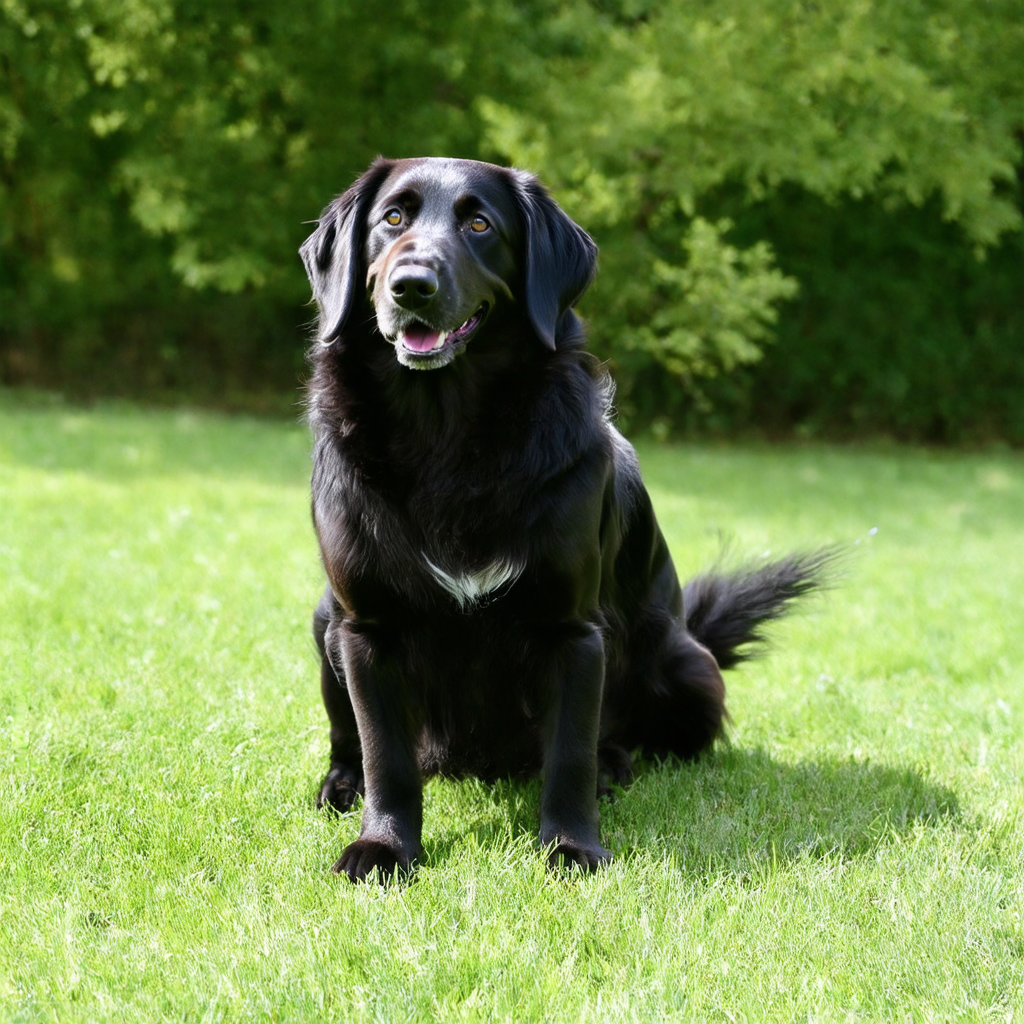} &
\includegraphics[width=0.12\linewidth]{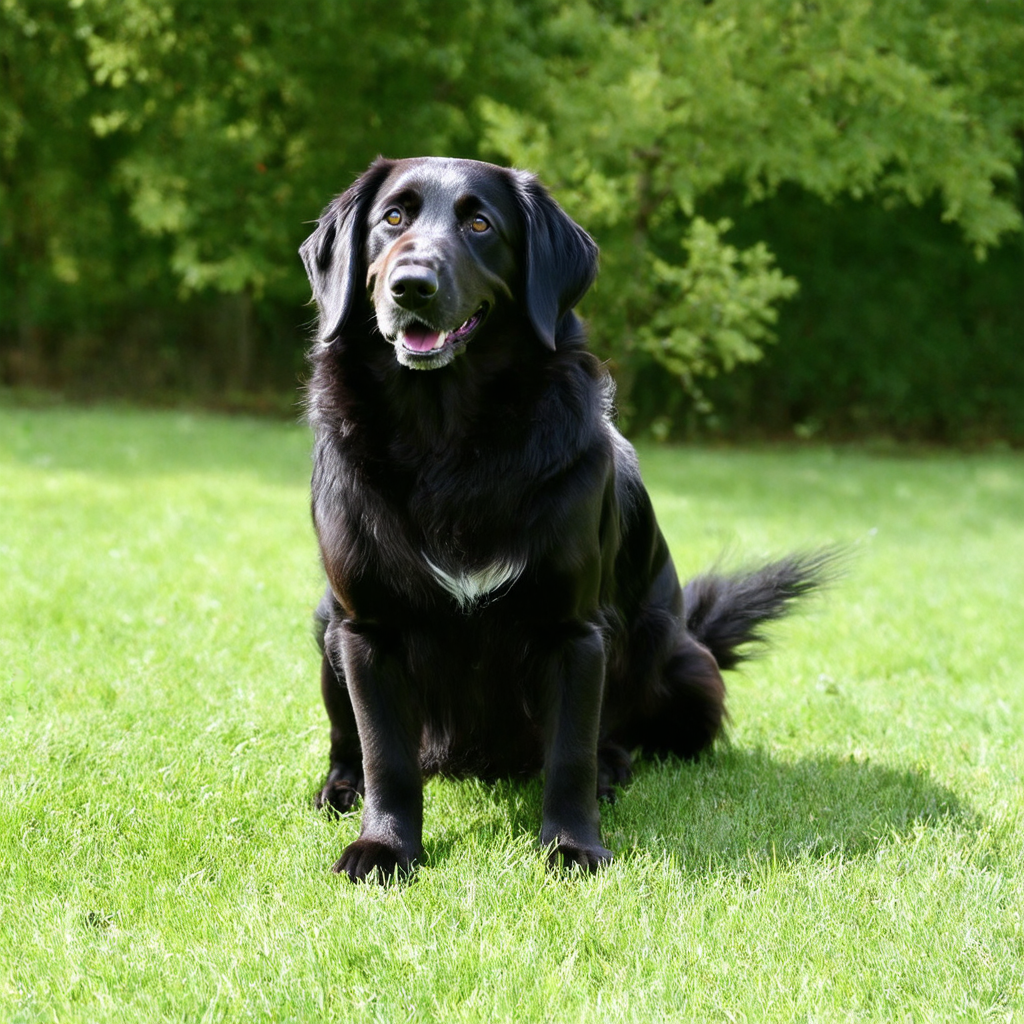} \\

\includegraphics[width=0.12\linewidth]{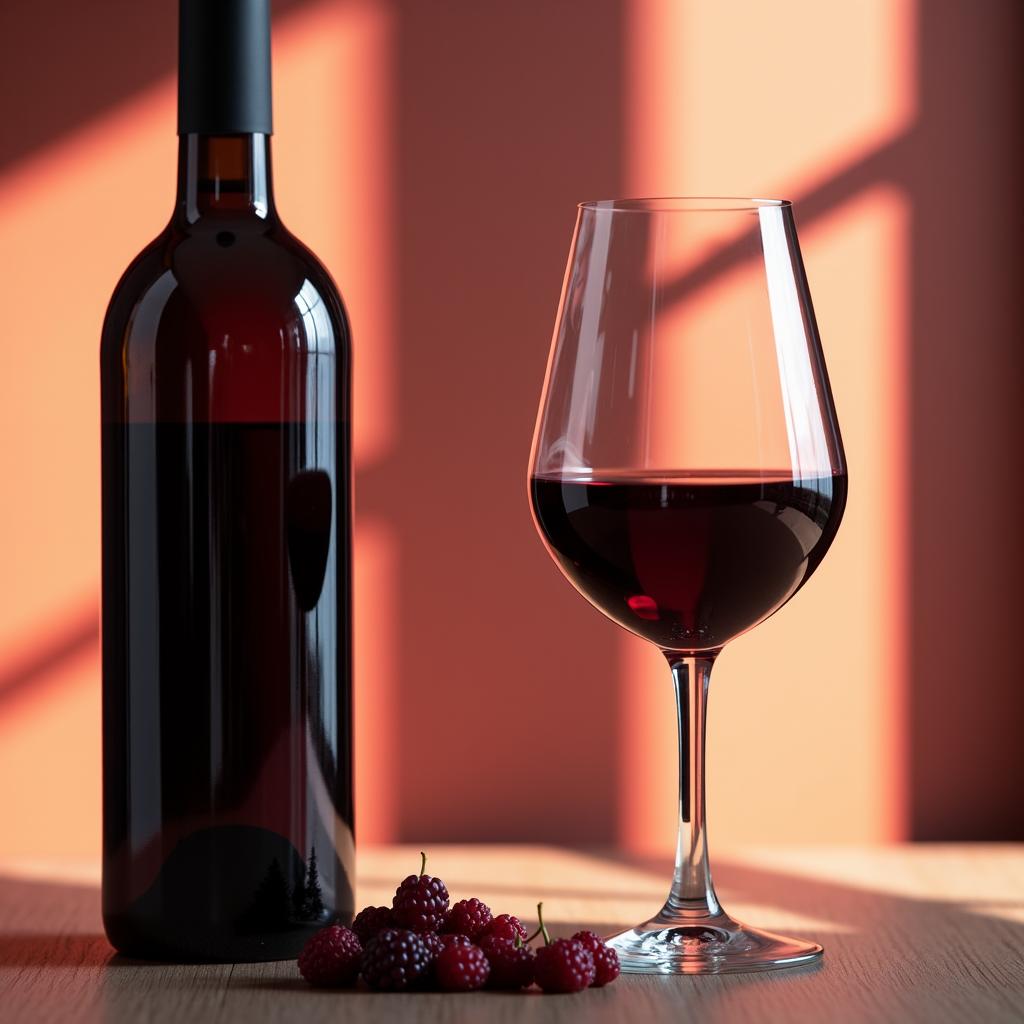} &
\includegraphics[width=0.12\linewidth]{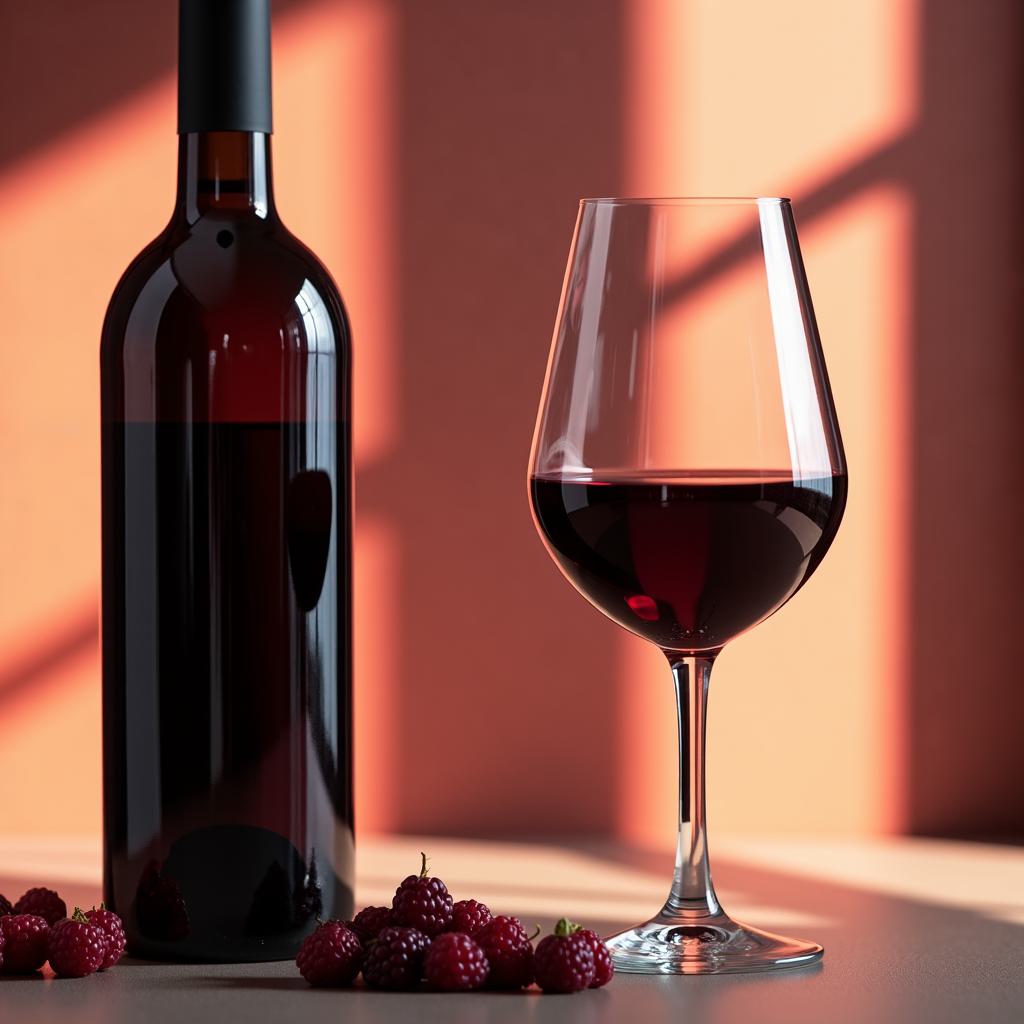} &
\includegraphics[width=0.12\linewidth]{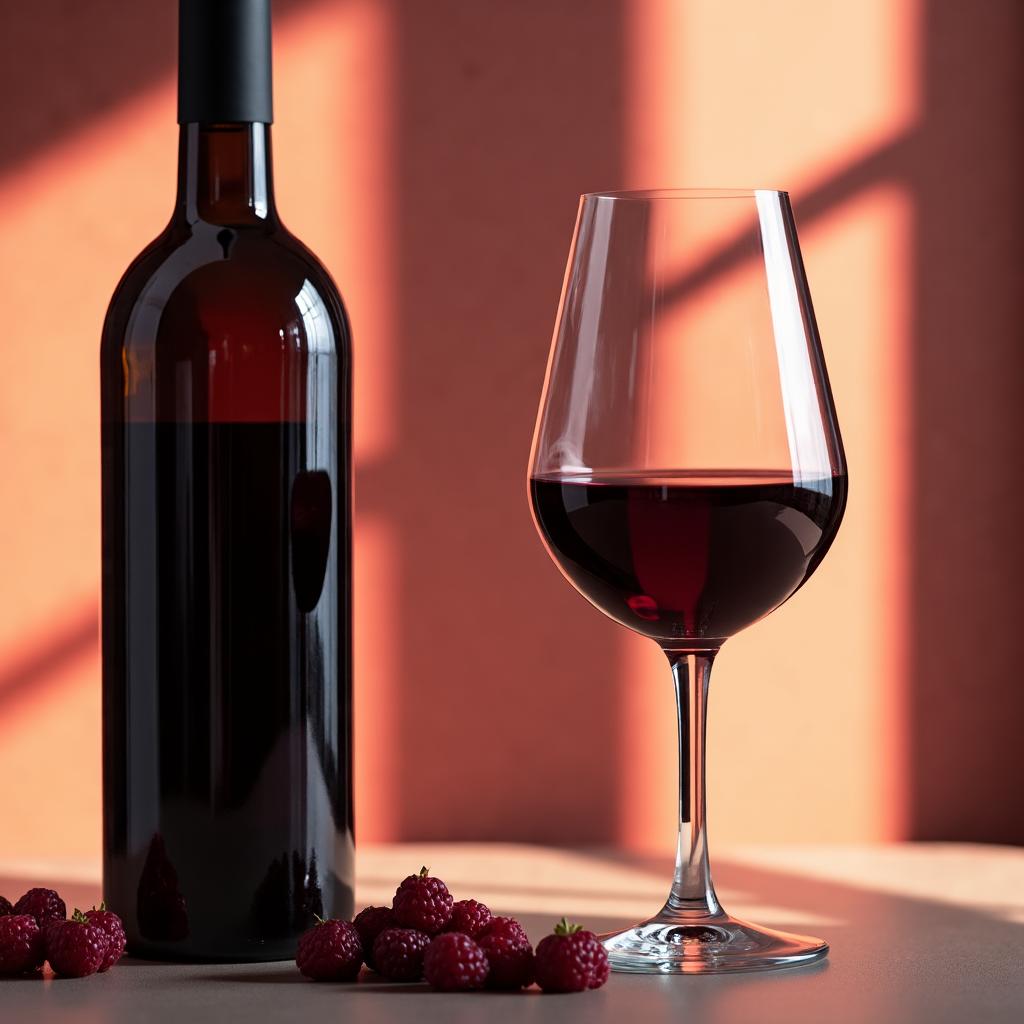} &
\includegraphics[width=0.12\linewidth]{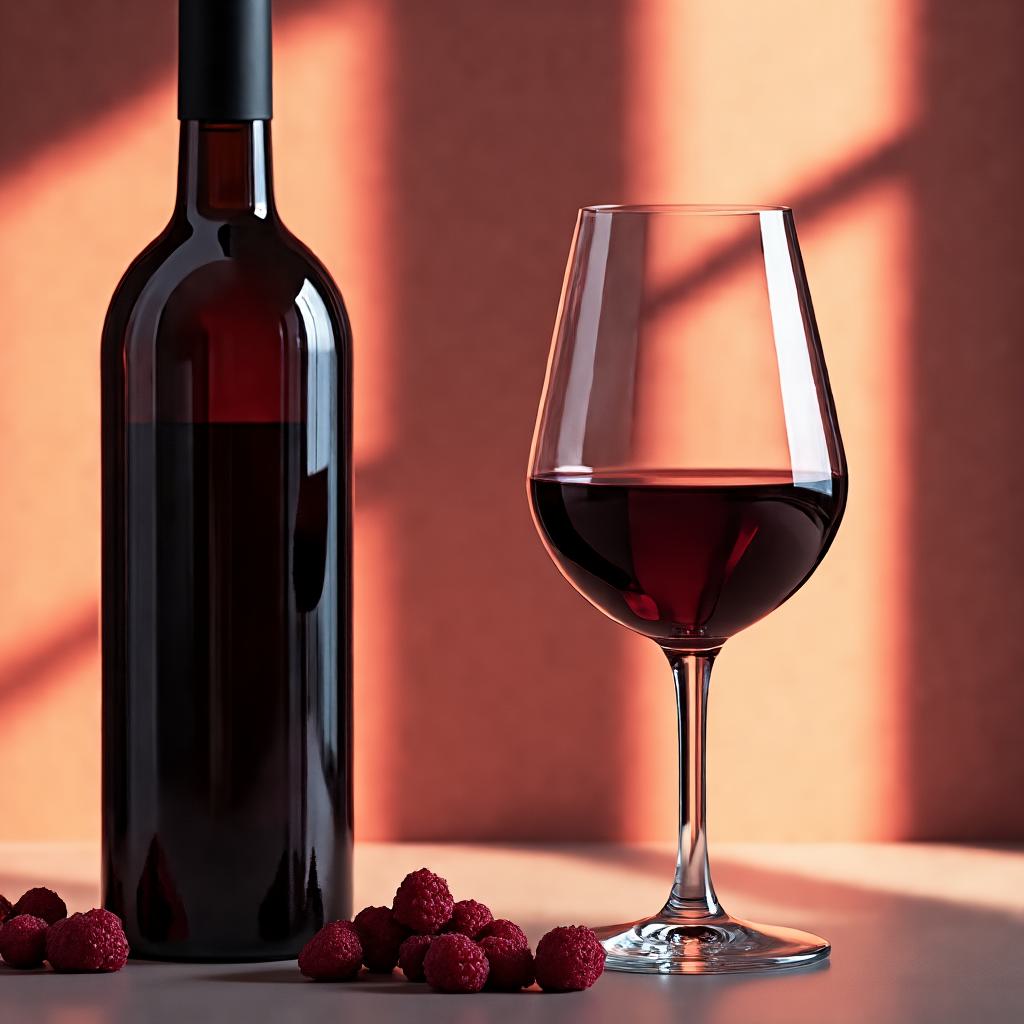} &

\includegraphics[width=0.12\linewidth]{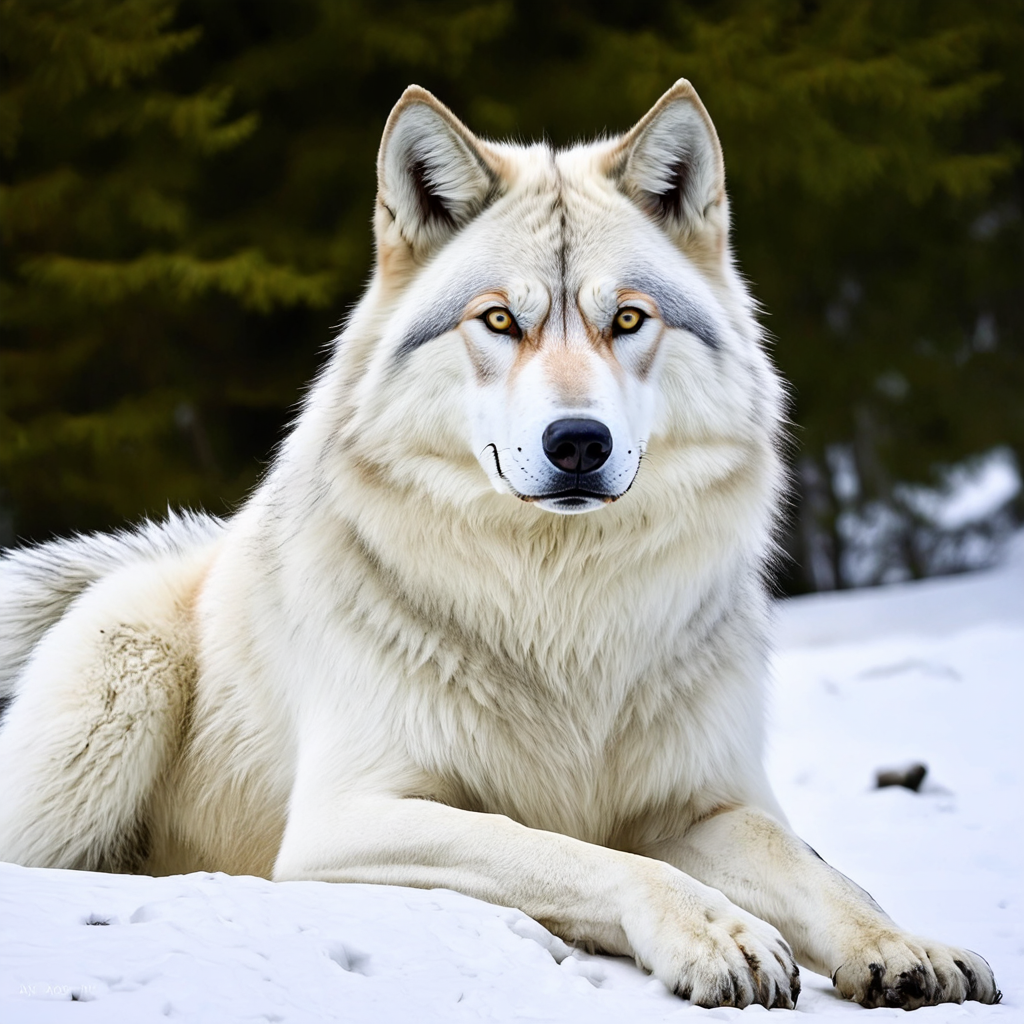} &
\includegraphics[width=0.12\linewidth]{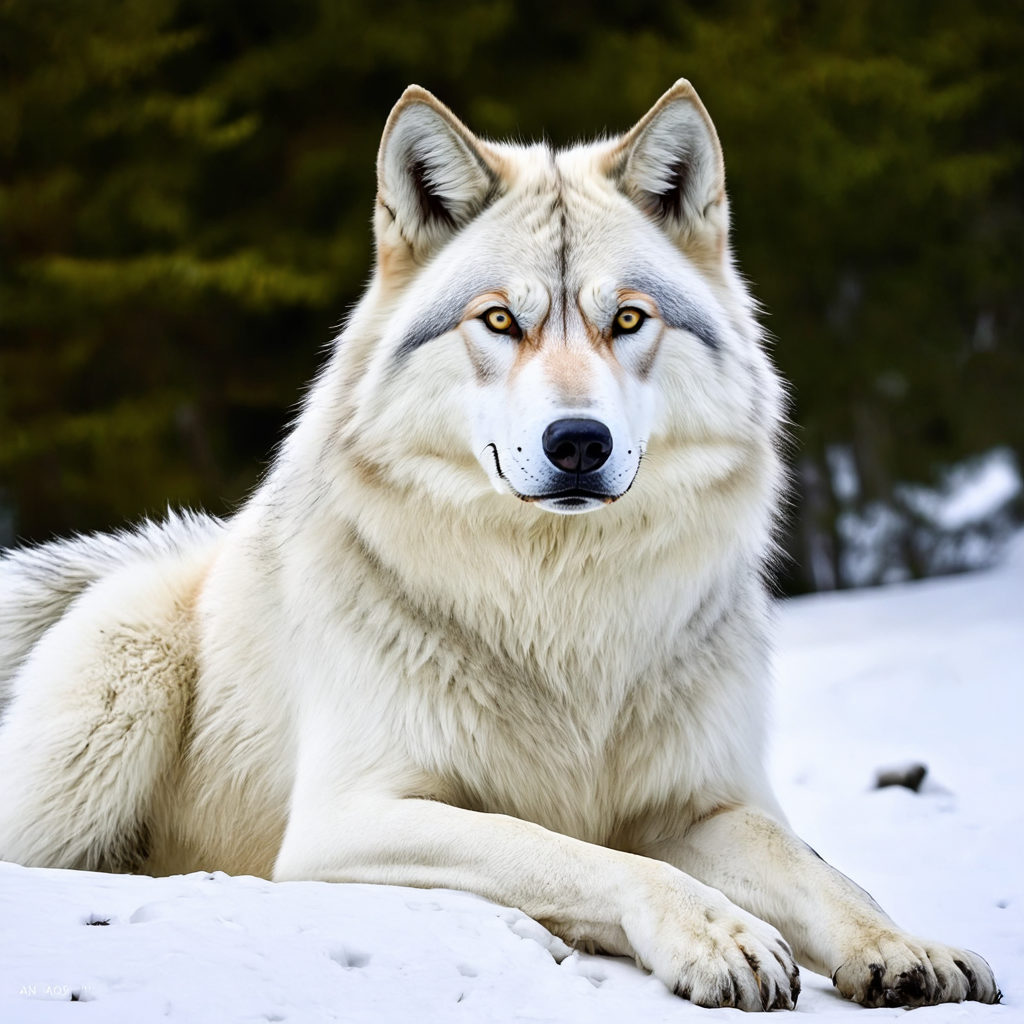} &
\includegraphics[width=0.12\linewidth]{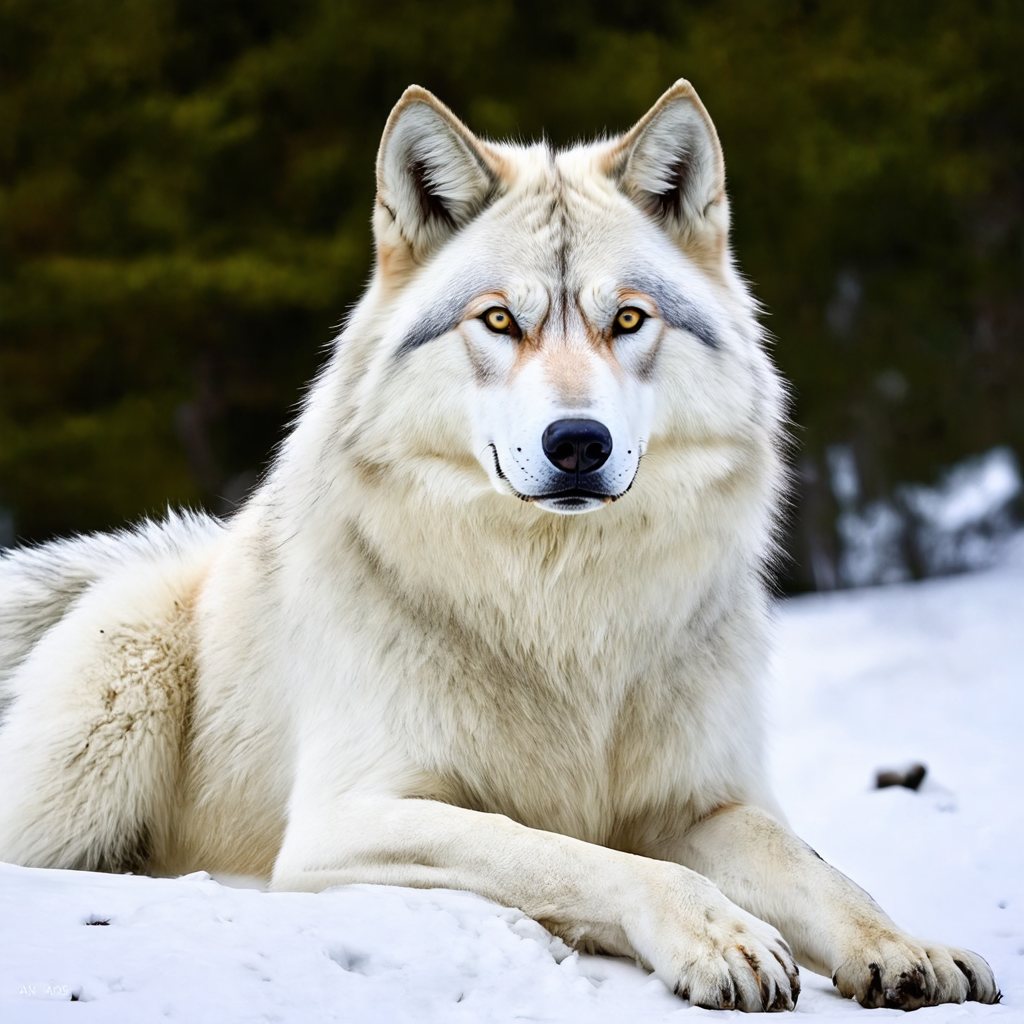} &
\includegraphics[width=0.12\linewidth]{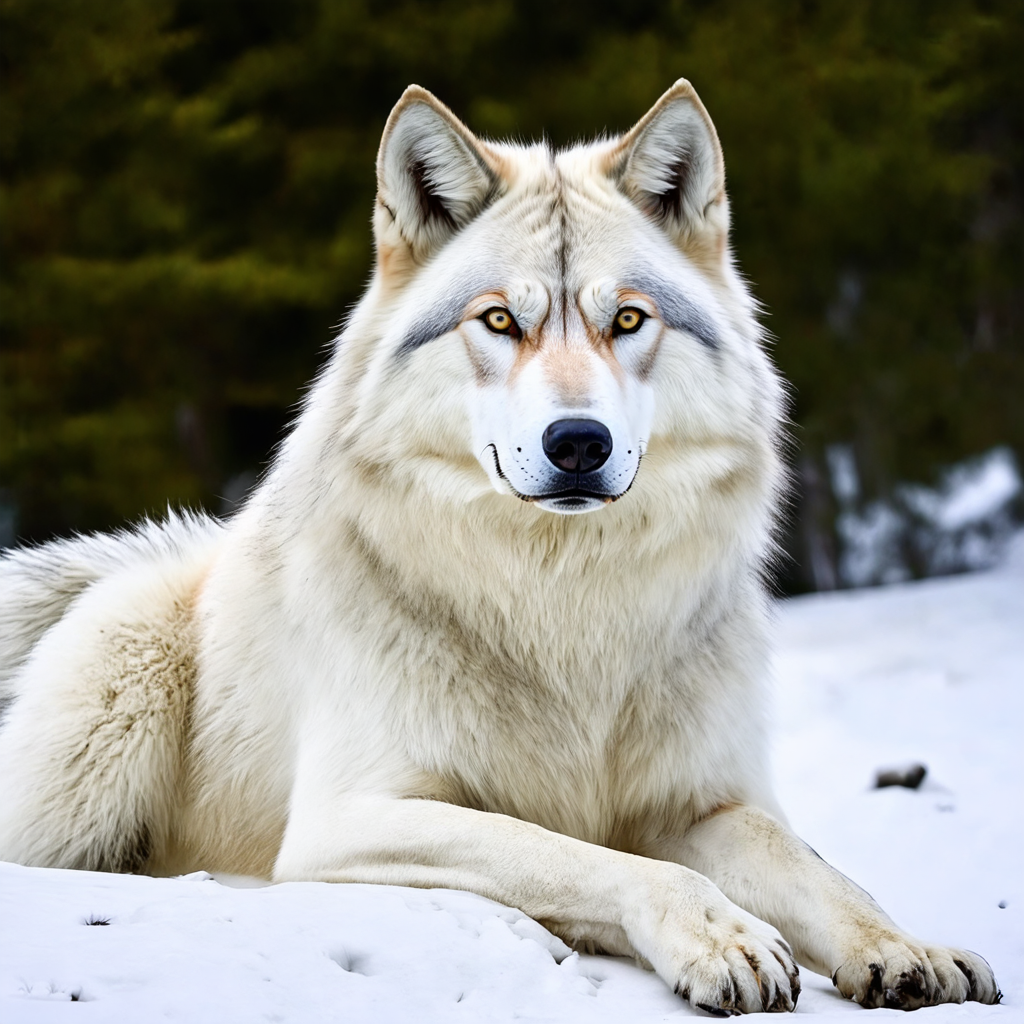} \\

\includegraphics[width=0.12\linewidth]{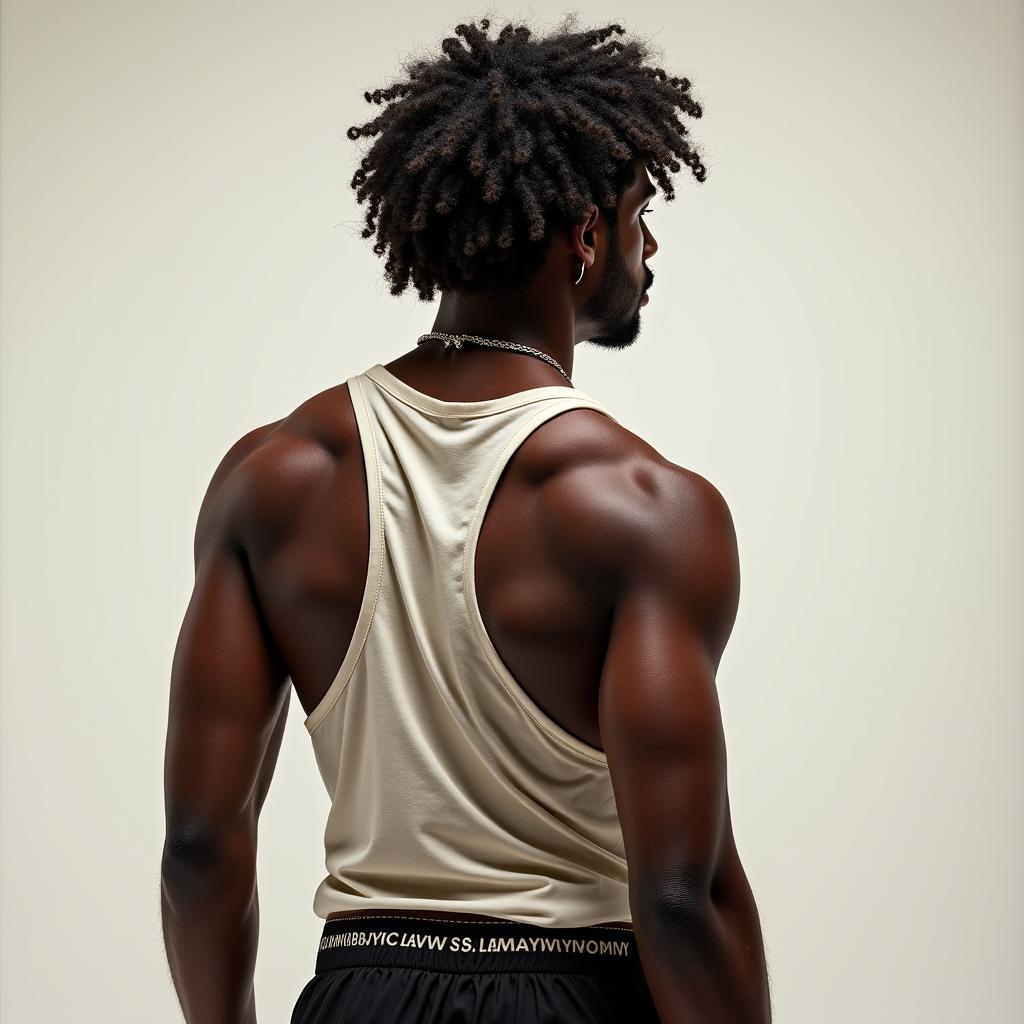} &
\includegraphics[width=0.12\linewidth]{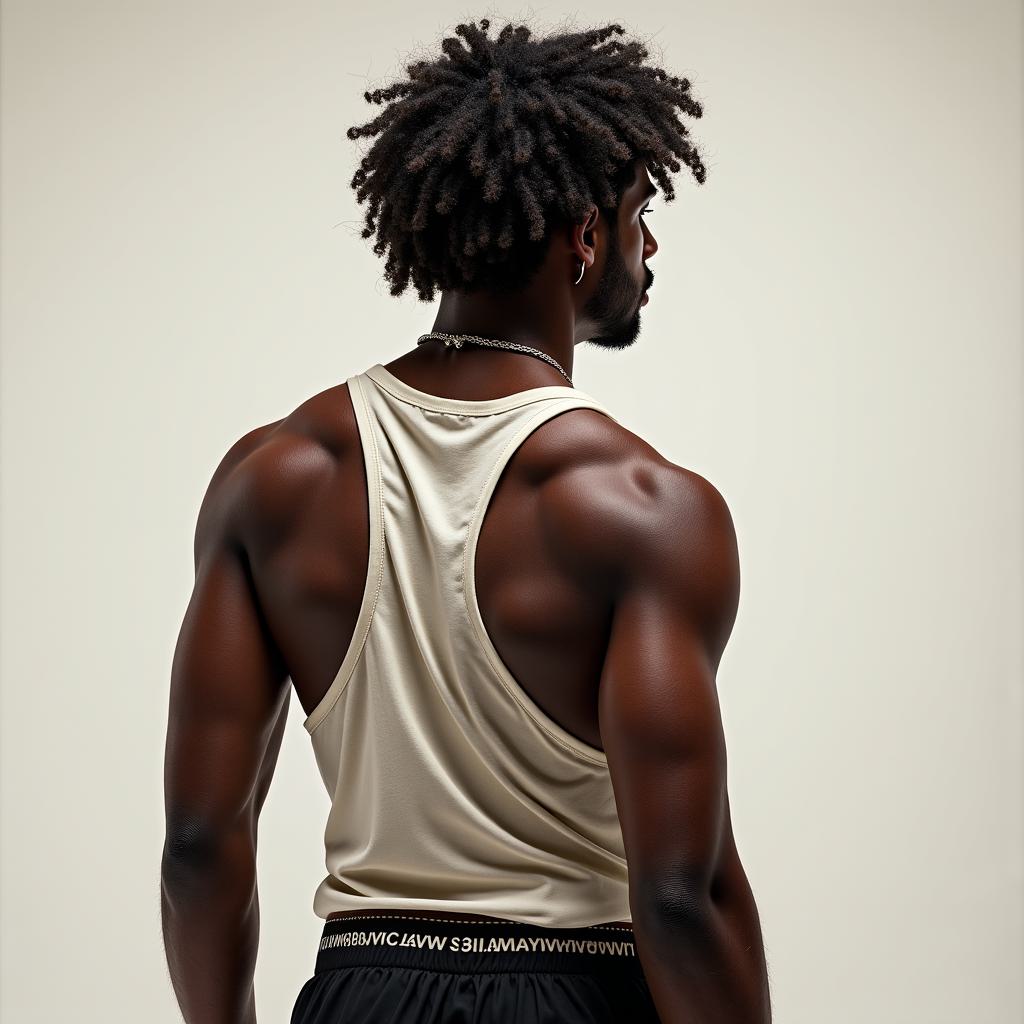} &
\includegraphics[width=0.12\linewidth]{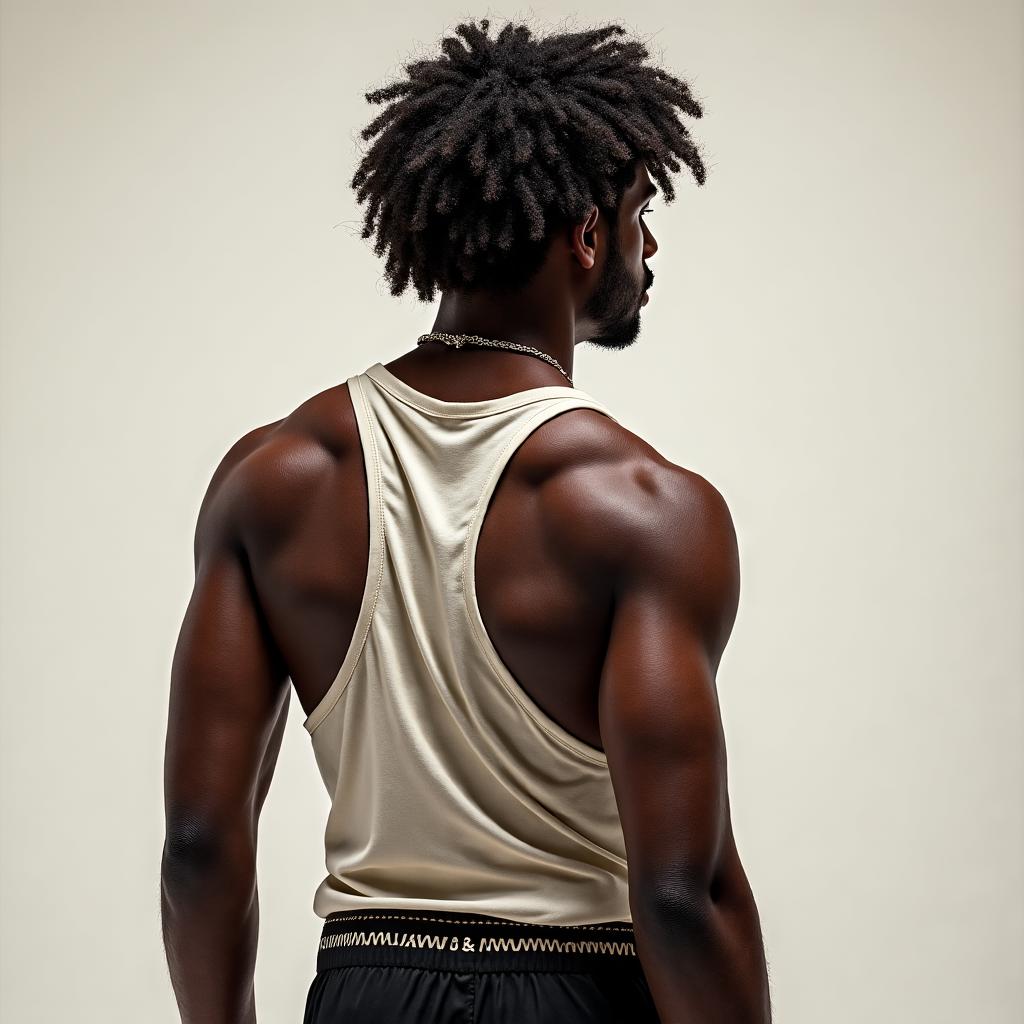} &
\includegraphics[width=0.12\linewidth]{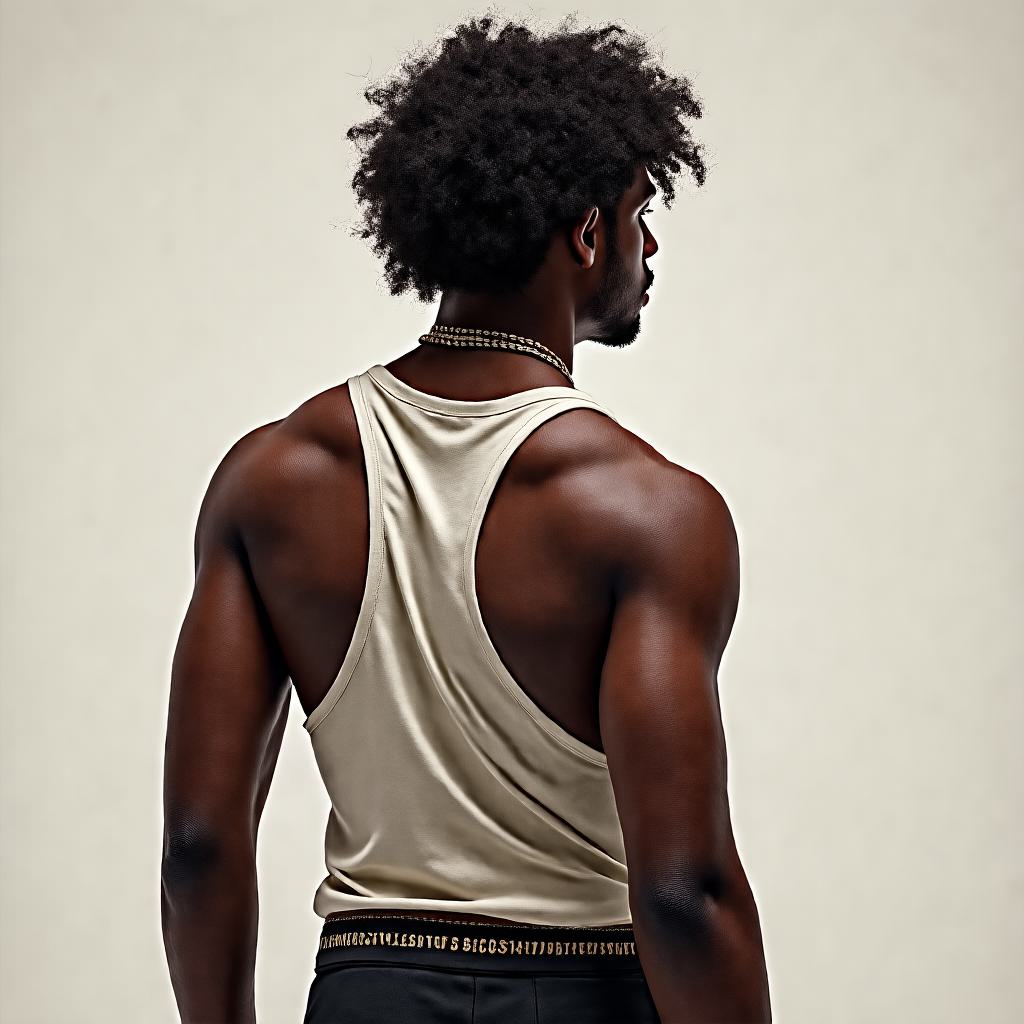} &

\includegraphics[width=0.12\linewidth]{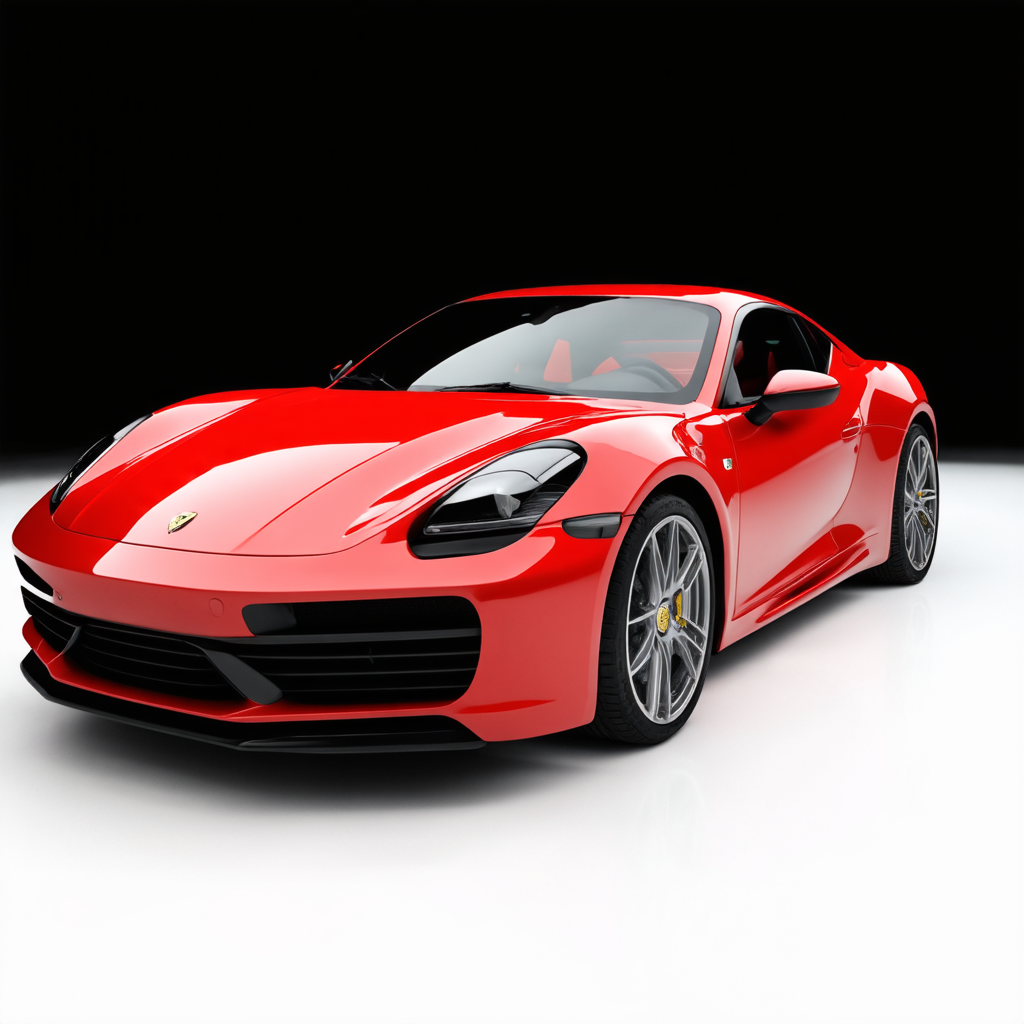} &
\includegraphics[width=0.12\linewidth]{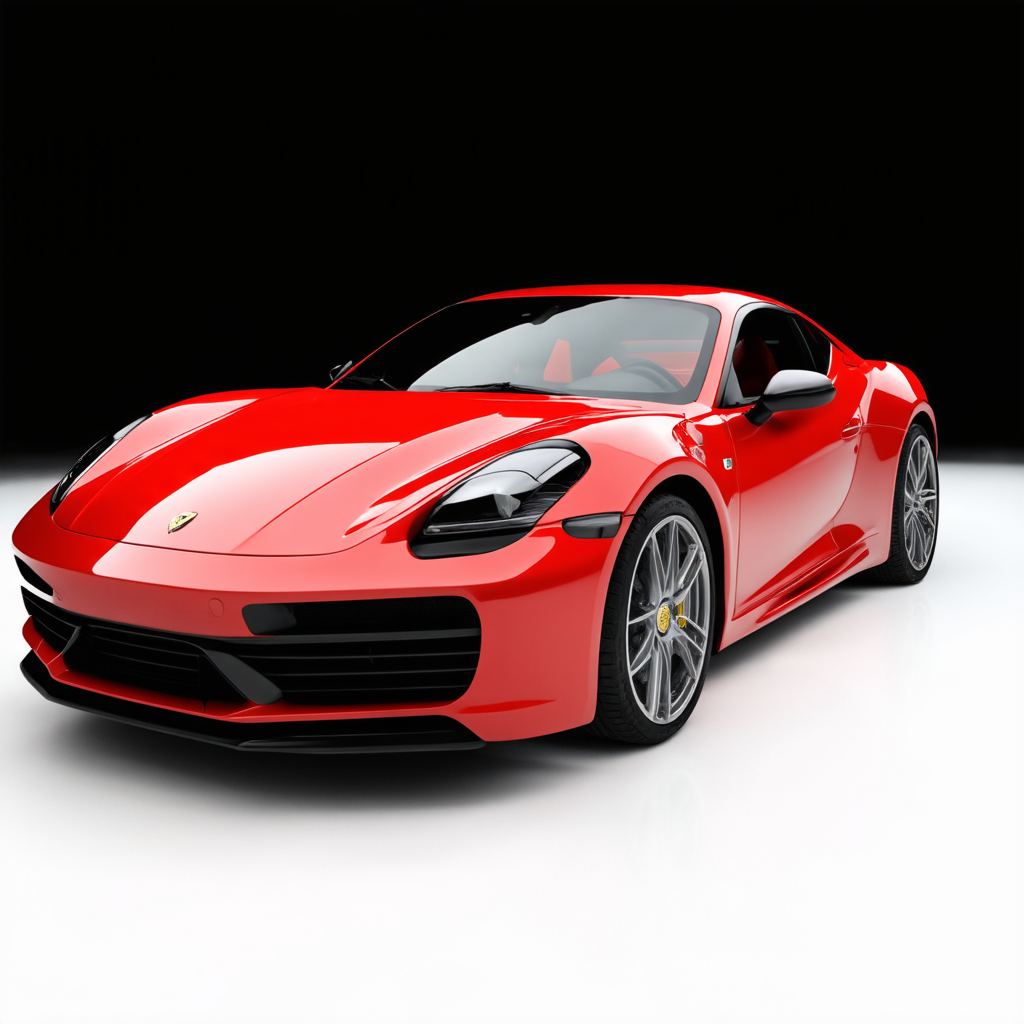} &
\includegraphics[width=0.12\linewidth]{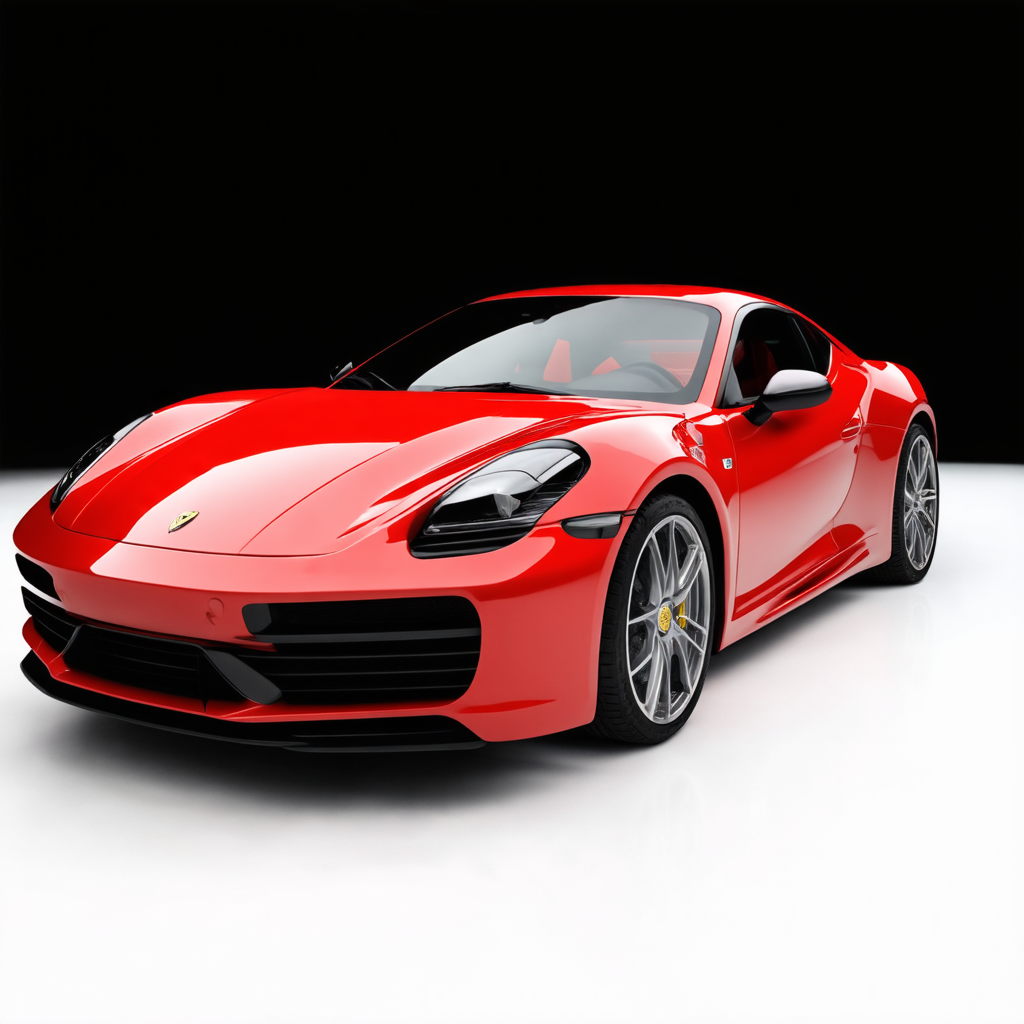} &
\includegraphics[width=0.12\linewidth]{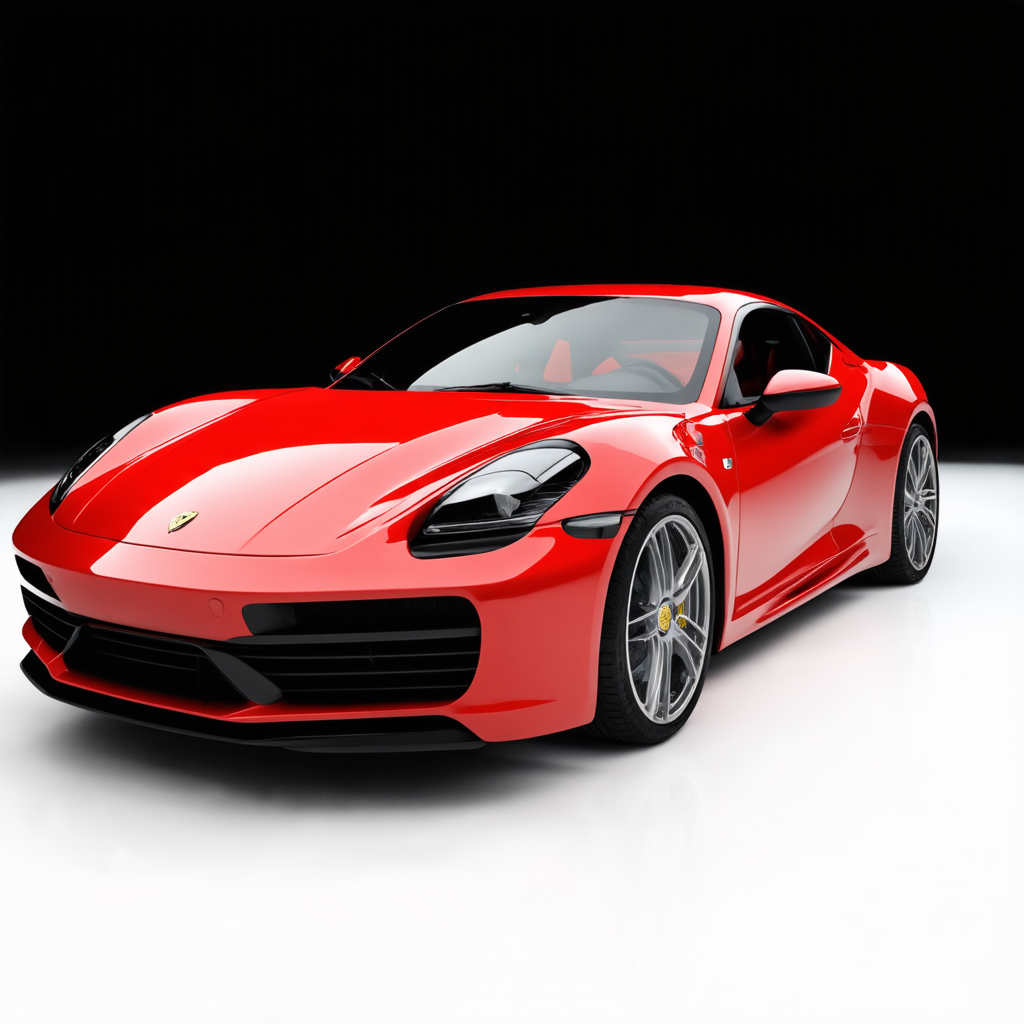} \\

\includegraphics[width=0.12\linewidth]{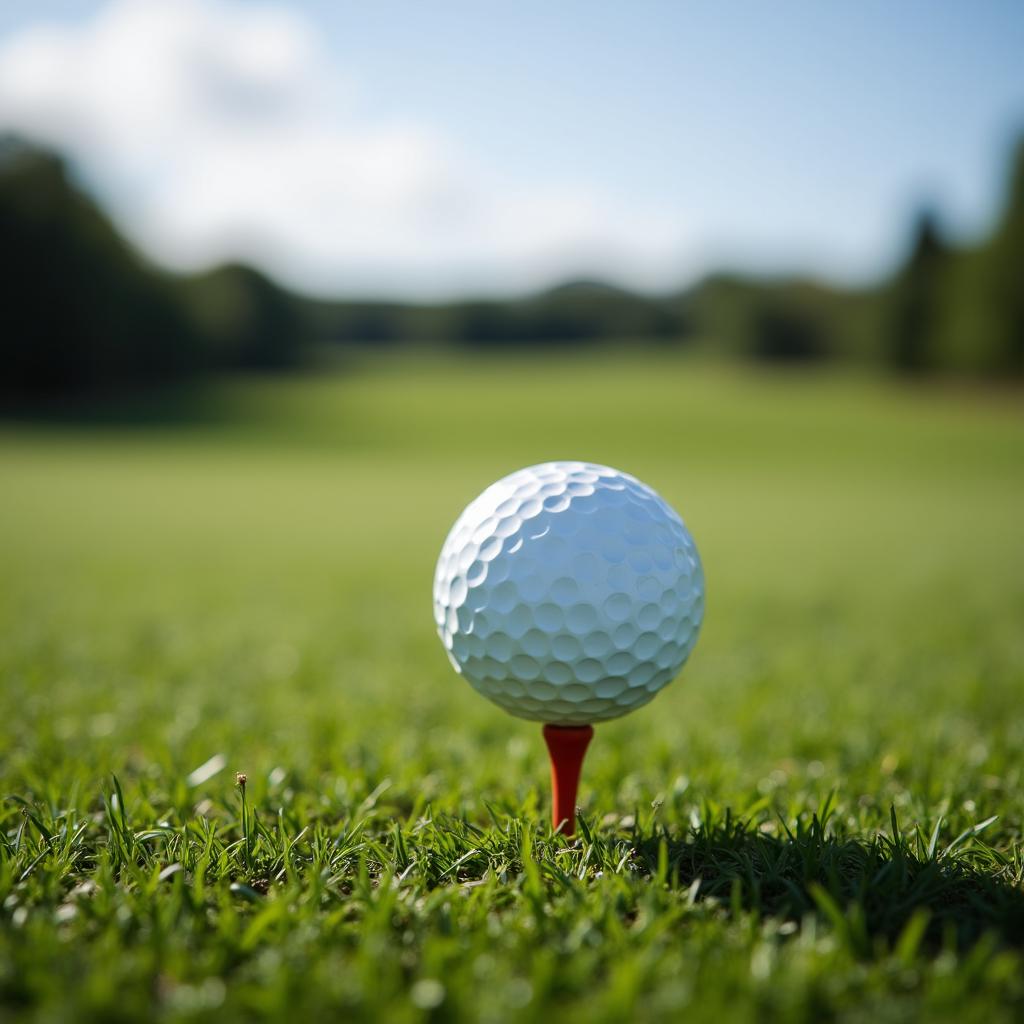} &
\includegraphics[width=0.12\linewidth]{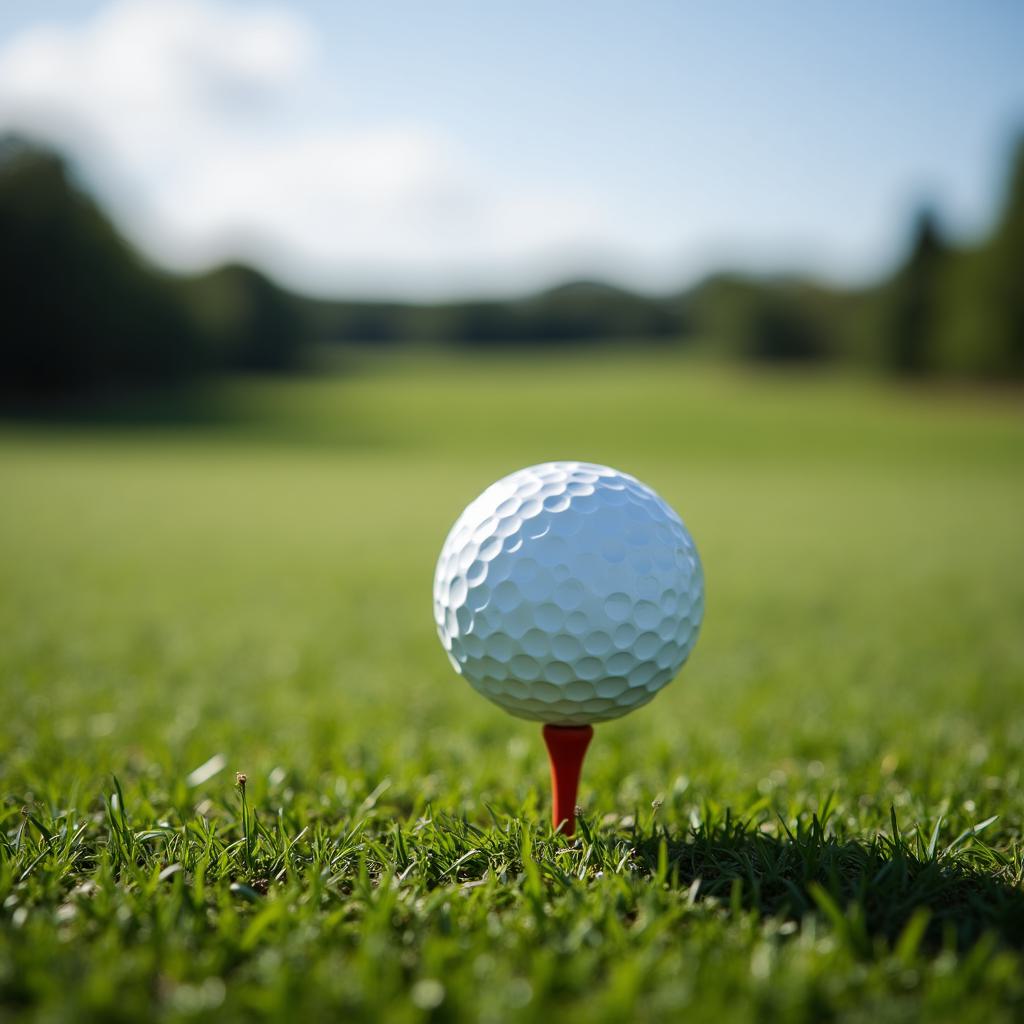} &
\includegraphics[width=0.12\linewidth]{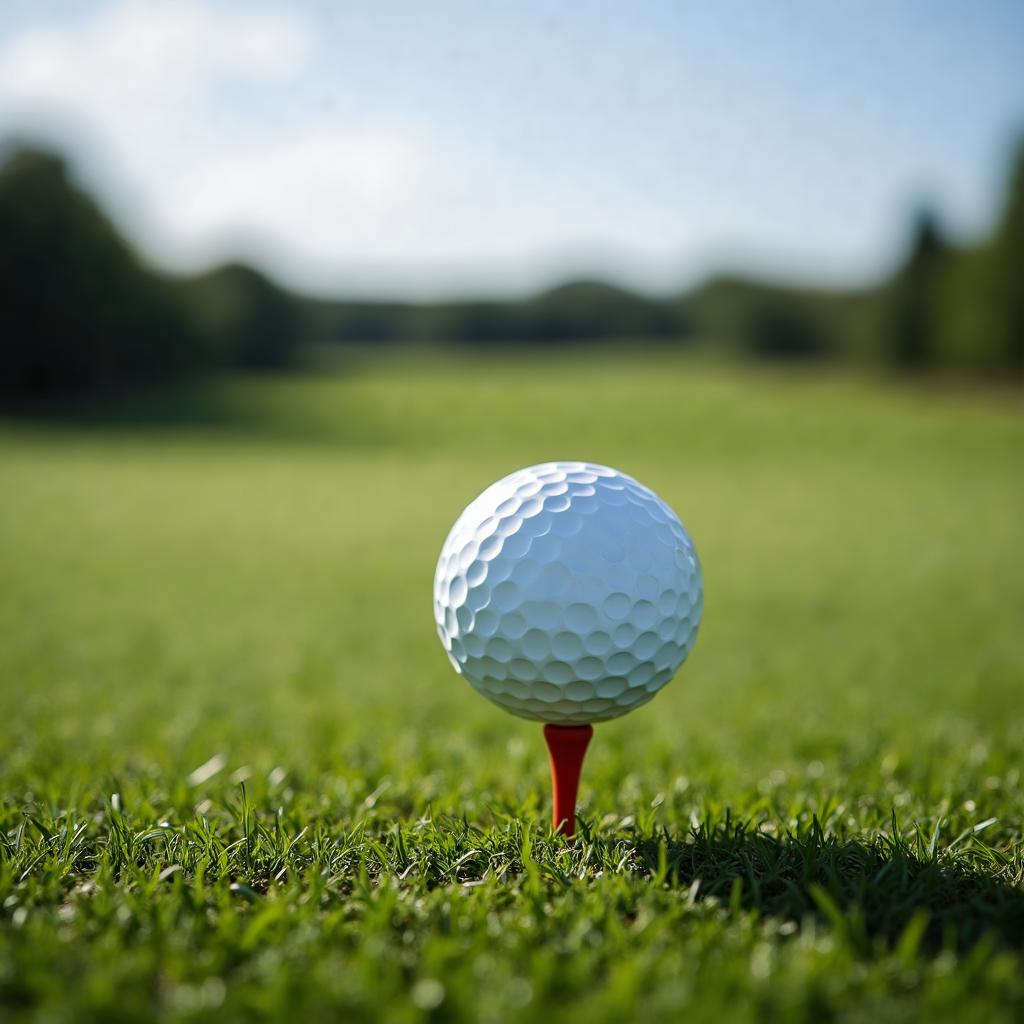} &
\includegraphics[width=0.12\linewidth]{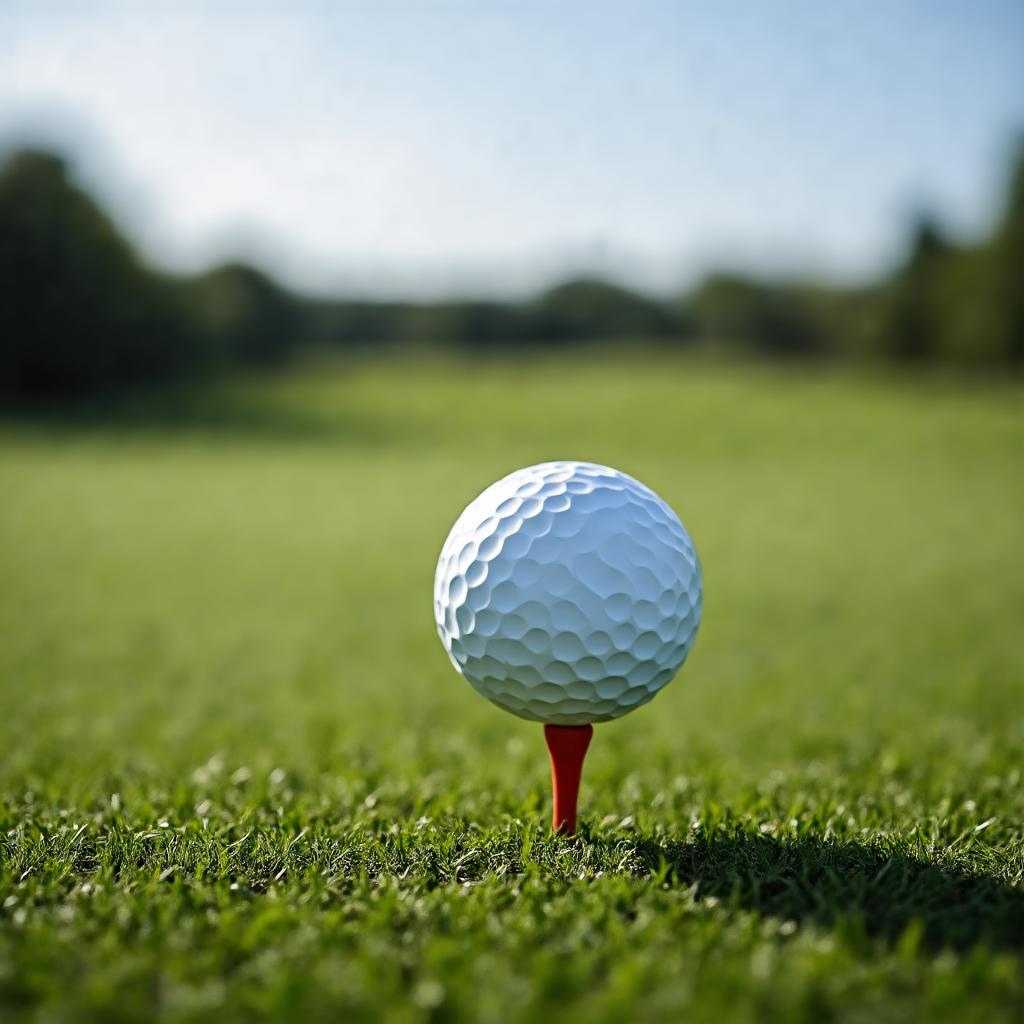} &

\includegraphics[width=0.12\linewidth]{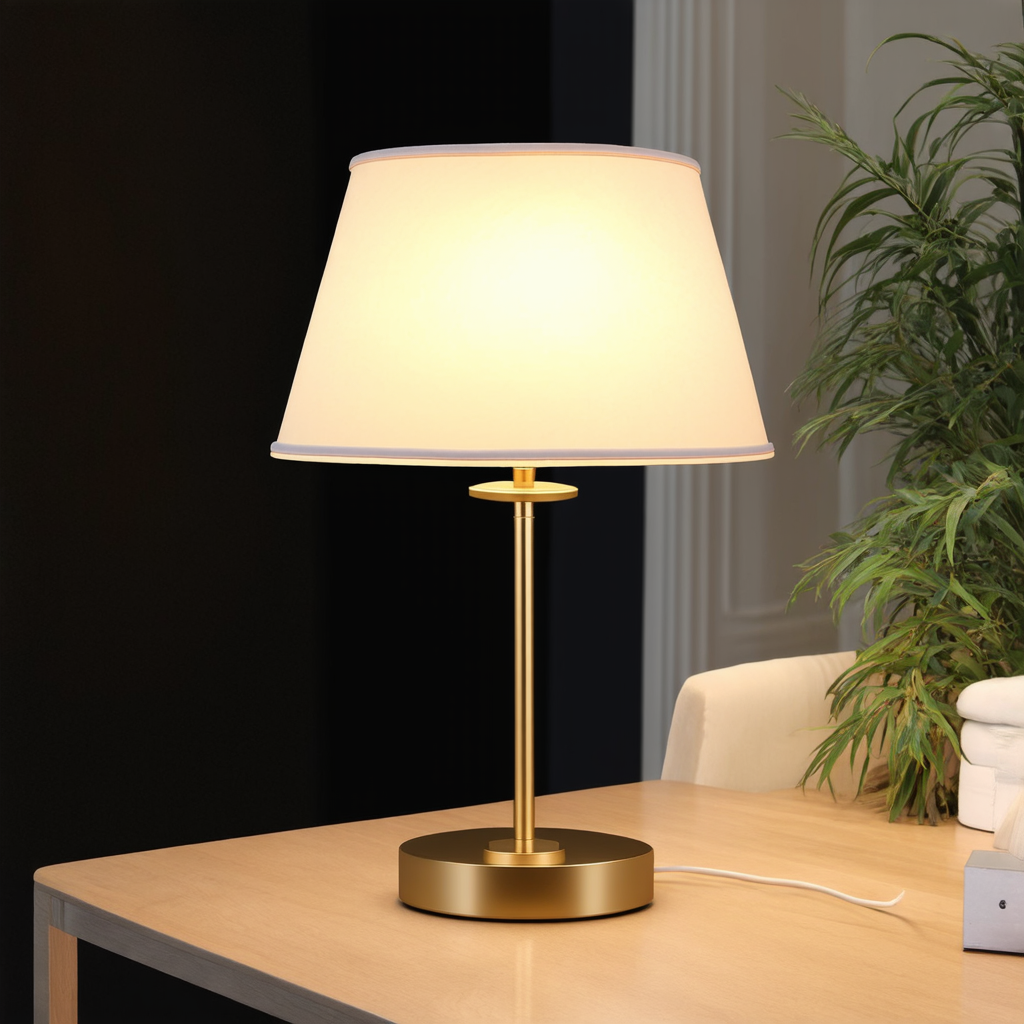} &
\includegraphics[width=0.12\linewidth]{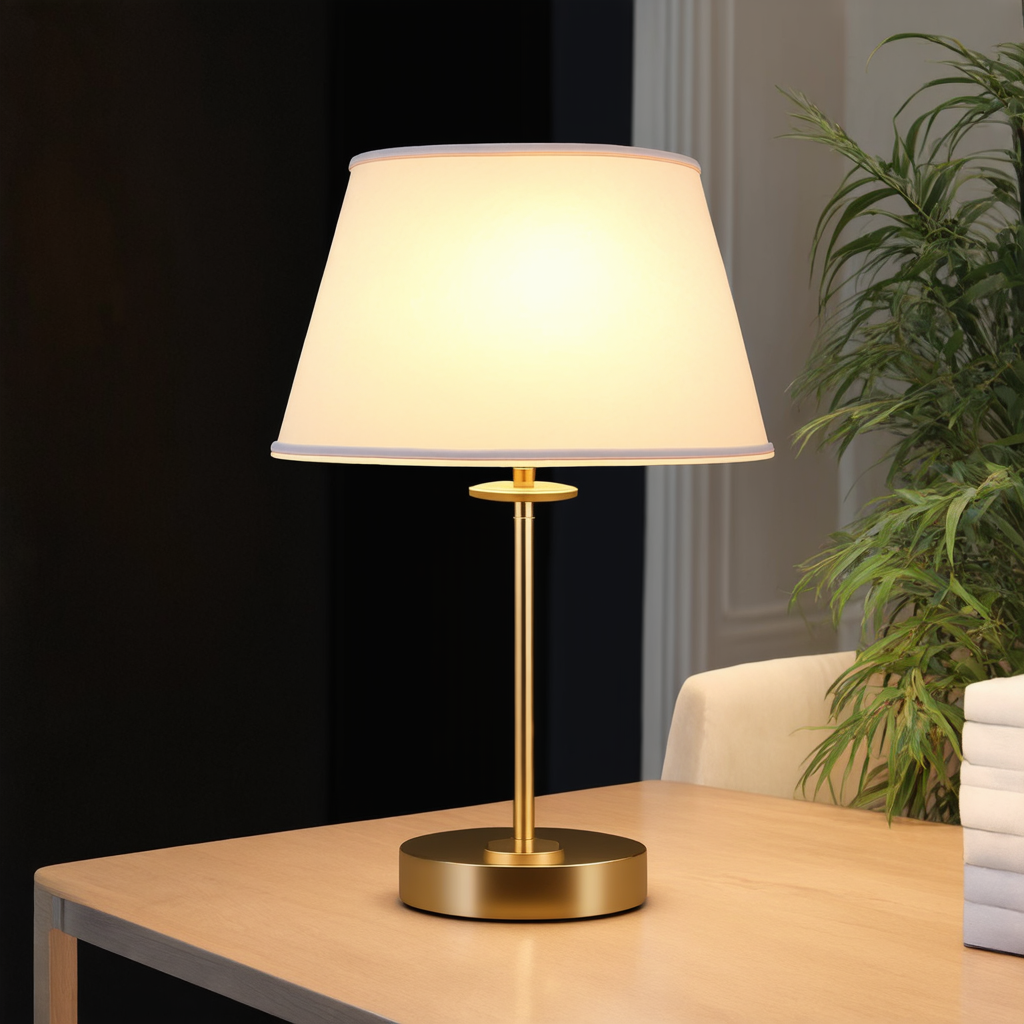} &
\includegraphics[width=0.12\linewidth]{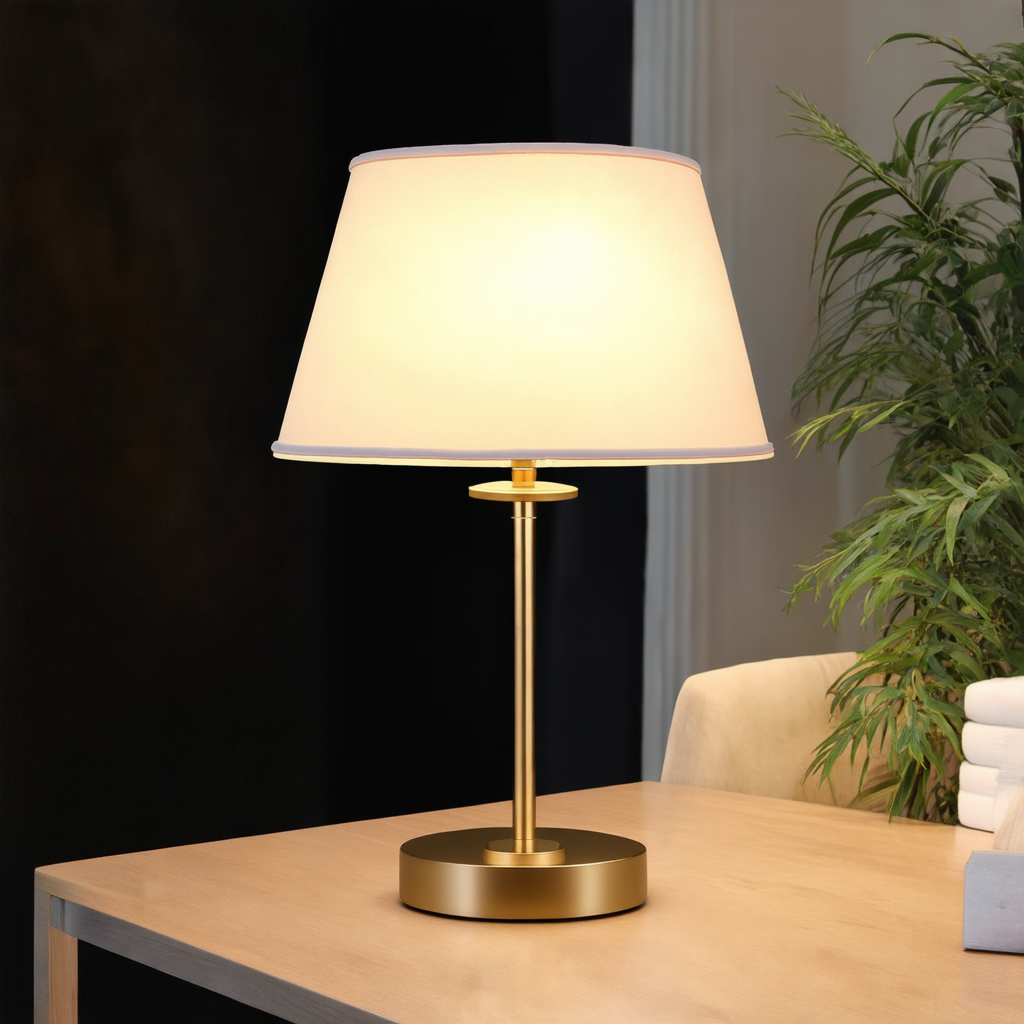} &
\includegraphics[width=0.12\linewidth]{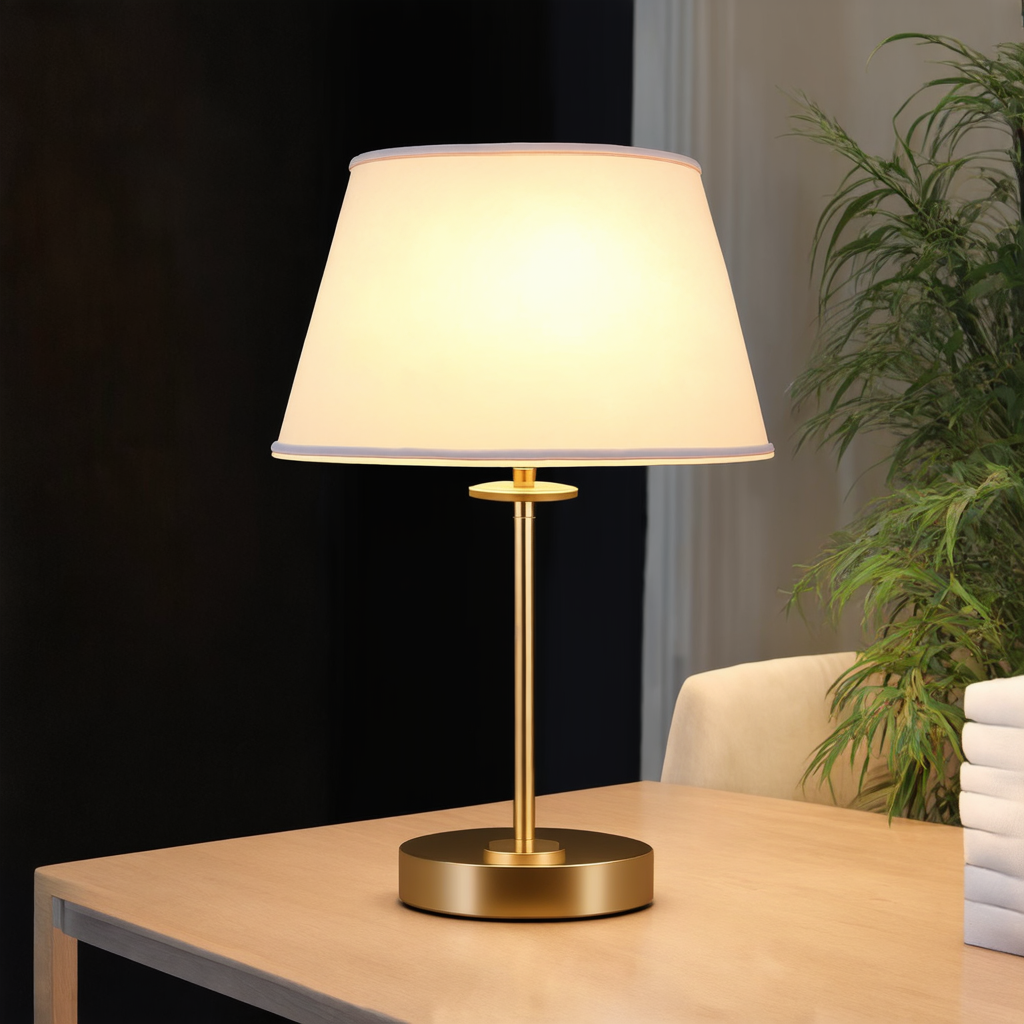} \\

\includegraphics[width=0.12\linewidth]{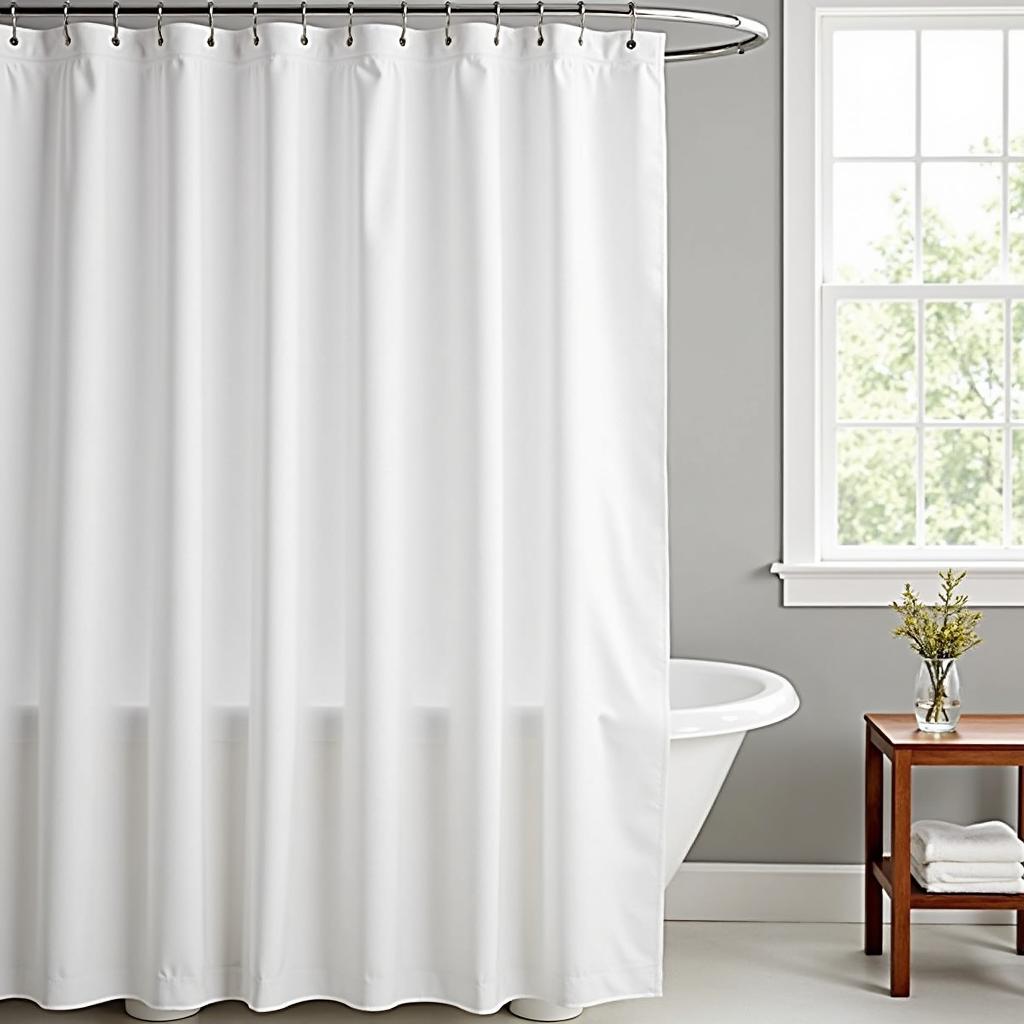} &
\includegraphics[width=0.12\linewidth]{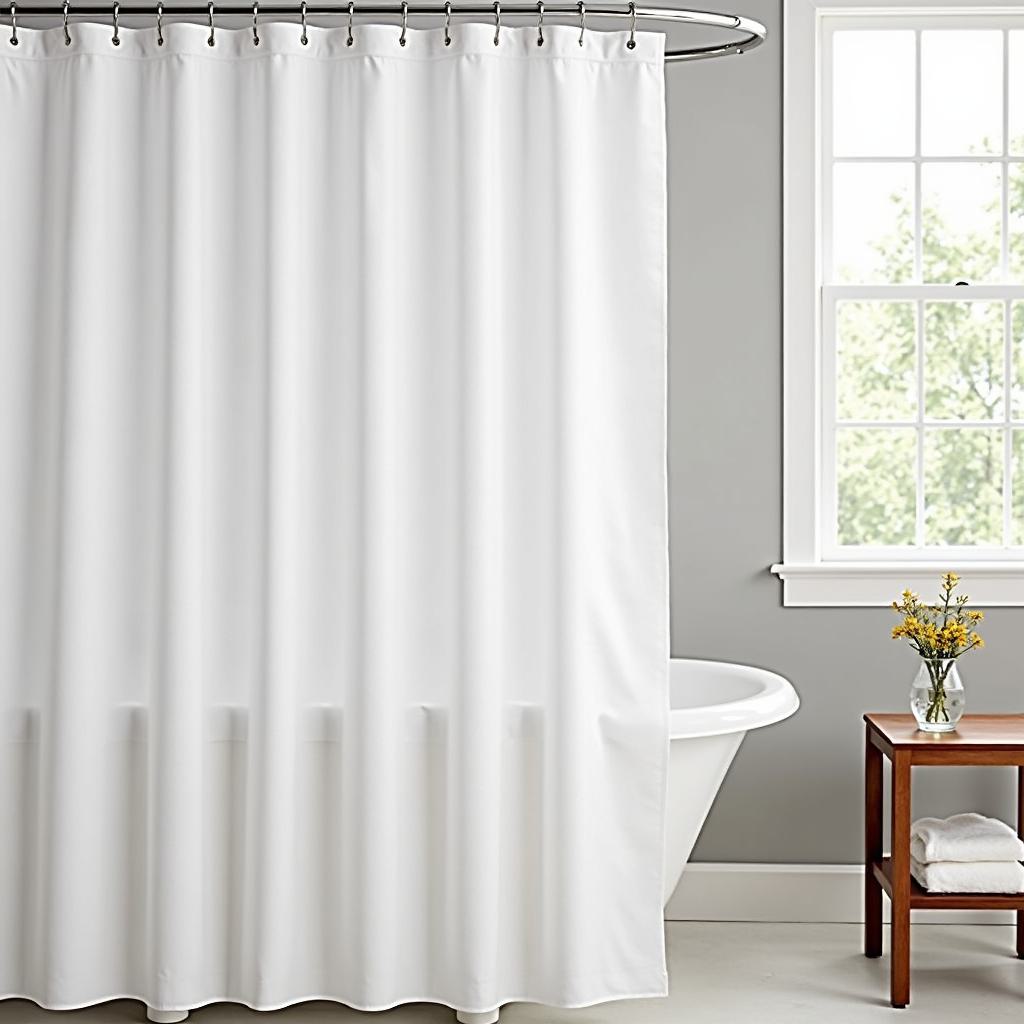} &
\includegraphics[width=0.12\linewidth]{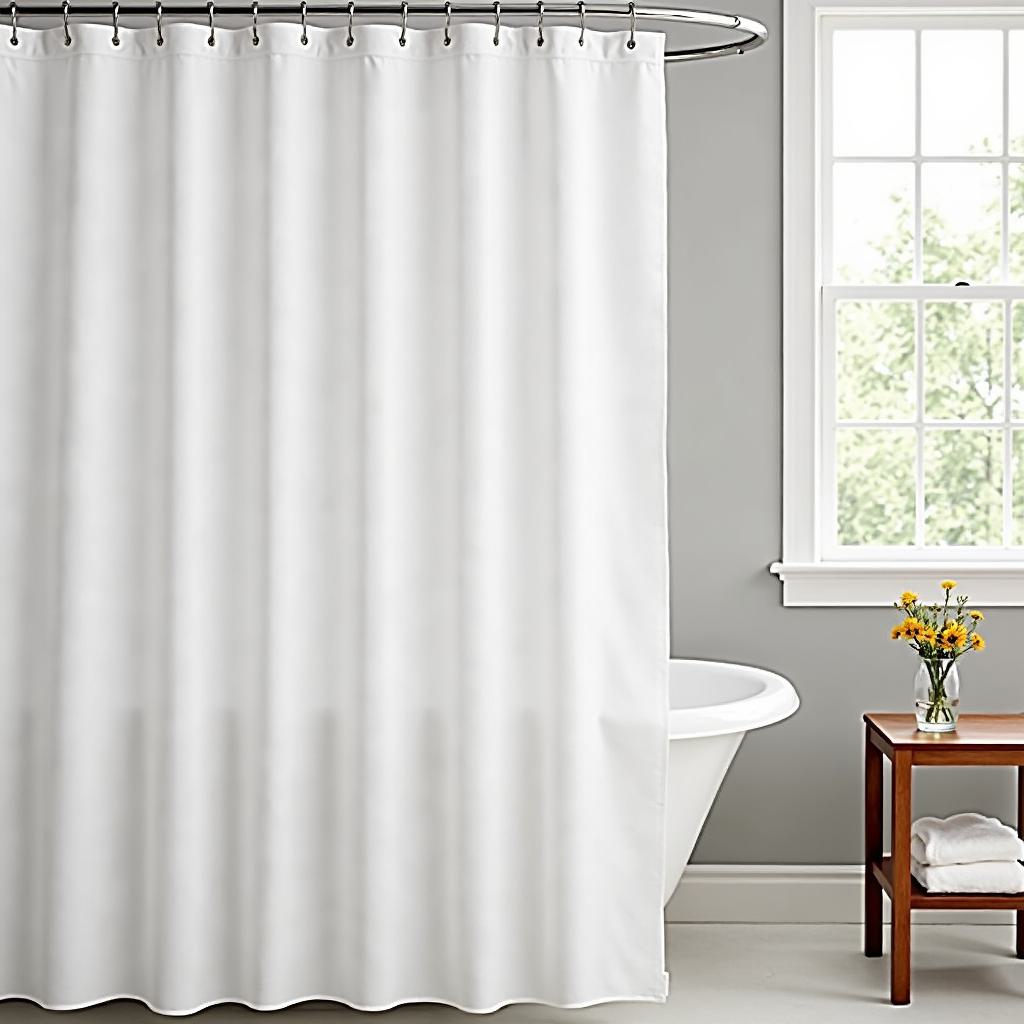} &
\includegraphics[width=0.12\linewidth]{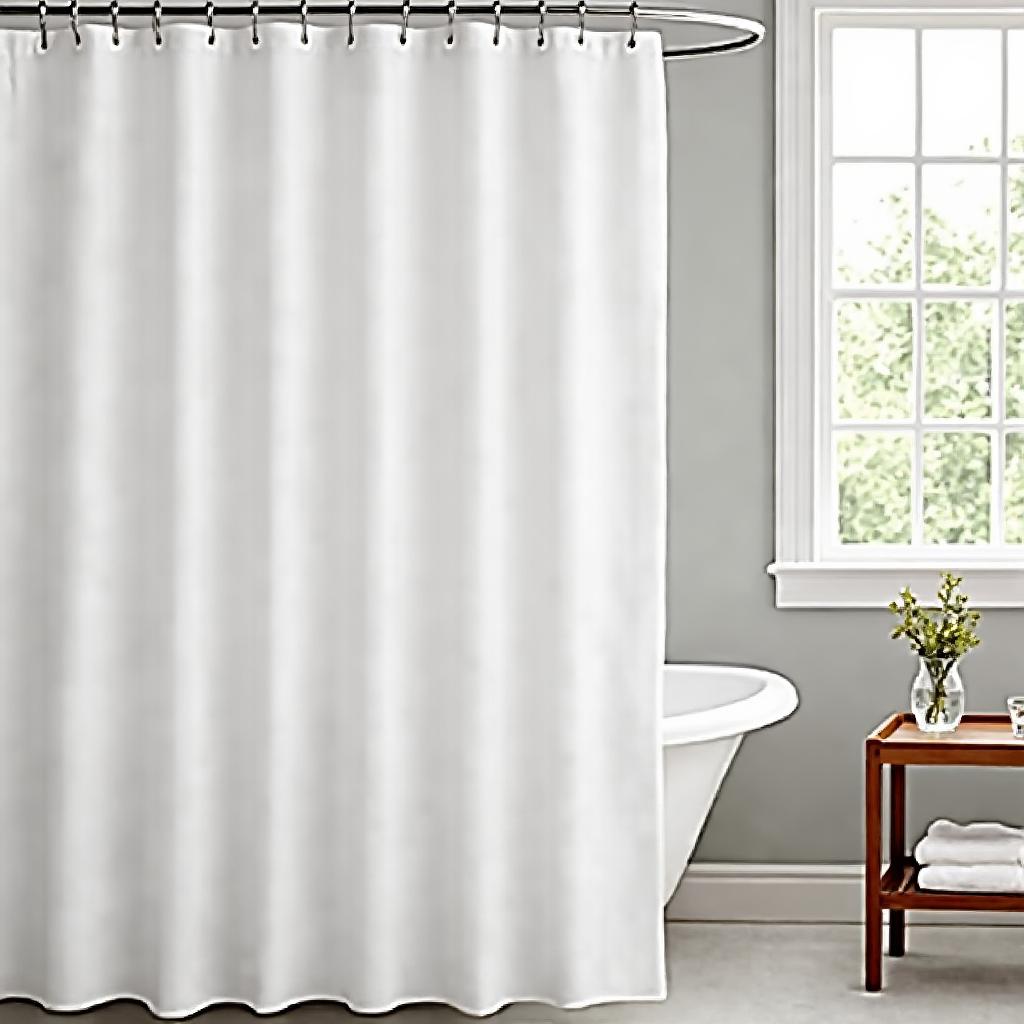} &

\includegraphics[width=0.12\linewidth]{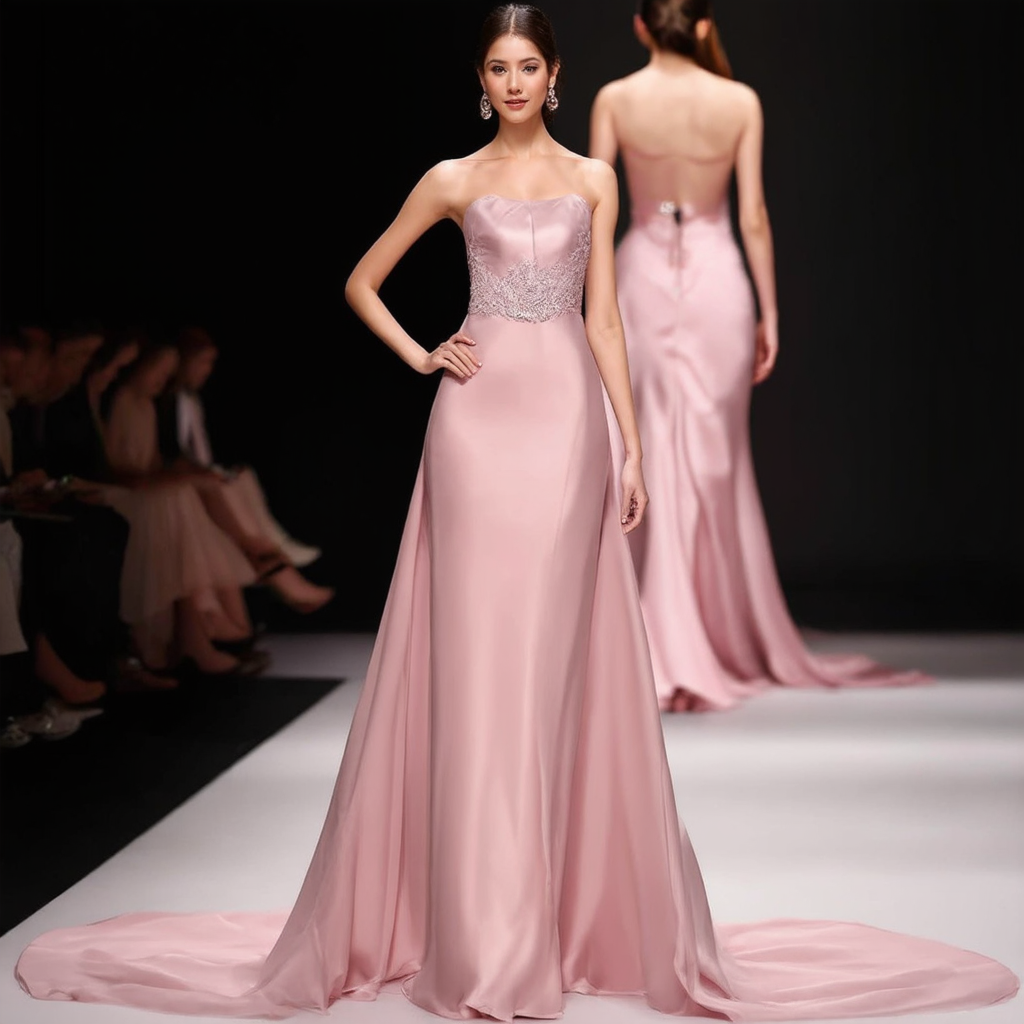} &
\includegraphics[width=0.12\linewidth]{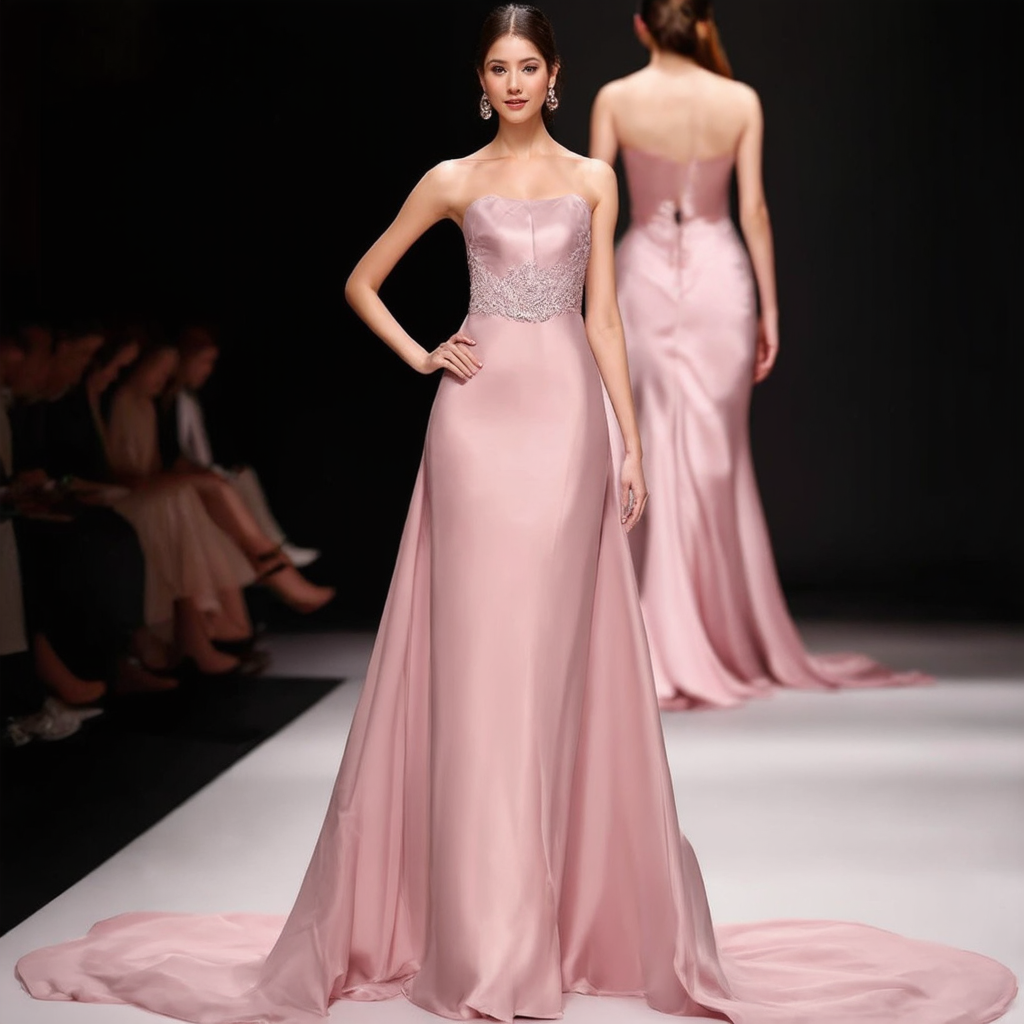} &
\includegraphics[width=0.12\linewidth]{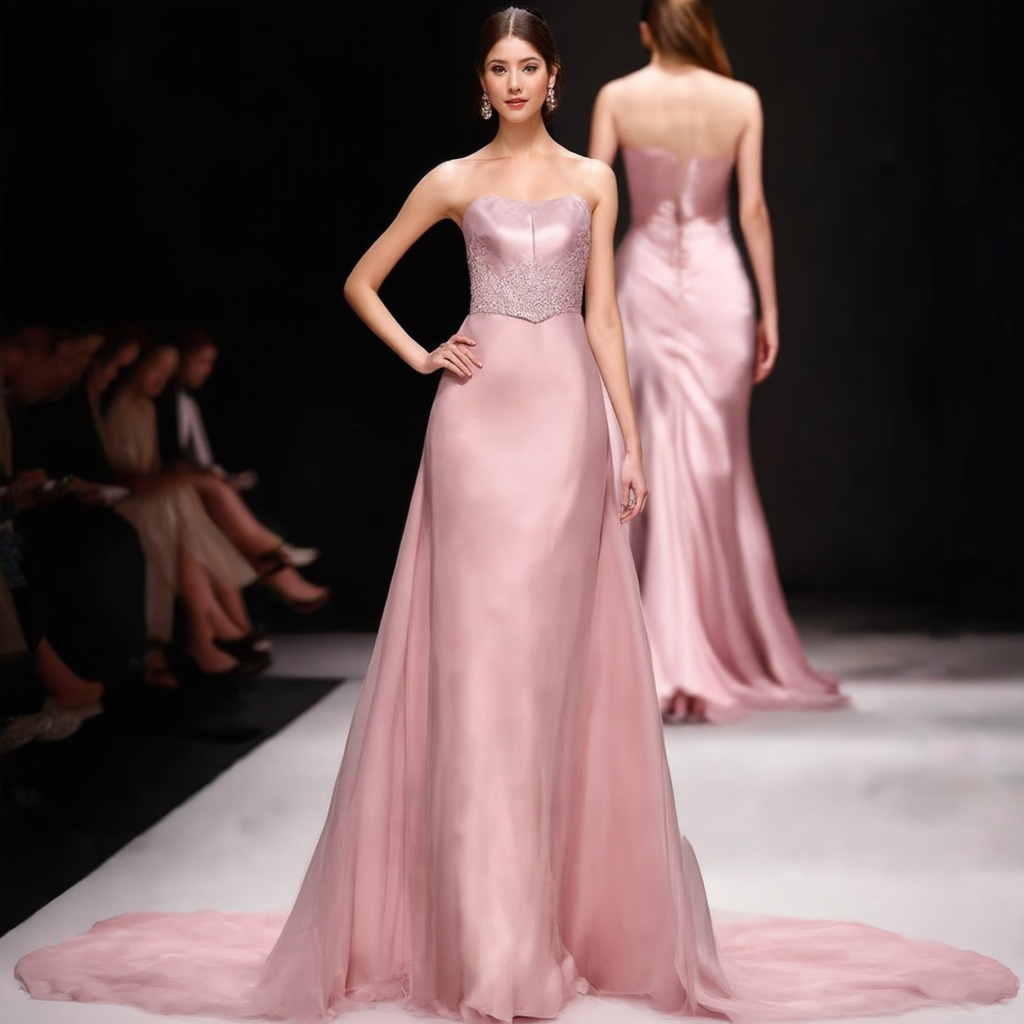} &
\includegraphics[width=0.12\linewidth]{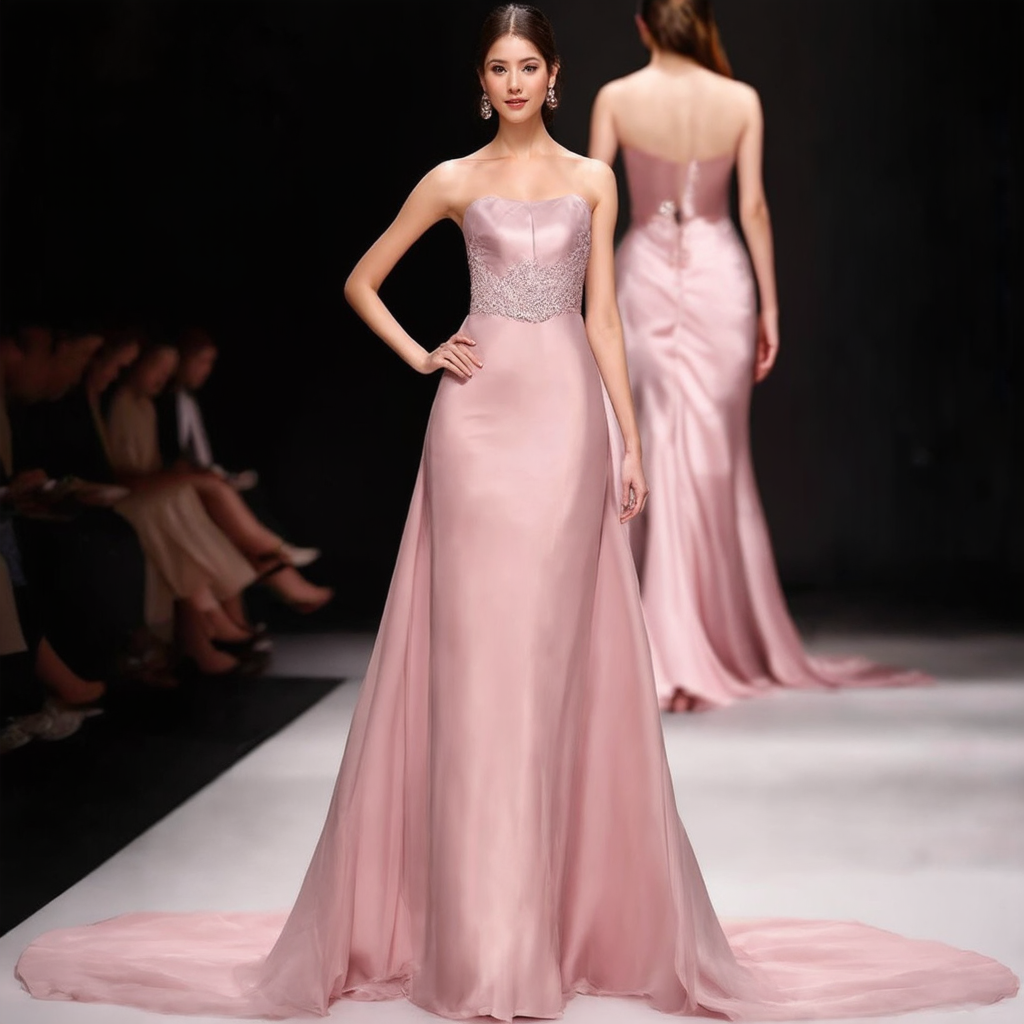} \\
\end{tabular}

\caption{
\textbf{Visualization results of DiTo under varying reduction ratios.} As the token reduction ratio increases, DiTo consistently maintains high visual fidelity and structural consistency.
}
\label{fig:dito_visualization_diffrent_ratios}
\vspace{-1.0em}
\end{figure}

\vspace{0.3em}
\noindent
\textbf{Baselines.}
We evaluate DiTo against an extensive set of representative TR methods, including ToMeSD, ToFu, SiTo, and ToMA. Although ToMeSD, ToFu, and SiTo were originally proposed for UNet models, they fundamentally operate by manipulating token similarity within a general TR pipeline consisting of matching, reduction, and recovery. From this perspective, their core mechanisms are not tied to a specific backbone architecture and can be naturally applied to DiTs. Accordingly, we include them as baselines by adapting their core components to the DiTs. Detailed implementations are provided in the Appendix.

\begin{table}[t]
\centering
\caption{Performance comparison of DiTo against prior methods for generating 1024$\times$1024 images on \texttt{FLUX} (35 sampling steps). Best values are highlighted.}
\label{tab:ratio_results}
\small
\setlength{\tabcolsep}{4pt}
\renewcommand{\arraystretch}{1.1}

\resizebox{\linewidth}{!}{%
\begin{tabular}{c|c|c c c c|c|c c}
\toprule
Ratio & Method & FID$\downarrow$ & PSNR$\uparrow$ & SSIM$\uparrow$ & LPIPS$\downarrow$ & CLIP$\uparrow$ &  Latency$\downarrow$ & $\Delta\downarrow$ \\
\midrule
Baseline & \texttt{Flux} & 31.62 & -- & -- & -- & 23.71 & 26.64 & 0\% \\
\midrule

\multirow{5}{*}{0.25}
& \texttt{DiTo} & \cellcolor{RoyalBlue!12}31.95  & \cellcolor{RoyalBlue!12}25.21 & \cellcolor{RoyalBlue!12}0.8552 & \cellcolor{RoyalBlue!12}0.2385 & 23.85 & \cellcolor{RoyalBlue!12}24.25 & -9.0\% \\
\cmidrule(lr){2-9}
& \texttt{ToMeSD} & 33.76 & 21.59 & 0.7817 & 0.3395 & 23.93  & 24.53 & -7.9\% \\
& \texttt{ToFu}   & 33.75 & 21.70 & 0.7851 & 0.3354 & \cellcolor{RoyalBlue!12}23.94 & 24.47  & -8.1\% \\
& \texttt{SiTo}   & 32.30 & 22.39 & 0.7965 & 0.3462 & 23.93 & 25.25 & -5.2\% \\
& \texttt{ToMA}   & 32.03 & 22.97 & 0.8112 & 0.2716 & 23.78 & 24.28 & -8.9\% \\
\midrule

\multirow{5}{*}{0.50}
& \texttt{DiTo} & \cellcolor{RoyalBlue!12}33.47 & \cellcolor{RoyalBlue!12}22.04 & \cellcolor{RoyalBlue!12}0.7757 & \cellcolor{RoyalBlue!12}0.3329 & \cellcolor{RoyalBlue!12}24.01 & 21.69 & -18.6\% \\
\cmidrule(lr){2-9}
& \texttt{ToMeSD} & 90.00 & 19.73 & 0.6172 & 0.5384 & 23.04 & 22.00 & -17.4\% \\
& \texttt{ToFu}   & 83.79 & 19.81 & 0.6328 & 0.5302 & 23.05 & 22.02 & -17.3\% \\\
& \texttt{SiTo}   & 61.05 & 20.17 & 0.6503 & 0.5245 & 23.45 & 22.54 & -15.4\% \\
& \texttt{ToMA}   & 43.71 & 20.49 & 0.6936 & 0.4386  & 23.90 & \cellcolor{RoyalBlue!12}21.62 & -18.8\% \\
\bottomrule
\end{tabular}%
}
\label{table:quantitative_results_flux}
\vspace{-1.5em}
\end{table}

\subsection{Experimental Results}
\textbf{Quantitative Results.}
\cref{table:quantitative_results_flux} presents quantitative comparisons between DiTo and existing TR methods (ToMeSD, ToFu, SiTo, and ToMA) on \texttt{Flux}. Overall, DiTo consistently outperforms all methods in terms of generation quality across multiple metrics, including FID, PSNR, SSIM, and LPIPS. Notably, DiTo demonstrates substantial improvements in perceptual reconstruction fidelity, achieving PSNR gains of 2.24–3.62 dB at a 0.25 ratio and 1.55–2.31 dB at a 0.50 ratio. These gains are also reflected in other key metrics, including SSIM and LPIPS. These results align with the design objective of DiTo, which explicitly aims to minimize recovery error by incorporating output token similarity into the TR process. As these metrics directly measure perceptual fidelity with respect to the vanilla model output, the consistent gains confirm that DiTo effectively achieves its intended objective. Furthermore, DiTo exhibits exceptional robustness to the reduction ratio; while existing methods suffer from substantial degradation in FID scores at a 0.50 ratio (ranging from 43.71 to 90.00), DiTo maintains a significantly more stable FID of 33.47.

Regarding semantic alignment, DiTo achieves CLIP scores (23.85–24.01) comparable to or even slightly exceeding the vanilla model (23.71), indicating that text-to-image consistency is well preserved. In terms of efficiency, by reusing matching results, DiTo achieves up to an 18.6\% latency reduction (21.69s), which is comparable to the runtime of ToMA—the fastest among the compared TR methods—while providing superior generation quality. \cref{table:quantitative_results_sd3} reports the results on \texttt{SD3}, where similar trends are consistently observed. These results demonstrate that DiTo generalizes well across different models and remains robust even with fewer diffusion timesteps (e.g., 35 steps in \texttt{Flux}).

\begin{table}[t]
\centering
\caption{Performance comparison of DiTo against prior methods for generating 1024$\times$1024 images on \texttt{SD3} (50 sampling steps). Best values are highlighted.}
\label{tab:ratio_results}
\small
\setlength{\tabcolsep}{4pt}
\renewcommand{\arraystretch}{1.1}

\resizebox{\linewidth}{!}{%
\begin{tabular}{c|c|c c c c|c|c c}
\toprule
Ratio & Method & FID$\downarrow$ & PSNR$\uparrow$ & SSIM$\uparrow$ & LPIPS$\downarrow$ & CLIP$\uparrow$ & Latency$\downarrow$ & $\Delta\downarrow$ \\
\midrule
Baseline & \texttt{SD3} & 27.67 & -- & -- & -- & 24.67 & 9.32 & 0\% \\
\midrule

\multirow{5}{*}{0.25}
& \texttt{DiTo} & \cellcolor{RoyalBlue!12}27.83 & \cellcolor{RoyalBlue!12}26.83 & \cellcolor{RoyalBlue!12}0.9160 & \cellcolor{RoyalBlue!12}0.1492 & 24.77 & \cellcolor{RoyalBlue!12}8.63 & -7.4\% \\
\cmidrule(lr){2-9}
& \texttt{ToMeSD} & 28.42 & 23.32 & 0.8697 & 0.2045 & 24.74 & 8.82 & -5.3\% \\
& \texttt{ToFu}   & 27.99 & 23.83 & 0.8835 & 0.1922 & 24.72 & 8.87  & -4.9\% \\
& \texttt{SiTo}   & 27.92 & 23.01 & 0.8682 & 0.2211 & \cellcolor{RoyalBlue!12}24.79 & 9.30 & -0.2\% \\
& \texttt{ToMA}   & 28.47 & 22.93 & 0.8624 & 0.2029 & 24.63 & 9.08 & -2.5\% \\
\midrule

\multirow{5}{*}{0.50}
& \texttt{DiTo} & \cellcolor{RoyalBlue!12}28.89
 & \cellcolor{RoyalBlue!12}23.16
 & \cellcolor{RoyalBlue!12}0.8510 & \cellcolor{RoyalBlue!12}0.2577 & \cellcolor{RoyalBlue!12}24.87 & \cellcolor{RoyalBlue!12}7.79 & -16.4\% \\
\cmidrule(lr){2-9}
& \texttt{ToMeSD} & 31.45 & 21.22 & 0.8090 & 0.3139 & 24.80 & 8.00 & -14.2\% \\
& \texttt{ToFu}   & 31.89 & 21.21 & 0.8112 & 0.3187 & 24.79 & 8.05 & -13.6\% \\
& \texttt{SiTo}   & 33.13 & 21.23 & 0.8046 & 0.3170 & 24.82 & 8.42 & -9.7\% \\
& \texttt{ToMA}   & 33.37 & 20.91 & 0.8049 & 0.3053 & 24.62 & 8.17 & -12.3\% \\
\bottomrule
\end{tabular}%
}
\label{table:quantitative_results_sd3}
\vspace{-1.0em}
\end{table}
\vspace{0.3em}
\noindent
\textbf{Qualitative Comparison.}
\cref{fig:visual_comparison_w_prior_tr} presents qualitative comparisons between DiTo and existing token reduction methods (ToMeSD, ToFu, SiTo, and ToMA) on \texttt{Flux} and \texttt{SD3}. Under the equivalent FLOPs reduction, DiTo consistently preserves higher visual fidelity and structural consistency across both models, whereas prior methods exhibit noticeable degradation. Additional qualitative comparisons are provided in the Appendix.
\cref{fig:dito_visualization_diffrent_ratios} further presents visualization results of DiTo across different reduction ratios on both models, demonstrating that image quality remains stable even as the reduction ratio increases. Additional visualizations are provided in the Appendix.

\vspace{0.3em}
\noindent
\textbf{Ablation \& Others.}
To validate our design decisions, we conduct comprehensive ablation studies (Appendix), evaluating the impact of the PMR threshold $\tau$ and penalty strength $\lambda$ to identify the optimal balance between efficiency and quality. Furthermore, we explore detailed configurations for TR, specifically examining the choice of reduction strategies (e.g., pruning/merging) and various similarity metrics. Finally, we provide a comparative analysis of FLOPs (Appendix) and memory consumption (Appendix) across different reduction methods. Our profiling confirms that while DiTo maintains matching results throughout the Reduction steps, the associated memory overhead is negligible, as the stored information consists of compact index metadata rather than high-dimensional token features.

\vspace{-0.5em}
\section{Conclusion}
\vspace{-0.5em}
In this work, we introduce \textbf{DiTo}, a novel paradigm that shifts toward output-centric token reduction to explicitly minimize recovery error. To achieve this, DiTo leverages the temporal consistency of DMs and establishes token correspondences using prior-step similarities as a proxy at a Matching step, which are then reused across subsequent Reduction steps. To optimize this scheduling, we propose PMR-guided Interval Scheduling to determine the optimal matching frequency. Additionally, to mitigate localized approximation errors and blocking artifacts from repeated reuse, we introduce Frequency-aware Token Matching with a selection frequency penalty. Extensive experiments demonstrate that DiTo effectively breaks the quality-efficiency trade-off, achieving a superior Pareto frontier compared to existing methods.



%
%
\bibliographystyle{splncs04}
\bibliography{main}


\end{document}